\documentclass[parskip=full, 10pt]{scrartcl}

\usepackage[left=2.5cm,right=2.5cm,
    top=2cm,bottom=2cm,bindingoffset=0cm]{geometry}
\usepackage{tipa}
\usepackage{graphicx} %
\usepackage{amsmath,amssymb}
\usepackage{pdflscape}
\usepackage{geometry}
\usepackage{booktabs}
\usepackage{multirow}
\usepackage{caption}
\usepackage{subcaption}
\usepackage{wrapfig}
\usepackage{xspace}
\usepackage[dvipsnames]{xcolor}
\usepackage{url}
\usepackage{float}
\usepackage{tabularray}
\usepackage[counterclockwise]{rotating}
\usepackage{xcolor}
\usepackage{color}
\usepackage{verbatim}
\usepackage{hyperref}
\usepackage{booktabs}
\usepackage{tablefootnote}
\usepackage{blindtext}
\usepackage{algorithm}
\usepackage{algpseudocode}
\usepackage[section]{placeins}
\usepackage{algorithmicx}

\usepackage{arabtex}
\usepackage{arabaux}
\usepackage{utf8}
\setcode{utf8}
\usepackage{hyperref}
\usepackage{booktabs}
\usepackage{mdframed}
\hypersetup{
    colorlinks=true,
    urlcolor=blue,
    citecolor=blue,
    linkcolor=blue
}
\usepackage{lipsum}
\usepackage{array}
\usepackage{wrapfig}
\usepackage[authoryear]{natbib}
\usepackage{enumerate}
\usepackage{authblk}

\restylefloat{table}
\usepackage[colorinlistoftodos,prependcaption,textsize=small]{todonotes}
\presetkeys{todonotes}{inline}{}

\renewcommand{\abstractname}{Abstract}

\NewDocumentCommand{\mourad}{ mO{} }{\textcolor{blue}{\textsuperscript{\textit{Mourad}}\textsf{\textbf{\small[#1]}}}}

\newcommand{\FS}{\textbf{Fanar Star}}
\newcommand{\FP}{\textbf{Fanar Prime}}
\newcommand{\bt}[1]{\textbf{#1}}
\definecolor{colour1}{gray}{1.0}
\definecolor{colour2}{gray}{0.95}

\makeatletter
\newcommand\fullwidthbox[1]{%
    \noindent\makebox[\textwidth]{%
    \begin{minipage}{\dimexpr\textwidth+2\fboxsep}
    \begin{mdframed}[
        backgroundcolor=blue!10,    %
        linecolor=black,            %
        linewidth=1pt,              %
        innertopmargin=10pt,        %
        innerbottommargin=10pt,     %
        innerleftmargin=10pt,       %
        innerrightmargin=10pt,      %
        skipabove=\baselineskip,    %
        skipbelow=\baselineskip     %
    ]
    #1
    \end{mdframed}
    \end{minipage}%
    }%
}

\makeatother
\title{Fanar \\
An Arabic-Centric Multimodal Generative AI Platform}

\author{FANAR TEAM\footnote{The author ordering is alphabetical by last name. See Section~\ref{sec:contribution} for contribution details.}}
\author{Ummar Abbas}
\author{Mohammad Shahmeer Ahmad}
\author{Firoj Alam}
\author{Enes Altinisik}
\author{Ehsannedin Asgari}
\author{Yazan Boshmaf}
\author{Sabri Boughorbel}
\author{Sanjay Chawla}
\author{Shammur Chowdhury}
\author{Fahim Dalvi}
\author{Kareem Darwish}
\author{Nadir Durrani}
\author{Mohamed Elfeky}
\author{Ahmed Elmagarmid}
\author{Mohamed	Eltabakh\thanks{Corresponding author: meltabakh@hbku.edu.qa}}
\author{Masoomali Fatehkia}
\author{Anastasios Fragkopoulos}
\author{Maram Hasanain}
\author{Majd Hawasly}
\author{Mus'ab Husaini}
\author{Soon-Gyo Jung}
\author{Ji Kim Lucas}
\author{Walid Magdy}
\author{Safa Messaoud}
\author{Abubakr	Mohamed}
\author{Tasnim Mohiuddin}
\author{Basel Mousi}
\author{Hamdy Mubarak}
\author{Ahmad Musleh}
\author{Zan	Naeem}
\author{Mourad Ouzzani}
\author{Dorde Popovic}
\author{Amin Sadeghi}
\author{Husrev Taha	Sencar}
\author{Mohammed Shinoy}
\author{Omar Sinan}
\author{Yifan Zhang}
\author{Ahmed Ali\footnote{work done while at QCRI}}
\author{Yassine El Kheir$^{\ddag}$}
\author{Xiaosong Ma$^{\ddag}$}
\author{Chaoyi Ruan$^{\ddag}$}

\affil{Qatar Computing Research Institute (QCRI), \\Hamad Bin Khalifa University }

\date{}
\begin{document}
\settransfont{\transliterationfont}

\maketitle
\newpage
\begin{abstract}
\begin{center}
\textbf{\abstractname} \\[2ex]
\end{center}

We present Fanar, a platform for  Arabic-centric multimodal generative AI systems, that supports 
language, speech and image generation tasks. At the heart of Fanar are
\FS\ and \FP, two highly capable Arabic Large Language Models (LLMs) that are best
in the class on well established benchmarks for similar sized models. \FS\ is a 7B (billion) parameter model that  was trained from
scratch on nearly 1 trillion clean  and deduplicated Arabic, English and Code tokens. \FP\ is a  9B parameter model  continually trained on the Gemma-2 9B  base model on the same 1 trillion token set. Both models are concurrently deployed and  designed to address different types of prompts transparently routed through a custom-built orchestrator. The Fanar platform provides many other capabilities including  a customized 
Islamic Retrieval Augmented Generation (RAG) system for handling religious prompts, a Recency RAG for summarizing information about current or recent events that have occurred after the pre-training data cut-off date. The platform provides additional cognitive capabilities including in-house bilingual speech recognition  
that supports multiple Arabic dialects, voice and image generation that is fine-tuned
to better reflect regional characteristics. Finally, Fanar provides an attribution service that can be used
to verify the authenticity of fact based generated content. 

The design, development, and implementation of Fanar was entirely undertaken at  Hamad Bin Khalifa University's Qatar Computing Research Institute (QCRI) and was sponsored by Qatar's Ministry of Communications and Information Technology to enable sovereign AI technology development.
\end{abstract}

\newpage 

\tableofcontents
\newpage
\section{Introduction}
\label{sec:intro}
\fullwidthbox{
A case for Arabic-centric Large Language Models is made including the limitations of obtaining Arabic data
and the distinctive characteristics of the language.
}

Large Language Models (LLMs) and Generative AI are becoming an integral part of day-to-day activities at the personal and enterprise level due to  their ability to carry out multifaceted language and cognitive tasks. Diverse applications, including writing assistants, translation services, customer support, software development and image generation, are proliferating and being offered as human productivity enhancing tools. While failure cases of LLMs tend to become viral, continued growth in their adoption is an indicator of their practical utility. From a computer science and AI research perspective, LLMs and their multimodal extensions have opened up scientific challenges that will continue to drive innovation in the research community. An important practical challenge that is far from being overcome  is the design and engineering of  high-quality and effective LLMs for non-English languages.  
The major bottleneck for non-English languages is the limited availability of large datasets which are currently necessary to match the performance of English-centric models. As is well known, the most ubiquitous language on 
the Web, from where most data is harvested, is English. The latest statistics from Common Crawl, a non-profit organization that takes a complete snapshot of the Web approximately once a month, show 
that English documents constitute 46\% of all textual web content, while other languages cap at around 6\%. Arabic, the spoken language of more than 400 million people and the official language of over 20 countries, constitutes around 0.5\% of web data\footnote{\url{https://commoncrawl.github.io/cc-crawl-statistics/plots/languages.html}}~\citep{Common-Crawl-open-web-scale-crawl}. Besides lack of data, another big hurdle is the cost of building an LLM, particularly from scratch, in terms of
both the required hardware and technical expertise. The geopolitical environment is trending towards a polarized world where access to high-end GPUs is often constrained and even when hardware is available, the monetary cost of training even a moderate size LLM in a reasonable amount of time can be prohibitive and out of reach for many organizations. Finally, substantial scientific and engineering expertise is required to undertake an LLM building exercise. While many organizations that have built private and open source LLMs release technical reports about their experience, in many cases, technical details are often left out making it difficult to replicate the process of building an LLM from scratch. A significant amount of deep knowledge across the entire 
LLM stack, ranging from data collection and cleaning, to pre-training, post-training, and deep computer systems knowledge is therefore required.

In this work, we introduce Fanar (meaning lighthouse in Arabic), an Arabic-centric Generative AI platform that  includes text-based LLMs,  speech and image generation systems, specialized Retrieval Augmented Generation (RAG) modules, and an attribution service to authenticate and correct facts in generated text. At the center of Fanar are  \FS\ and \FP\ two 7B and 9B parameter LLMs respectively that are trained on nearly 1 trillion clean and deduplicated  Arabic, English and Code tokens. \FS\ was trained from scratch while \FP\ was continually pre-trained on the Gemma-2 9B model on the same 1 trillion data set. The two models work in concert and make up for the lack of Arabic data, especially in technical domains. Additionally, Fanar includes a new customized morphologically
aligned Arabic tokenizer as well as benchmarks that evaluate cultural capabilities that will be of independent interest to the Generative AI research community. 

The rest of this report is structured as follows. In Section~\ref{sec:arabic}, we
give a brief overview of the Arabic language, its footprint, dialects and setup the
socio-linguistic context. A high level overview of the major
components of the Fanar platform and how they interact with each other is provided in Section~\ref{sec:fservices}. The efforts to collect, clean and integrate   a large Arabic data set of nearly 1 trillion tokens are detailed in Section~\ref{sec:arabic_data_curation}, followed by the design of our specialized Arabic tokenizer in Section~\ref{subsec:data:tokenization}. The  model architectures of both \FS\ and \FP\ and the pre-training pipeline  are the subject of Section~\ref{sec:modeling}. A comprehensive overview of post-training steps
and the necessary adaptations for Arabic are then provided in Section~\ref{sec:pt}. The performance
of both models  on standardized and new culturally aware benchmarks  is described in Section~\ref{sec:benchmark}. 
Arabic speech services and the regionally representative image generation capabilities of Fanar
are introduced in Section~\ref{sec:multimodal}. The Retrieval Augmented Generation (RAG) systems for
Islamic content, recent events and important biographies for fact-related response
attribution are the focus of Section~\ref{sec:rags}. Section~\ref{sec:future} concludes with a reflection on
building a large scale GenAI system and a discussion on a future road map for Fanar.

\section{Arabic Language}
\label{sec:arabic}
\fullwidthbox{
An overview of the Arabic language and its distinctive characteristics is provided, motivating the need for developing language technologies specifically tailored for Arabic.
}

The Arabic language, a member of the Semitic language family, is spoken by over 310 million people across the Middle East, North Africa, and the Arabian Peninsula (MENA region) \citep{10.1111/lnc3.12202}, and by more than 467 million individuals across 60 countries worldwide \citep{gregory2021advancing}. Beyond its geographic and linguistic reach, Arabic carries immense spiritual significance as the liturgical language of over 2  billion Muslims who engage in daily prayers and religious practices in Arabic. It is the official or co-official language of 25 countries and holds a prominent place in linguistic and cultural studies due to its complexity and global significance \citep{gregory2021advancing}. Arabic exists as a spectrum of linguistic forms, ranging from Classical Arabic, which is used primarily in religious and classical literary texts, %
and Modern Standard Arabic (MSA), used in formal settings and communications, to a wide array of colloquial dialects spoken across the MENA region and widely used on social media platforms. The rich diversity and significance of Arabic, its complex grammar and structure, and  the colloquial-nuances make it both a fascinating and challenging for language technology development \citep{farghaly2009arabic,habash2010introduction}. %

At the core of Arabic's linguistic system lies a derivational morphology system where words are typically derived from roots that are fit into morphological templates and accept prefixes and suffixes.  %
Roots, typically composed of three consonants (though occasionally four or five), encode fundamental semantic %
meanings. These roots combine with specific morphological patterns to produce words that convey nuanced meaning. 

For example, the root \textit{k-t-b} generates words such as \textit{kitAb} (‘book’), \textit{maktabap} (‘library’), \textit{kutub} (‘books’), \textit{kAtib} (‘writer’), \textit{maktwb} (‘written’), and many other nouns and adjectives, as well as verb conjugations that account for tense, gender, and person\footnote{Buckwalter encoding is used to represent Arabic text.}~\citep{watson2002phonology}. This non-linear system and Arabic's intricate inflectional paradigms, including the use of prefixes and suffixes that mark tense, mood, gender, and number and attaches determiners and pronouns, distinguishes it markedly from Indo-European languages. While Arabic derives a vast array of words from a limited set of roots, Indo-European languages such as English typically rely on concatenative morphology and a more extensive lexicon to achieve similar flexibility~\citep{ferguson1959diglossia,watson2002phonology,10.4236/iim.2017.92003,10.24093/awejtls/vol6no3.9}. Arabic morphology allows for words such as \RL{وَبِكِتَابِهِمْ} \textit{wabikitAbihim} (‘and with their book’) and \RL{فَأَسْقَيْنَاكُمُوهُ} \textit{fa’asqaynākumūhu} (‘And We gave it (water) to you to drink.’).

The sociolinguistic phenomenon of diglossia further complicates the computational processing of Arabic. Modern Standard Arabic (MSA), derived from Classical Arabic, serves as the formal language for media, education, and political discourse. However, it is not the native language of any Arab speaker. Instead, native speakers use regional dialects, which differ significantly from MSA in phonology, syntax, and lexicon~\citep{ferguson1959diglossia}. These dialects are influenced by other languages, such as English, Berber, French, Persian, Turkish, and Aramaic, and can often be mutually unintelligible~\citep{farghaly2009arabic,al2017dialects}.

Another defining feature of Arabic is its script, which is derived from the Nabataean alphabet~\citep{healey2012brief}. The script, similar to other Semitic scripts, is written from right to left, and its adaptability allows it to represent unrelated languages such as Persian and Urdu. However, in most languages that use the Arabic script, short vowels are not explicitly represented, necessitating diacritization for accurate interpretation. This introduces challenges for NLP tasks, particularly those requiring diacritization or semantic disambiguation~\citep{10.20961/prasasti.v6i1.43127,10.3115/v1/w14-3601}.

Although Arabic has many speakers and a profound cultural impact, the amount of Arabic content on the web accounts for only $0.5\%$ of all online data, which presents significant challenges for technological development (Figure~\ref{fig:panel}). The linguistic characteristics, cultural significance, and remarkable diversity of Arabic underscore the pressing need to develop language technologies specifically tailored to this language. Integrating linguistic expertise and anthropological insights into the development and evaluation processes is essential for effectively addressing the unique complexities inherent in Arabic. By doing so, we can ensure that these technologies not only meet the technical challenges of the language but also honor its rich cultural heritage and nuanced variations.

\section{Fanar Platform Services}
\label{sec:fservices}
\fullwidthbox{
Fanar platform services are introduced including \FS\ and
\FP, speech and image generation capabilities and RAG features
}

\begin{figure}[ht!]
    \centering
  
    \includegraphics[width=0.9\textwidth]{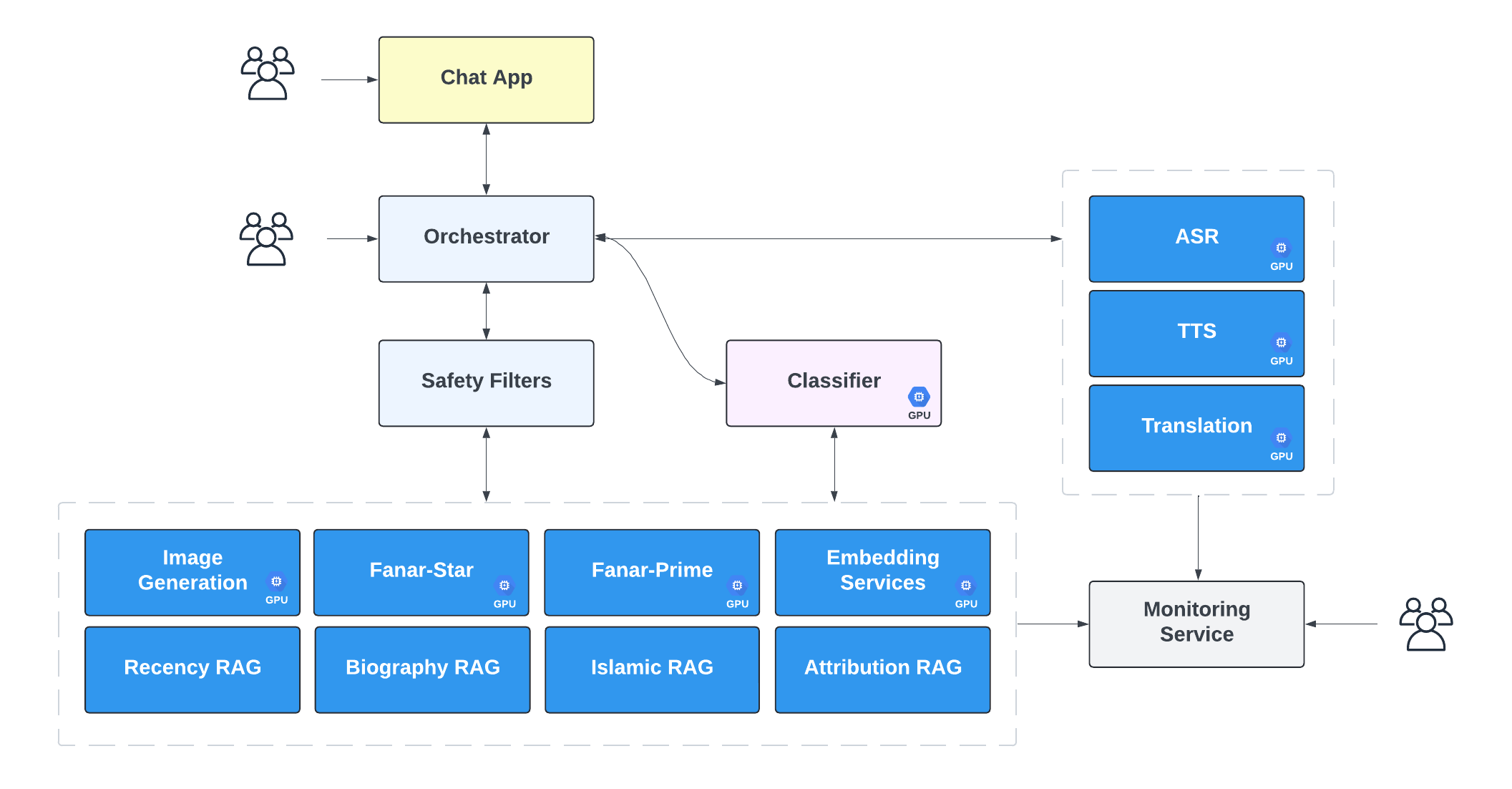}
    \caption{The GenAI services provided through the Fanar platform. The Orchestrator is responsible for routing prompts to the appropriate 
    services depending upon the nature of the request. All responses
    are safety-checked to adhere to responsible AI guidelines and
    cultural alignment.}
    \label{fig:fo}
\end{figure}

The Fanar platform is organized into a set of services coordinated through an \bt{Orchestrator} as shown in Figure~\ref{fig:fo}. 
Requests from the Chat App or an API are either sent directly to speech and translation services or passed through a safety filter and then classified to be processed by other LLM-driven services. Prompts for image generation are also first passed through a safety filter.
We briefly summarize the core services that make up the Fanar family. More details will be given in subsequent sections.

\noindent
{\FS}: This is the flagship 7 billion-parameter LLM trained entirely from scratch using a meticulously designed two-stage curriculum approach. This model leverages a refined implementation of the decoder-only Transformer architecture \citep{NIPS2017_3f5ee243}, inspired by the architectural principles of OLMo \citep{groeneveld2024olmo} and the LLaMA family \citep{touvron2023llama2}. The pre-training process begins with a multi-epoch phase comprising two initial epochs over a diverse corpus of 1 trillion tokens, distributed across Arabic (40\%), English (50\%), and programming code (10\%). In subsequent epochs, the token count is reduced to 0.8 trillion through targeted filtering using an education classifier, with an increased focus on Arabic content (50\%), accompanied by proportionate adjustments for English (40\%) and code (10\%). The pre-training concludes with a carefully designed cool-down stage, incorporating an additional 100 billion tokens from high-quality datasets curated in-house and gradually diminishing the learning rate to zero. Detailed descriptions of the model architecture and pre-training recipe are provided in Section~\ref{sec:modeling}. \FS\ undergoes a comprehensive post-training stage consisting of Supervised Fine-Tuning (SFT) and Direct Preference Optimization (DPO) \citep{rafailov2024direct} to achieve robust alignment with safety and ethical considerations, as detailed in Section~\ref{sec:pt}. The deployment architecture incorporates an orchestration mechanism whereby prompts other than Islamic or STEM domains, as determined by a specialized classifier, are routed to \FS.

\noindent
{\FP}: This model uses continual pre-training to build upon the Gemma-2-9B base model \citep{Riviere2024Gemma2I}, which was itself initially pre-trained on 8 trillion tokens using knowledge distillation from a larger model. Our approach begins with strategic vocabulary pruning, reducing the original 250,000-token vocabulary to 128,256 tokens to optimize compatibility with our training data. The model comprises 459 million embedding parameters and 8.32 billion non-embedding parameters, totaling 8.78 billion parameters. The continual pre-training process for \FP\ mirrors the two-stage curriculum strategy employed for \FS\ but is limited to a single epoch, followed by a cool-down stage. The data mixture and filtering criteria are aligned with those of \FS\, ensuring a balanced representation of Arabic, English, and code content. Post-training of \FP\ aligns closely with the methodology applied to \FS\, including SFT and DPO for safety and value alignment. During orchestration, \FP\ is designated to handle STEM and reasoning-related prompts, routed through the specialized classifier.

\noindent
\textbf{Speech Recognition (SR):} Fanar enables natural interaction through speech -- the most effortless and natural form of human communication. 
The orchestrator integrates a state-of-the-art Arabic-English Automatic Speech Recognition (ASR) system. 
The advanced ASR system supports -- \textit{(i)} multiple Arabic dialects, e.g., Egyptian, Gulf and Levantine; \textit{(ii)} non-native Arabic accents; and \textit{(iii)} diverse code-switching scenarios, including both dialectal variations within Arabic (e.g., Modern Standard Arabic (MSA) $\leftrightarrow$ Egyptian dialect (EGY)) as well as seamless transitions between English and Arabic (Ar $\leftrightarrow$ En). 
These capabilities collectively empower Fanar to accommodate  dialectal Arabic speakers and  foster inclusivity for non-native Arabic speakers. Details of ASR are provided in Section~\ref{sec:multimodal}.

\noindent
\textbf{Text-to-Speech (TTS):} To enable better accessibility, the Platform integrates Arabic and English text-to-speech systems. The TTS systems leverage Diffusion Transformer with ConvNeXt V2 \citep{chen-etal-2024-f5tts} for better text-speech alignment during in-context learning, without the extra modules like grapheme/phoneme alignment, duration predictor, text encoder, or any aid of codec for semantic information infusion. For details, see Section \ref{sec:multimodal}.

\noindent
\textbf{Image Generation (IG):} The Platform provides support for image generation that is  aligned for reflecting Arab and Islamic preferences. The Stable Cascade model is used as it has a much smaller latent
space compared to the well known Stable Diffusion model and its variants and is optimized for both faster fine tuning and inference. In the current landscape of state-of-the-art (SOTA) image models, such as Stable Cascade, biases are evident when generating images from neutral prompts. These models predominantly depict elements of Western cultures, including people, cuisine, and scenery. Additionally, there is a notable lack of accurate representation when generating images related to Middle Eastern topics. This includes details such as culturally appropriate attire, diverse skin tones, and iconic regional landmarks. Our approach for fine-tuning image generation
to reflect local  cultural values is provided in Section~\ref{sec:multimodal} along with concrete examples.

\noindent
\textbf{Retrieval Augmented Generation (RAG)}: A RAG system retrieves relevant information from external data sources for a given input prompt which can then be passed as contextual information to the LLM~\citep{lewis_retrieval-augmented_2020}. By grounding the generated response on the provided context, it can help improve the accuracy of generated responses~\citep{ram_-context_2023}.
Fanar currently provides four RAG systems for controlled
content generation in specific domains.
These are: (i) Attribution RAG for providing supporting evidence (references) for fact-related queries. For example, if the prompt is ``What is the length of the river Nile?" the response will 
be validated against Wikipedia and corrected if there is a mismatch; (ii) Recency RAG for information that is post the checkpoint date
of the pre-training corpus. As an example, for the prompt ``What is the latest weather in Doha?" the system will extract information from
selected verified websites and summarize the information; (iii) Islamic RAG provides content from authoritative websites for Islam related prompts; and finally (iv) Biography RAG for ensuring that accurate information is generated for
well known people in the region and beyond. More details about these four RAG systems are provided in Section~\ref{sec:rags}.

\noindent
\textbf{Translation:}  Fanar provides a specialized service for translation from Modern Standard
Arabic (MSA) to other Arabic dialects and directly from English to the dialects. As parallel data
in this space is highly limited, we fine-tune an existing sequence-to-sequence transformer model
for dialectal translations. Benchmarking details are provided in Section~\ref{subsubsec:data:en-to-msa}. The translation systems 
build upon 15 years of expertise within QCRI in MSA and dialectal Arabic translations.

\section{Pre-training Data }
\label{sec:arabic_data_curation}
\fullwidthbox{
Pre-training data composition for Arabic, English and code is presented. Data filtering pipeline, especially for Arabic is described. The role of machine translation to expand coverage
of Arabic data 
is highlighted.
}

Data is a critical building block for modern AI systems in general and for LLMs in particular. We describe the pre-training data composition for both English and Arabic and the use of machine translation to augment both MSA and dialectal Arabic data.   Given the scarcity of Arabic data, our syntactic and semantic filtering and cleaning approaches are more nuanced compared to  English. 

\subsection{Composition of the Fanar Pre-training Data}

\begin{figure}[hbt!] %
\centering
    \includegraphics[width=0.9\linewidth]{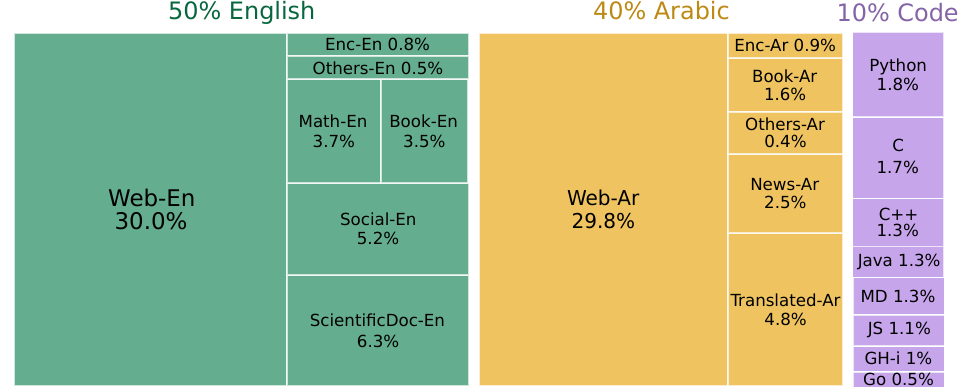}
\caption{Composition of the Fanar pre-training data sources distributions.}
\label{fig:data-mix-fanar}
\end{figure}

\begin{figure}[hbt!] %
  \begin{minipage}[b]{.45\linewidth}
    \centering
    \begin{tabular}{@{}l| l | c @{}}
    \toprule
    \textbf{Domain} & \textbf{Doc Source} & \textbf{Fanar Tokens} \\
    & & (in Billions) \\ 
    \midrule
    \multirow{7}{4em}{English} & Web & 307.8 \\
    & Scientific & 64.6 \\
    & Social Media & 53.4 \\
    & Math & 37.9 \\
    & Books & 35.9 \\
    & Encyclopedic & 8.2 \\
    & Others & 5.1 \\
    \midrule
    \multirow{6}{4em}{Arabic} & Web & 305.7 \\
    & Translated & 49.3 \\
    & News & 25.6 \\
    & Books & 16.4 \\
    & Encyclopedic &  4.1 \\
    & Others & 9.2 \\
    \midrule
    Code & GitHub & 102.6 \\ 
    \bottomrule
    \end{tabular}
    \captionof{table}{Fanar pre-training data includes $\sim$1T cleaned tokens in Arabic, English, and Code, sourced from diverse origins.}
    \label{tab:data-mix-fanar}
  \end{minipage}
  \hfill
  \begin{minipage}[b]{.45\linewidth}
    \centering
    \includegraphics[width=1\linewidth]{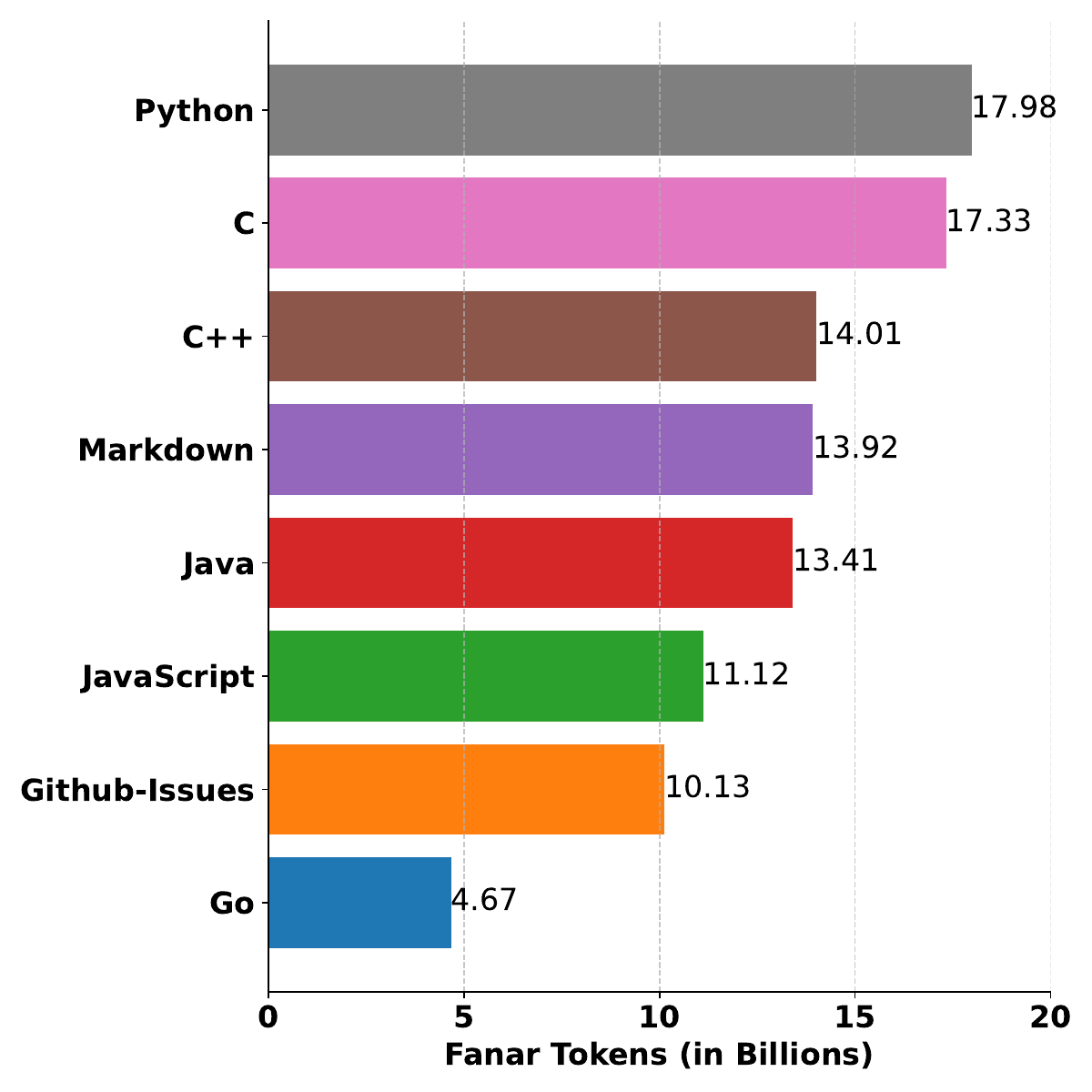}
    \captionof{figure}{Code data composition in our pre-training data.}
    \label{fig:prog-langs}
  \end{minipage}

\end{figure}

To pre-train Fanar, we curated a dataset comprising of 1 trillion tokens spanning Arabic, English, and computer code. The tokens were sourced from  diverse  origins, including web documents, scientific articles, encyclopedic entries, mathematical problems, books, news articles, and source code from common programming languages. The diversity of data is instrumental in enabling the model to exhibit robust performance across a wide array of tasks. Detailed distribution of the data sources for each domain are presented in Figure \ref{fig:data-mix-fanar}, while Table \ref{tab:data-mix-fanar} outlines the token counts for these sources. Recognizing that different corpora contribute uniquely to model training, we aimed to balance corpora that enhance language understanding, facilitate knowledge reasoning, and improve task-specific performance. We employed rigorous preprocessing, including data cleaning, filtering, and deduplication, to ensure the quality of the data. The final data mixture was determined based on extensive ablation studies and are described in Section~\ref{subsec:data:tokenization}.

\subsubsection{English Data Composition}

The English component of our pre-training dataset encompasses approximately 513 billion tokens, derived from a diverse range of sources many of which have been utilized in other representative LLMs.  These include:
(i) web documents from preprocessed Common Crawl sources, including C4 \citep{10.5555/3455716.3455856}, RefinedWeb \citep{penedo2023the},  DCLM \citep{li2024datacomplmsearchgenerationtraining}, Dolma \citep{soldaini-etal-2024-dolma} and RedPajama \citep{weber2024redpajamaopendatasettraining} ensuring broad web-based content representation, (ii) scientific documents derived from RedPajama-arXiv and PeS2o \citep{peS2o} datasets, (iii) social media data extracted from Pushshift Reddit dataset \citep{baumgartner2020pushshiftredditdataset} to capture conversational and informal language patterns, (iv) mathematical data from sources like Algebraic-Stack  \citep{azerbayev2024llemma} and Open-Web-Math 
 \citep{paster2024openwebmath} datasets to capture complex reasoning and mathematical language, (v) books from public domain book data from Project Gutenberg via Dolma corpus and (vi) 
encyclopedic content  from Wikipedia dumps and MegaWika \citep{barham2023megawikamillionsreportssources} for structured, factual information. This rich and varied collection ensures a comprehensive representation of English-language data across multiple disciplines and styles.

\subsubsection{Arabic Data Composition}
Recognizing the limited availability of high-quality Arabic pre-training data, we curated an extensive set of 410 billion Arabic tokens. The data spans multiple varieties of Arabic, including MSA, Classical Arabic, and dialectal content encompassing a diverse range of sources: (i) web documents that were crawled in-house and preprocessed for quality, (ii) news articles and books spanning literature, religion, politics, culture
and history, (iii) encyclopedic content from Arabic Wikipedia (iv) classical and contemporary Arabic poetry and (v) in-house machine
translated books, STEM papers and encyclopedic documents to ensure English-Arabic language alignment.

\subsubsection{Code Data Composition}

Our code data subset comprises approximately 102 billion tokens, representing around 10\% of the pre-training dataset. This subset was sourced primarily from The Stack \citep{kocetkov2023the}, a collection of permissively licensed GitHub projects across a large number of programming languages. We sub-selected data from common programming languages in The Stack, including Python, C, C++, Java, Go, and JavaScript. Additionally, we included Markdown and GitHub Issues data to provide contextual code understanding. Figure \ref{fig:prog-langs} presents the detailed code data composition.

\subsection{Data Curation, Cleaning, and Standardization}
The Arabic data (both web and curated) is collected from a multitude of sources, each having a different file format (e.g., txt, HTML, XML, JSON, zip) and text granularity (e.g., lines, paragraphs, articles, complete books, poems). We homogenize these varying formats into the format used by datasets such as Dolma~\citep{soldaini-etal-2024-dolma}. Files are collections of JSON records \footnote{For example, an article is considered as one record.} with the following fields: \textbf{{\em ``id''}} which serves as a unique identifier; \textbf{{\em ``text''}} which contains the \textbf{core} text content at the granularity of the original data\footnote{Given a website, we automatically detected repeated headers,  footers, and inner sections and links for each source, and removed them from all its articles to improve quality signals.}; \textbf{{\em ``metadata''}} which contains any additional information about the record such as creation date and source URL, and \textbf{{\em ``quality\_signals''}} capturing a set of quality scores collected at the record level, which are described in more detail in Section~\ref{sec:qualitySyn}. 
All text goes through a simple cleaning process where all HTML and JavaScript tags are removed and white spaces (including tabs, newlines, trailing escape characters) are normalized.
\subsection{Data Quality and Filtering}
\label{sec:qualitySyn}
As the data comes from a variety of sources, it contains a lot of noise and low quality content. We do not want to train the model on low quality data as it would adversely affect the quality of generation. As the data does not come with any pre-existing score of quality, we utilize a filtering pipeline that only retains high quality data based on their syntactic and semantic characteristics. This pipeline extends and tailors existing work on data filtering to the Arabic language.
\subsubsection{Syntactic Filtering}
\label{sec:qualitySyn_}
Existing Arabic LLMs, such as \textbf{Jais}~\citep{sengupta2023jais}, apply hard-coded cutoffs based on heuristics to judge the quality of a given text and filter it out if it fails to pass certain thresholds (e.g, special symbols should not exceed 20\% of the content). In our approach, we implemented 20 of the most widely used \textbf{quality signals} described in the \textbf{RedPajama} dataset\footnote{ \url{https://www.together.ai/blog/redpajama-data-v2}}. These heuristic-based quality signals determine the quality of a given text. They cover a variety of measures such as the number of sentences, the number of words, the ratio of symbols and punctuation to words among others. We also removed records with insufficient amounts of Arabic. 

We modified all quality signals to handle Arabic texts properly. Some examples include adding right-to-left punctuation marks and considering digits written in Arabic/Hindi alphabets, diacritics, ligatures,  special symbols, Farsi and decorated characters. %
Other quality signals in RedPajama are handled as part of the deduplication and model-based filtering described in the next subsections.\\

Furthermore, rather than determining the cutoff thresholds for each of these quality signals in an ad-hoc manner, we utilize a systematic approach. For a given quality signal X, we divide X’s score, which is typically between 0 and 1, into 10 histogram ranges with a fixed bin width of 0.1, and then distribute the dataset records based on their scores to one of the 10 buckets. Then, by manually investigating random hundred samples from each bucket, we make more informative decisions on the cutoffs to apply to all datasets. For instance, we observed that setting the threshold for the fraction of unique words in the content to 0.2 effectively identifies a significant portion of advertisement content due to its repetitive nature. \\

\begin{figure}[ht]
\centering
\includegraphics[width=0.4\linewidth]{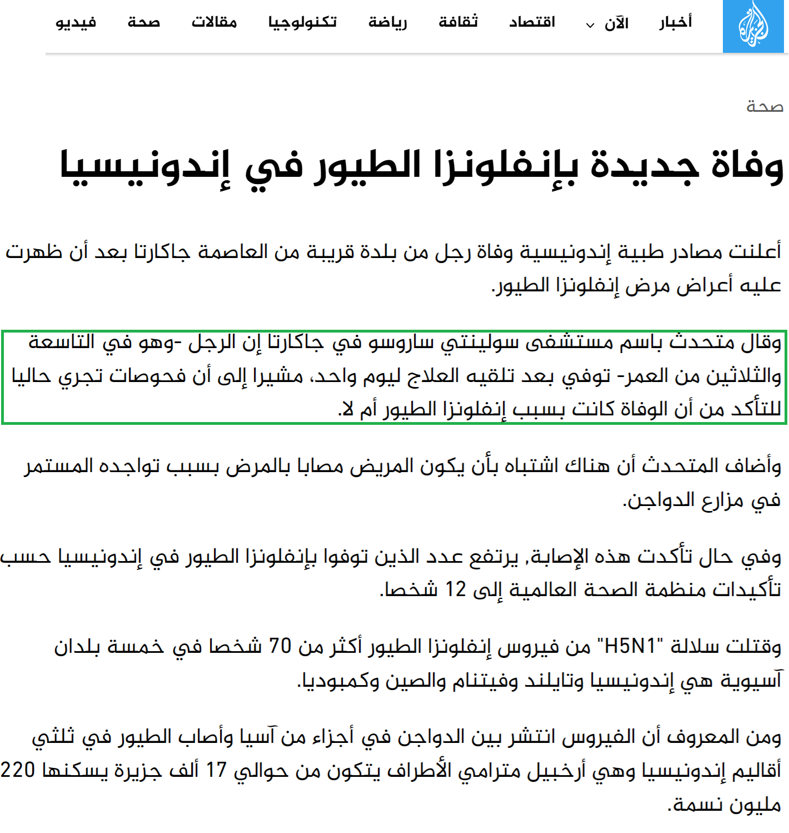}
\caption{High-quality Arabic article failed in a C4 filter. Only one paragraph (not three) has at least 200 characters.}
\label{fig:c4_3_parag}
\end{figure}

Existing web datasets, such as \textbf{C4}~\citep{dodge2021documentinglargewebtextcorpora}, process CommonCrawl data using uniform filtering rules across \textit{all} languages. We hypothesize that while these rules may be suitable for English, they may not be as effective for Arabic. For instance, one such filter excludes web pages containing fewer than three paragraphs, each with a minimum of 200 characters. 
Figure 
\ref{fig:c4_3_parag} illustrates a high-quality Arabic article that would be excluded by the aforementioned filter, as only one paragraph meets the criteria. To address this, we adjusted the filtering rules based on empirical observations from actual Arabic data, as previously described.

\subsubsection{Semantic Filtering}
To remove unwanted content from web data, we used the ASAD system 
\citep{hassan-etal-2021-asad} to detect offensive and profane language, and adult content. We plan to increase the accuracy of hate speech detection in ASAD and use it to filter out hateful content. In addition, we sampled 20 articles from the most common domains from web data and gave them to annotators to estimate the quality and usefulness of each domain.

\subsubsection{Model-based Filtering}
\label{subsec:edu-filter}
Recent progress on data preparation for LLM training~\citep{li2024datacomplmsearchgenerationtraining} has shown the advantage of data filtering using models that predict data quality. While our syntactic and semantic filters removed the majority of noisy data, there is some additional data such as Ads and SEO content that passed the first two filtering phases and require advanced filtering. We explored several approaches in this direction. We introduced two model-based filters that we used for the preparation of pre-training data:
\begin{itemize}
    \item Perplexity filtering: We used KenLM models which are probabilistic n-gram language models for  fast perplexity estimation \citep{heafield2011kenlm}.  These models are trained on Wikipedia content for several languages. For our perplexity filtering, we used a pre-trained model for Modern Standard Arabic (MSA)\footnote{https://huggingface.co/edugp/kenlm}. We assessed the perplexity distributions and defined a threshold per dataset to filter data with the highest ~5\% perplexity. In the last training epoch, an additional perplexity filtering is applied to remove low perplexity content. At that training stage, these documents are becoming too easy and do not give high learning opportunity for the LLM.
    \item Filtering using Education classifier: We build an education classifier following the approach introduced in FineWeb-Edu for English Web data~\citep{lozhkov2024fineweb-edu}. In Fine-Web-edu-ar, a translation for FineWeb-Edu to Arabic is proposed. A small NLLB-500M model is used to translate FineWeb-Edu. The data quality in Fine-Web-edu-ar is poor due to the use of small Machine Translation model. Alternatively, we construct a native Web education Arabic dataset:  First we sampled randomly 1M documents from our web corpus.  We used an Qwen-2.5-73B-Instruct to annotate these documents for the classifier training~\citep{qwen2.5}. The LLM is prompted to assign a score between 0 and 5 reflecting the richness of the document in educational content. We used the same annotation prompt as for FineWeb-Edu. Minor adjustment to the labeling prompt are made to accommodate the Arabic content. In order to train the classifier, we selected a multilingual embedding that is good for Arabic language \citep{bge-m3}. We trained a classification head on top of the embedding to score the education level of web documents. The average accuracy of the classifier on a validation set is about 70\% which is comparable to the results on English FineWeb-Edu classifier. We used the education classifier filter low education score with values 0 and 1. Inspection of the low education content revealed documents several unfiltered Ads and adult content. The filtered data correspond approximately to 20\% of the data.
\end{itemize}

\subsection{Data Deduplication}
\label{sec:qualityDedup}
Data deduplication is important in maintaining and controlling the high quality of the data. As such, the problem of scalable near duplicate detection and its benefit in training LLMs has been explored extensively. \citep{lee2022deduplicatingtrainingdatamakes,broder1997resemblanceofdocuments,bogen2013scalableduplicatedetection,elmagarmid2006duplicate}  
In Fanar, we apply two types of data deduplication, namely {\em exact-match dedup}, and {\em approximate-match dedup}. Additionally, we perform URL deduplication on all web-based data. We implemented a pipeline that can scale well to large datasets under limited compute and memory. We perform both inter-dataset and cross-dataset deduplication. 
\vspace{2mm}

{\textbf{Exact-Match Deduplication:}}
We implement a hashmap based approach to put similar objects in the same bucket. We control the number of buckets such that each one on average is around 10GB to 20GB and peak memory usage does not exceed 1TB. Then, in a parallel fashion, each bucket is processed by sorting its records and eliminating duplicates. We then reconstruct the files in the same order as the original ones using a merge-sort operation utilizing the unique id of each record. 
\vspace{2mm}

{\textbf{Approximate-Match Deduplication:}}
Exact-match deduplication, although computationally efficient, it has its own limitations. For the approximate-match deduplication (also known as fuzzy deduplication), we adopt the same approach used in other datasets~\citep{slimpajama2024, tokpanov2024zyda13tdatasetopen, brown2020languagemodelsfewshotlearners, zeng2021pangualphalargescaleautoregressivepretrained}, which is the min-wise locality sensitive hashing LSH technique~\citep{broder1997resemblanceofdocuments}. 
We experimented with various parameter configurations, and converged to the following parameters, {\em gram size} is set to 8, 
the {\em number of bands (b)} is set to 12, and {\em band length (r)} is set to 11 which results in the approximated {\em Jaccard similarity threshold} being  0.8 and {\em signature length} as 132. The algorithm is highly scalable and we used 350 CPU cores with a peak memory usage of 1TB and completed in about 12 hours.

\subsection{Machine Translation}
\label{subsec:data:diaMT}
Leveraging over a decade of advancements in Arabic NLP, QCRI has consistently set benchmarks in machine translation and resource creation for Arabic and its underrepresented dialects. Our Shaheen machine translation system \citep{sajjad-etal-2017-neural} exemplifies this legacy, offering a state-of-the-art precision in translating between English and Modern Standard Arabic (MSA). This expertise enabled us to tackle the dual challenge of creating robust datasets and training state-of-the-art Fanar LLM addressing both linguistic and cultural nuances. 

Large Language Models (LLMs) have demonstrated impressive capabilities, primarily due to their extensive size and the diversity of the data they are trained on. However, this advantage is heavily skewed towards high-resource languages like English, leaving low-resource languages, such as Arabic, at a significant disadvantage. To bridge this gap, researchers and practitioners have increasingly turned to synthetic data generation methods, with Machine Translation (MT) emerging as a prominent approach. By translating existing English datasets into low-resource languages, MT facilitates the creation of larger and more diverse datasets, thereby enhancing the ability and performance of LLMs for these underrepresented languages. This approach not only expands linguistic resources but also enables access to rich knowledge repositories, including scientific literature, encyclopedias, and other genres that are predominantly available in English, enriching the breadth and depth of training data for LLM.

\subsubsection{English-to-MSA Translation}
\label{subsubsec:data:en-to-msa}

To initiate our exploration of MT systems, we benchmarked several open-source and commercial solutions, including state-of-the-art models such as \textbf{shaheen} \citep{sajjad-etal-2017-neural}, \textbf{mbart} \citep{liu-etal-2020-multilingual-denoising}, \textbf{madlad400} \citep{kudugunta2023madlad400}, \textbf{helsinki} \citep{ostling-etal-2017-helsinki}, \textbf{nllb 3.3B} \citep{nllbteam2022language}, \textbf{gpt4} \citep{openai2024gpt4}, and \textbf{S-T5} based on Ara-T5 \citep{nagoudi2022arat5}.\footnote{The Ara-T5 system does not perform translation out-of-the-box, we fine-tuned it a mixed domain of 515K sentences \citep{joty-etal-2015-avoid}.} These systems were evaluated using AraBench \citep{sajjad-etal-2020-arabench}, which provides a diverse range of test sets spanning multiple domains. The evaluation revealed that different systems excelled in specific domains (results summarized in Table~\ref{tab:en-to-msa-bench-marking}). Consequently, we selected three systems: shaheen, nllb, and S-T5 for translating specialized content across diverse domains.

\begin{table}[htbp]
  \centering
  \resizebox{\textwidth}{!}{
  \begin{tabular}{lcccccccc}
    \toprule
    \textbf{Test Name}                   & \textbf{shaheen} & \textbf{mbart} & \textbf{madlad400} & \textbf{helsinki} & \textbf{nllb} & \textbf{gpt4} & \textbf{S-T5} & \textbf{Category} \\
    \midrule
    bible         & 4.6     & 3     & \textbf{7.9}    & 4.3      & 6    & 6.3  & 5.1    & \multirow{1}{*}{Religious} \\
    iwslt.11               & \textbf{14}      & 10.9  & 12.3   & 11.6     & 12.5 & 10.9 & 12.1   & \multirow{4}{*}{Spoken } \\
    iwslt.12               & 15.7    & 14.4  & 16.4   & 15.4     & \textbf{16.9} & 14.9 & 16     & \\
    iwslt.13               & 16.1    & 15.6  & \textbf{19.5}   & 16.9     & \textbf{19.5} & 16.6 & 18.5   & \\
    iwslt.14               & 13.9    & 13.4  & 15.7   & 13.3     & \textbf{16.5} & 13.1 & 14.9   & \\
    news.04                 & 24.2    & 9.7   & 22.1   & 19.8     & 21.6 & 19.1 & \textbf{32.6}   & \multirow{3}{*}{News} \\
    news.05                 & 25.3    & 7.7   & 21.3   & 18.4     & 21.8 & 17.9 & \textbf{36.6}   & \\
    news.test06                 & 15.5    & 5.7   & 12.8   & 10.1     & 12.5 & 12.4 & \textbf{20.4}   & \\
    ldc\_web\_eg.test  & \textbf{6.3}     & 4.7   & 5.5    & 2.2      & 5.6  & 5.6  & 5.4    & \multirow{2}{*}{General} \\
    summa-AJ      & 20.9    & 10.9  & 19.2   & 18       & 20.8 & 20.8 & \textbf{24.7}   & \\
    summa-BBC     & 19.3    & 13.7  & 20.3   & 16.5     & 20.8 & 20.8 & 21.5   & \\
    qed\_1                 & \textbf{24.7}    & 7.8   & 6.4    & 14.6     & 18.7 & 7.8  & 6.7    & \multirow{1}{*}{Education} \\
    qed\_2                 & \textbf{19.1}    & 9.6   & 6.3    & 11.8     & 17.4 & 11.2 & 8.5    & \\
    travel                 & \textbf{21.1}    & 12.7  & 14     & 13.3     & 16.1 & 17.9 & 14.9   & \multirow{1}{*}{Travel} \\
    mayo                   & 13.6    & 11.9  & 20.4   & 14.5     & 22   & \textbf{20.7} & 12.4   & \multirow{1}{*}{Health} \\
    \midrule
      \textbf{Average} & \textbf{17.0} & 10.1 & 14.7 & 13.4 & \textbf{16.6} & 14.4 & \textbf{16.7} &  \\
    \bottomrule
  \end{tabular}
  }
\caption{Bench-marking English-to-MSA Machine Translation Systems}
\label{tab:en-to-msa-bench-marking}
\end{table}

\begin{table}[ht]
\centering
\begin{tabular}{@{}lrrrrr@{}}
\toprule
                  & \multicolumn{1}{l}{shaheen} & \multicolumn{1}{l}{nllb } & \multicolumn{1}{l}{S-T5} & \multicolumn{1}{l}{No Best System} & \multicolumn{1}{l}{Total Samples} \\ \midrule
Books             & \textbf{53.36\%}            & 50.59\%                       & 32.81\%                       & 11.86\%                            & 253                               \\
STEM Papers       & 28.05\%                     & \textbf{58.54\%}              & 14.63\%                       & 18.29\%                            & 82                                \\
Wiki Encyclopedia & 28.26\%                     & \textbf{68.48\%}              & 31.52\%                       & 10.87\%                            & 92                                \\ \bottomrule
\end{tabular}
\caption{Human evaluation scores for English-to-MSA translation. Annotators were allowed to mark multiple best systems for each sample, or explicitly mark if none of the translations were good. The scores indicate the percentage of samples for which a system was in the best translations list.}
\label{tab:en-to-msa-human-eval}
\end{table}

We chose three specific genres, namely Books, STEM Papers, and Wiki Encyclopedia, for translation from English to MSA. To determine the most suitable system for each genre, we selected a small sample from each category and translated them using our chosen systems. Subsequently, we conducted a human evaluation on this subset to identify the best-performing system for each genre. In addition, we calculated COMET scores \citep{rei-etal-2020-comet} as an additional evaluation metric. The evaluations are presented in Tables \ref{tab:en-to-msa-human-eval} and \ref{tab:en-to-msa-comet-eval}. Based on the outcomes, we opted for \textbf{Shaheen} for translating books\footnote{Note that although the COMET scores favored the \textbf{Nllb} system for all genres, we ultimately selected \textbf{Shaheen} based on human evaluation.

} and the \textbf{Nllb 3.3B} model for translating STEM papers and Wiki Encyclopedia.

\begin{table}[ht]
\centering
\begin{tabular}{@{}lrrr@{}}
\toprule
                  & \multicolumn{1}{l}{shaheen} & \multicolumn{1}{l}{nllb} & \multicolumn{1}{l}{S-T5} \\ \midrule
Books             & 0.7219                      & \textbf{0.7301}                        & 0.6865                        \\
STEM Papers       & 0.6829                      & \textbf{0.7664}                        & 0.6562                        \\
Wiki Encyclopedia & 0.7143                      & \textbf{0.7735}                        & 0.6654                       \\ \bottomrule
\end{tabular}
\caption{COMET evaluation scores for English to MSA translation}
\label{tab:en-to-msa-comet-eval}
\end{table}

\subsubsection{MSA-to-Dialect Translation}
\label{subsec:mt-into-dialects}

Despite its rich diversity, dialectal Arabic remains significantly underrepresented in Large Language Models (LLMs). To address this gap during the development of Fanar, we developed English-to-Dialectal Machine Translation models, complemented by human post-editing, to create robust evaluation benchmarks. These benchmarks facilitate translation between Modern Standard Arabic (MSA) and two major dialects: Egyptian (Egy) and Levantine (Lev) Arabic. By extending MSA-based resources like ArabicMMLU to dialectal contexts, we provide valuable tools for assessing LLMs' comprehension of dialectal Arabic.

 We fine-tuned two machine translation models: \textbf{AraT5} \citep{nagoudi-etal-2022-arat5} and \textbf{NLLB} \citep{nllbteam2022language} and experimented with several variants of these models, with sizes ranging from 600M to 3.3B parameters. In our preliminary experiments, we found the NLLB 3.3B model to surpass AraT5 and its smaller variants, post fine-tuning with dialectal data.  We carried out ablation studies using different data mixtures on the NLLB 3.3B model. We shortlisted three systems per dialect using BLEU scores as our primary criterion and used human evaluation to select the best system for each dialect. Table~\ref{tab:combined-scores} provides a summary of the evaluation results across various dialectal test sets within the community. More details on this  can be found in \citep{mousi2024aradicebenchmarksdialectalcultural}.

\begin{table}[ht!]
\centering
\scalebox{0.9}{%
\begin{tabular}{cccccccccc}
\toprule
 & \multicolumn{4}{c}{\textbf{MSA-to-LEV Models}} & \multicolumn{4}{c}{\textbf{MSA-to-EGY Models}} \\
\cmidrule(lr){2-5} \cmidrule(lr){6-9}
 & \textbf{OSACT} & \textbf{SADID} & \textbf{LDC} & \textbf{D2M} & \textbf{ARZEN} & \textbf{D2M} & \textbf{LDC} & \textbf{MADAR} \\
\midrule
S1 & 9.8 & 12.7 & 6.2 & 11.0 & 1.8 & 57.3 & 11.8 & 17.7 \\
S2 & 9.8 & 11.8 & 6.3 & 11.7 & 17.3 & 57.2 & 12.3 & 17.6 \\
S3 & 9.7 & 11.8 & 7.0 & 47.8 & 15.8 & 55.0 & 11.2 & 17.9 \\
S4 & 5.92 & 8.42 & 3.46 & 4.89 & 1.88 & 7.53 & 2.83 & 6.02 \\
\bottomrule
\end{tabular}
}
\caption{SacreBLEU scores for MSA-to-LEV and MSA-to-EGY models. \textit{S1} to \textit{S4} represent different configurations: \textbf{MSA-to-LEV}: \textit{S1} = UFAL, \textit{S2} = +LDC, MADAR, PADIC, D2M, \textit{S3} = +LDC, MADAR, PADIC, D2M, \textit{S4} = GPT4 zero-shot. \textbf{MSA-to-EGY}: \textit{S1} = MADAR + D2M + LDC, \textit{S2} = MADAR + D2M + LDC + Arzen, \textit{S3} = MADAR + D2M + LDC + Arzen + BOLT, \textit{S4} = GPT4 zero-shot.}
\label{tab:combined-scores}
\end{table}

\section{Tokenization}
\label{subsec:data:tokenization}
\fullwidthbox{
The importance of tokenization in LLMs is explained. The limitations of BPE type
tokenizers for Arabic are explained. A new morphologically aware tokenizer
algorithm is introduced.
}

Tokenization is a foundational step in any natural language processing (NLP) pipeline, segmenting text into tokens such as bytes, characters, subwords, words, or multi-word units. The quality of tokenization directly impacts downstream tasks, as errors can propagate through the pipeline, ultimately degrading the performance of downstream applications \citep{sajjad-etal-2017-challenging, adel2018overview}. Tokenization has a rich history in NLP, with methods ranging from simple whitespace splitting to advanced statistical and neural approaches \citep{smit2014morfessor, otani-etal-2020-pre}. Tokenization plays a critical role in Large Language Models (LLMs), influencing their efficiency, context length, and even their precisions \citep{dagan2024getting}. While tokenization-free approaches also exist in LLM research as an alternative \citep{clark2022canine,deiseroth-etal-2024-free}, most successful models (e.g., Gemma, LLaMA, and the OpenAI GPTs) continue to rely on Byte Pair Encoding (BPE)-based tokenizers, along with its underlying assumptions and limitations.

\subsection{Byte-Pair Encoding and its Limitations}
BPE, originally introduced as a traditional text compression algorithm \citep{shibata1999byte}, was first proposed for use in machine translation in 2016  as a text tokenizer \citep{sennrich-etal-2016-neural}. Since then, it has been widely adopted in NLP and LLMs due to its efficiency in managing vocabulary size, handling out-of-vocabulary words, prioritizing frequent patterns, and, to some extent, improving upon morphology-based tokenizers \citep{sennrich-etal-2016-neural}. Despite its widespread success, Vanilla BPE has notable limitations: (i) its greedy algorithm, (ii) inefficiencies in cross-lingual settings where similar words may use different character variations, and (iii) amount of character information is not equal in different languages. These shortcomings have spurred modifications, such as BPE dropout \citep{provilkov-etal-2020-bpe}, sampling-based BPE~\citep{asgari2019probabilistic,asgari2020subword}, byte-level extensions \citep{wang2020neural}, multilingual BPEs \citep{liang-etal-2023-xlm}. 

\subsection{Challenges of Byte-Pair Encoding and Arabic}

\textbf{Arabic Morphology and BPE:} The additive nature of BPE makes it well-suited for suffixing languages like English. However, languages such as Arabic present unique challenges due to their root-and-pattern morphology, complex derivational systems, and their classification as primarily infixing languages \citep{mcomber1995morpheme, versteegh2014arabic}. Consequently, traditional BPE and byte-level approaches \citep{sengupta2023jais} often fail to effectively capture the intricate morphological structures of Arabic, underscoring the need for more advanced tokenization strategies tailored to the linguistic properties of such languages.

Analyzing the output of vanilla BPE on Arabic text, we observed that Arabic's morphological structure, characterized by its infixes and root-pattern system, is not well-suited to BPE. As a result, tokens are often segmented in a morphologically meaningless manner, introducing unnecessary ambiguity. For instance, the word \RL{الرحمن} (Al-Rahman, ``The Merciful'') is segmented into \RL{من} (min, ``whom'') + \RL{ال} (al, ``the'') + \RL{رح} (rah, an incomplete fragment). Here, \RL{من} (min), a frequent token in Arabic, is semantically unrelated to the original word \RL{الرحمن}. This segmentation forces the model to disambiguate these unrelated components, complicating the learning of meaningful embeddings. On the other hand, purely morphological segmentation in language models has also shown limitations, as it does not align with the frequent patterns present in natural language usage \citep{durrani-etal-2019-one}.

\textbf{Byte-level vs. Char-level Tokenizer:} Since Arabic script characters typically require more than one byte, the use of byte-level BPE (BBPE) is inefficient and demands a larger number of merging steps. Moreover, byte-level patterns fail to preserve character similarity in many cases. Unlike English, where accented characters often share a byte with the base character, Arabic's encoding structure exacerbates this inefficiency. Similarly, adopting character-level tokenization approaches, such as those used by OpenAI, is also inefficient, particularly during text generation, and will limit the context-size.

\subsection{Fanar Morphology-based Tokenizer}

This observation motivated us to design a tokenization approach that combines the strengths of morphological segmentation and the statistical efficiency of byte-pair encoding (BPE), resulting in the Fanar morphologically aware tokenizer, MorphBPE~\footnote{U.S. provisional patent application number: 63/679,403.}. The core idea is to align the BPE algorithm with the morphological structure of Arabic (or other morphologically rich languages) by modifying it to respect morpheme boundaries during the token merging process.

In the MorphBPE algorithm, we start by initializing the vocabulary with individual characters from the text. The training corpus is then segmented using morphological segmentation, ensuring that the structural morphemes are identified before applying any statistical operations. As shown in Algorithm 1, the algorithm iteratively computes byte-pair frequencies and selects the most frequent byte-pair merge candidate. However, unlike standard BPE, MorphBPE includes a modified step (line 5 in Algorithm 1), which ensures that merges do not cross morpheme boundaries. This modification preserves the morphological integrity of the text while enabling statistical efficiency during tokenization.

The iterative process continues until the desired vocabulary size is reached. At each step, the vocabulary is updated to include the newly merged tokens, balancing the morphological structure of the language with statistical considerations for improved representation in language models.

\begin{algorithm} \caption{Morphologically-aware BytePair Encoding (MorphBPE)} \begin{algorithmic}[1] \State Initialize vocabulary with individual characters \State Segment the training corpus using morphological segmentation \While{number of merges $<$ desired vocabulary size} \State Compute byte-pair frequencies \State \textbf{Modified Step:} Find the most frequent byte pair that does not cross morpheme boundaries \State Merge the most frequent byte pair into a new symbol \State Update the vocabulary with the merged symbol \EndWhile \end{algorithmic} \end{algorithm}

By respecting the morphological structure during tokenization, MorphBPE addresses the inefficiencies and ambiguities of standard BPE when applied to morphologically rich languages like Arabic. This approach enhances the ability of language models to learn meaningful embeddings that better capture linguistic nuances.

\subsection{Tokenization Evaluation}

Tokenization evaluation can be conducted using \textbf{intrinsic} or \textbf{extrinsic} metrics. Below, we discuss key intrinsic metrics and their applications:

\noindent\textbf{(i) Fertility:} Fertility measures the ratio of the number of tokens produced by a tokenizer compared to a baseline tokenizer, typically one that uses whitespace splitting. A lower fertility score is often interpreted as indicative of a better tokenizer, as it suggests more efficient representation. However, this argument can be disputed, particularly for agglutinative languages like Turkish, where meaningful representation for large language models (LLMs) requires more tokens to capture the underlying morphological structure and ensure sufficient context for each surface form.

\noindent\textbf{(ii) Perplexity:} Perplexity measures the likelihood of a held-out text for a trained model is another commonly used intrinsic evaluation metric. However, comparing perplexity across different tokenizers is valid only if their vocabulary sizes are the same. Otherwise, differences in vocabulary size render the comparisons non-equivalent. 

\noindent\textbf{(iii) Morphological Alignment Score (Proposed Metric):}
We propose a new intrinsic evaluation metric, the \textbf{Morphological Alignment Score}, which assesses how well the tokenization aligns with the underlying morphological segmentation of words. To calculate this, we use a pairwise alignment score based on dynamic programming, ensuring that the order of matching tokens with segmented morphemes is preserved. This method provides a quantitative measure of how effectively a tokenizer respects the morphological structure of the language.

\subsection{Fanar Tokenizer Preprocessing and Training}
Preprocessing is a crucial step in the BPE process to ensure accurate frequency counts of character patterns. We designed an extensive and peer-reviewed preprocessing pipeline for Arabic script-based languages, including Arabic, Persian, Urdu, and others, to normalize all related scripts into a standard form. As part of this process, we remove diacritics from Arabic text, as most data is not diacritized. However, diacritic characters are still included in the tokenizer vocabulary to allow enhancement during supervised fine-tuning (SFT). To maintain the integrity of the preprocessing, all data has been converted into the HuggingFace format.

The Fanar Tokenizer was trained on the complete Arabic dataset used for model development. By adhering to a vocabulary size that is a multiple of 1024, the tokenizer aligns with modern hardware architectures, such as GPUs and TPUs, which optimize processing in 1024-sized blocks \citep{10.20944/preprints202402.1702.v1}. This alignment improves throughput and reduces latency by enabling efficient token batch processing.

Drawing from LLaMA’s use of a 32K vocabulary size for English and code \citep{10.18653/v1/2023.emnlp-main.534}, we identified optimal morphological alignment at 45K tokens for Arabic. To ensure both efficiency and flexibility, we combined 32K English tokens with 45K Arabic tokens, pruned infrequent merges, and finalized a vocabulary size of \(75 \times 1024 = 76,800\). This size accommodates reserved tokens for multimodality, diacritics, and other special cases.

\subsection{Fanar Tokenizer Evaluation Results}
\begin{figure}[ht!]
    \centering
    \includegraphics[width=\textwidth]{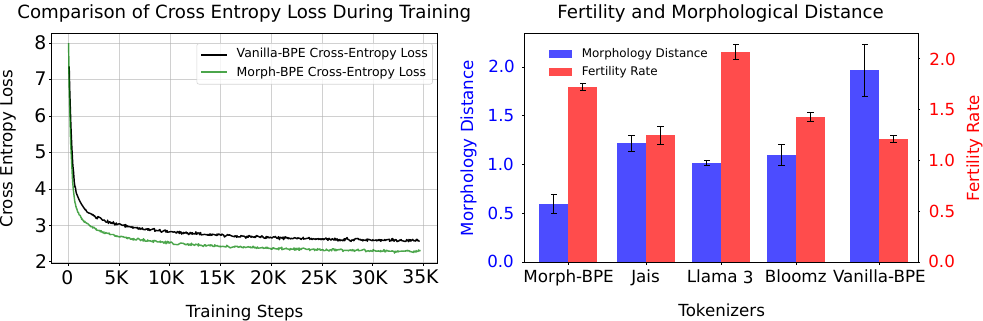}
    \caption{Intrinsic evaluation of Fanar Tokenizer (Morph-BPE). From left to right: \textbf{(i)} Cross-entropy loss comparison between Morph-BPE and vanilla-BPE, Morph-BPE demonstrates faster convergence and smaller loss than vanilla-BPE for the same vocabulary size. \textbf{(ii)} Comparison of fertility and morphological alignment among existing Arabic and Multilingual Tokenizers.}
    \label{fig:fanartokeval}
\end{figure}

The evaluation of the Fanar Tokenizer was conducted across the mentioned metrics, including fertility, morphological distance, and perplexity, comparing vanilla BPE with Morphological BPE. Figure \ref{fig:fanartokeval}-(i) shows that the Fanar Morph Tokenizer achieves a lower training loss compared to vanilla BPE, demonstrating its efficiency and faster convergence. Figure \ref{fig:fanartokeval}-(ii) highlights that the Fanar Morph Tokenizer attains the highest alignment with morphology while maintaining reasonable fertility. These results illustrate that Morphological BPE not only preserves morphological structure but also improves model performance, reducing perplexity loss and accelerating convergence across various model sizes.

\section{Modeling and Pre-training}
\label{sec:modeling}
\fullwidthbox{
The pre-training recipe for both \FS\ and \FP\ is detailed. The multi-epoch and phased approach for language data mixtures
and curriculum training are highlighted.
}

We describe the model architectures of both \FS\ and \FP\, our choice of training stages and our choice of data mixture. We also discuss our ablation studies that informed these choices.

\subsection{Model Architecture}

Both \FS\ and \FP\ are refined versions of the classical decoder-only Transformer architecture. They respectively have 7.1B and 8.78B parameters~\citep{NIPS2017_3f5ee243}. \FS\ reuses the architecture of OLMo \citep{groeneveld2024olmo} and the Llama family \citep{touvron2023llama2} (trained from scratch) while \textbf{\FP}\ is built upon the Gemma-2-9B base model \citep{Riviere2024Gemma2I} (continually trained). Table \ref{tab:arch-fanar} compares the architectures. The vocabulary size of \FS\ is 76,800 while that
of \FP\ is 128,256. The vocabulary of \FP\ is pruned down from the original 250,000. It is also notable that the embedding dimensionality of \FS\ is larger (4096) compared to that of \FP\ (3584).

\begin{table}[hbt!]
\centering
\begin{tabular}{l|c c}
\toprule
 & \textbf{\FS} & \textbf{\FP} \\
\midrule
Num Layers          & 32         & 42        \\
Attention Heads     & 32         & 16        \\
Model Dimension     & 4096       & 3584      \\
Hidden Size         & 4096       & 3584      \\
Intermediate Size   & 11008      & 14336     \\
Pre-Normalization   & \texttt{RMSNorm} & \texttt{RMSNorm} \\
Post-Normalization  & None       & \texttt{RMSNorm} \\
Positional Embeddings & \texttt{RoPE} & \texttt{RoPE} \\
Attention Variant   & \texttt{Full} & \texttt{GQA} \\
Biases              & None       & None      \\
Activation          & \texttt{SwiGLU} & Approximated \texttt{GeGLU} \\
Context Length      & 4096       & 4096      \\
Batch Size (samples) & 1344      & 1071      \\
Batch Size (tokens)  & $\sim$5.5M & $\sim$4.4M \\
Vocab size          & 76,800     & 128,256   \\
Weight Tying        & False      & False     \\
\midrule
Embedding Parameters     & 314M & 459M  \\
Non-embedding Parameters & 6.79B & 8.32B \\
\textbf{Total Parameters} & \textbf{7.1B} & \textbf{8.78B} \\
\bottomrule
\end{tabular}
\caption{Overview of the model configuration of Fanar Star and Fanar Prime.}
\label{tab:arch-fanar}
\end{table}

\subsection{Ablation Studies}
\label{subsec:ablation-fanar}

We ablated key design choices related to data filtering and data mixture composition on 1B-parameter models. We used 50 to 100 Billion tokens to train a model for each configuration of interest. We then benchmarked the resulting models on multiple Arabic-translated tasks, including HellaSwag, OpenBookQA, PIQA, and BoolQ. While the results were consistent across benchmarks, we report statistics on HellaSwag as a representative. We used the best performing design configurations to train our 7B-parameter model. The design choice of the tokenizer was already described in Section~\ref{sec:arabic_data_curation}.

\subsubsection{Comparing Data Filtering Strategies}
We compared two data filtering approaches: (i) Our in-house Fanar filtering recipe (detailed in Section \ref{sec:qualitySyn}) and (ii) The data filtering methodology from the Jais model \citep{sengupta2023jais}.

\begin{wrapfigure}[17]{r}{7.5cm}
   \centering
   \includegraphics[width=0.45\textwidth]{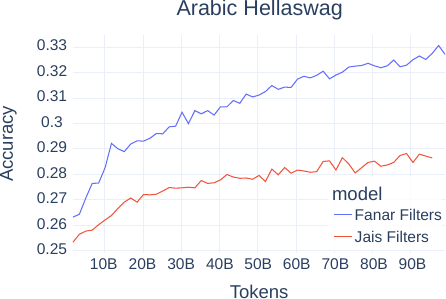}
   \caption{Ablation results on HellaSwag. Fanar and Jais filtering recipes are applied separately on the data and 1B-parameter models are trained on them.}
   \label{fig:ablation-data-filter}
\end{wrapfigure}

\noindent
Figure \ref{fig:ablation-data-filter} compares a few data filtering recipes. The Fanar filtering recipe achieves and improvement of about four points on the Arabic HellaSwag benchmark compared to models trained with the Jais data filtering recipe. Notably, models trained with the Jais-filtered data exhibited earlier performance plateauing, whereas those trained on Fanar-filtered data showed a consistent upward performance trajectory. The improvement in our filtering approach can be attributed to a multi-stage filtering process, including perplexity-based filtering and quality classification. 

\subsubsection{Comparing Data Mixture Composition}
\label{subsec:ablation-datamix}
\begin{figure}[hbt!]
   \centering
   \includegraphics[width=1\textwidth]{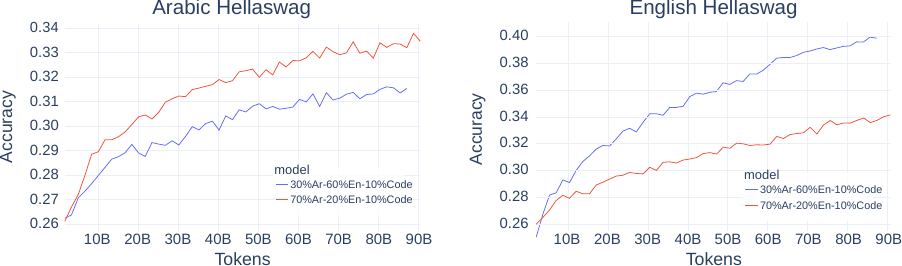}
   \caption{Ablation results on data mixture composition. 1B-parameter models were trained on different data mixtures. 
   }
   \label{fig:ablation-data-mix}
\end{figure}

We investigated different Arabic-to-English token ratios which we present in Figure \ref{fig:ablation-data-mix}.  
The 70:20 Arabic, English ratio (red line) improved Arabic HellaSwag performance
by two points.  However, this adjustment led to a 6-point degradation in English benchmark performance. On the other hand, 
30:60 Arabic, English ratio (blue line) yielded higher English performance but
at a cost to Arabic benchmarks.
Given the disproportionate degradation in English performance compared to the Arabic gains, we adopted a dynamic data-mixing strategy: during the initial phase of pre-training, we kept the English tokens ratio higher and we made progressive adjustments by increasing the Arabic tokens ratio towards the end of the training. 

\noindent

\subsection{{\FS} Pre-training}

\subsubsection{Training Recipe}
Fanar Star was pre-trained using a two-stage curriculum approach specifically designed to address the challenges of limited Arabic language data while maintaining robust multilingual capabilities. Our training recipe comprises a multi-epoch pre-training phase followed by a cool-down phase. This design leverages both breadth and depth in data utilization, ensuring optimal model performance across diverse linguistic and domain-specific tasks. The details of each phase are outlined below.

\begin{itemize}
    \item \textbf{Stage-1: Multi-Epoch Pre-training Phase.} We trained Fanar using for four epochs. This is aligned with findings by \citep{muennighoff2023scaling}, which demonstrate that training data can stay fresh for four epochs. During the first two epochs, the training data comprised a consistent token composition of 40\% Arabic, 50\% English, and 10\% Code. This initial training phase aims to establish a broad cross-lingual foundation.  For the last two epochs, additional filtering was applied using an education classifier (see Section~\ref{subsec:edu-filter} for details), resulting in a 20\% reduction in token volume. Concurrently, we adjusted the data mixture to prioritize Arabic language proficiency by increasing its proportion to 50\%, while reducing the English token share to 40\% and maintaining 10\% for Code tokens. The results provided in this report for Fanar Star are based on completed three epochs of pre-training. The training of the fourth epoch is in progress. 

    \item \textbf{Stage-2: Cool-Down Phase.} Recent studies suggest that training large language models on a subset of high-quality data in the final stages of pre-training significantly enhances their downstream task performance \citep{blakeney2024does, dubey2024llama3herdmodels}. Inspired by this, the cool-down phase of our training recipe involved curating a high-quality data subset, comprising approximately 100 billion tokens from carefully selected Arabic and English sources. In this phase, the model continued training on the curated dataset, with the learning rate linearly annealed to zero, starting from the final learning rate of the multi-epoch pre-training phase. 
\end{itemize}

\begin{figure}[hbt!] %
   \centering
   \includegraphics[width=0.45\textwidth]{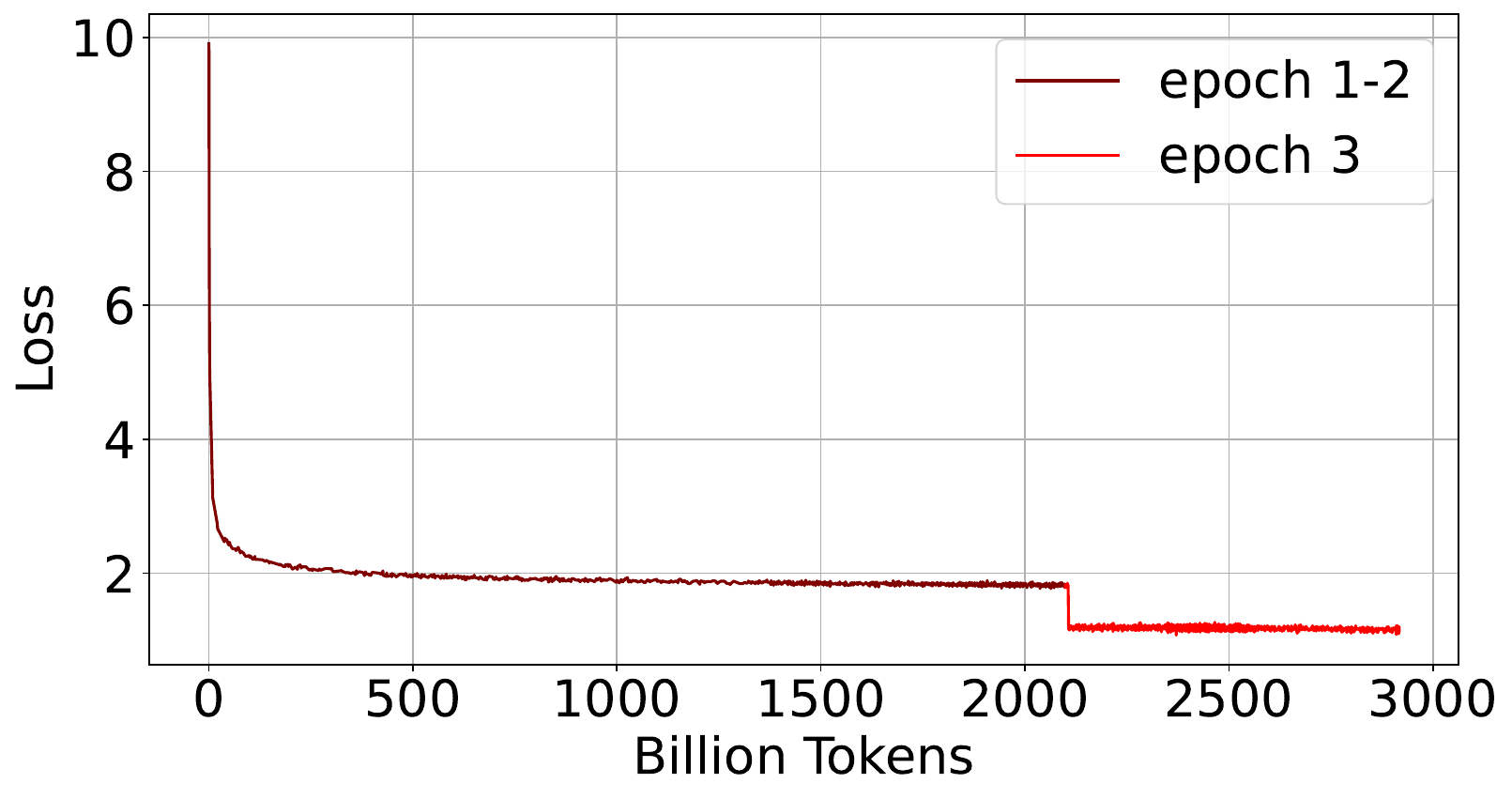} 
   \includegraphics[width=0.45\textwidth]{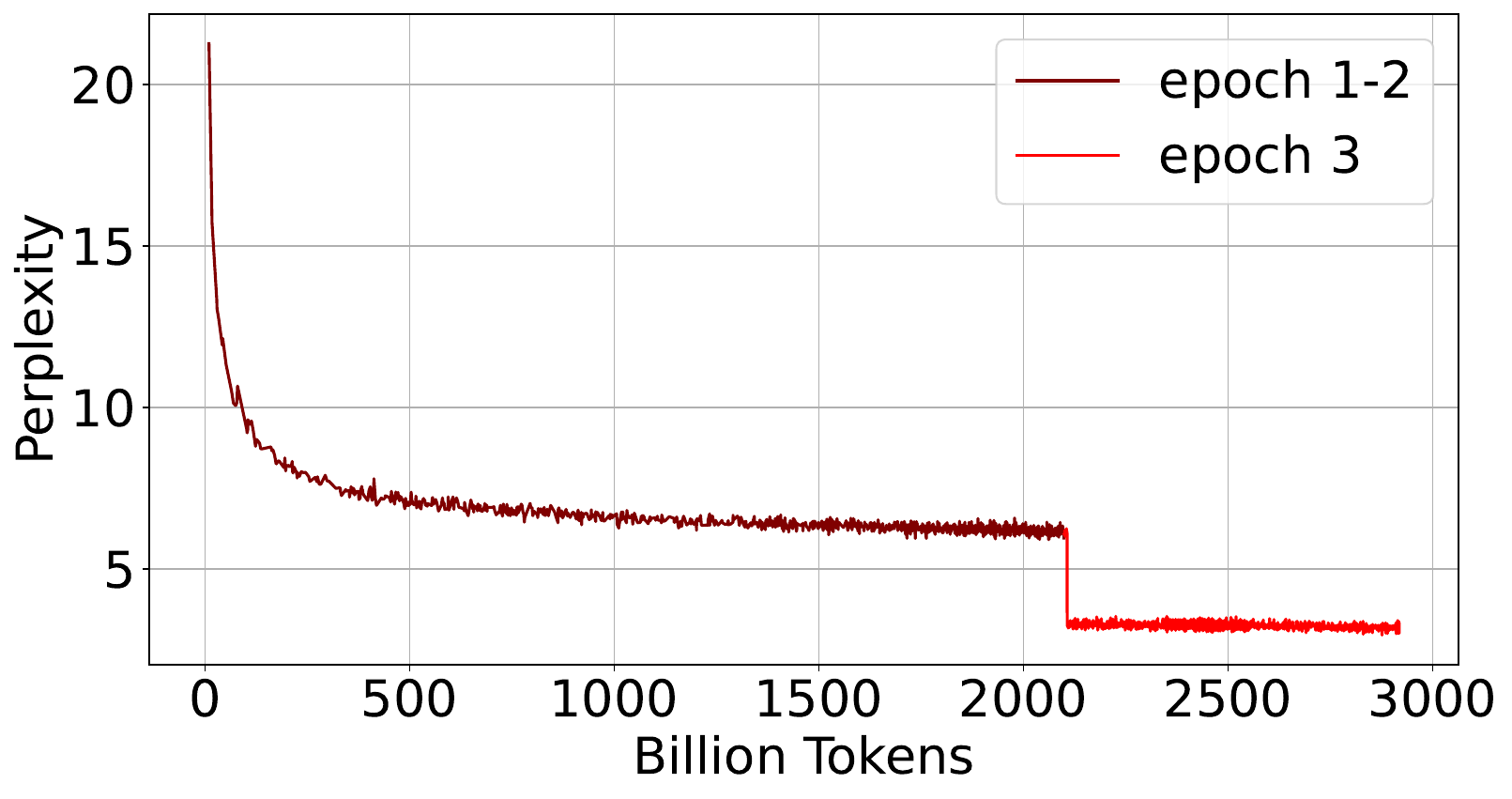}  
   \caption{Training loss and perplexity Curves. The additional filtering applied in epoch 3 is reflected in the reduction of loss and perplexity.}
   \label{fig:train-loss}
\end{figure}

\begin{figure}[hbt!] %
   \centering
   \includegraphics[width=0.45\textwidth]{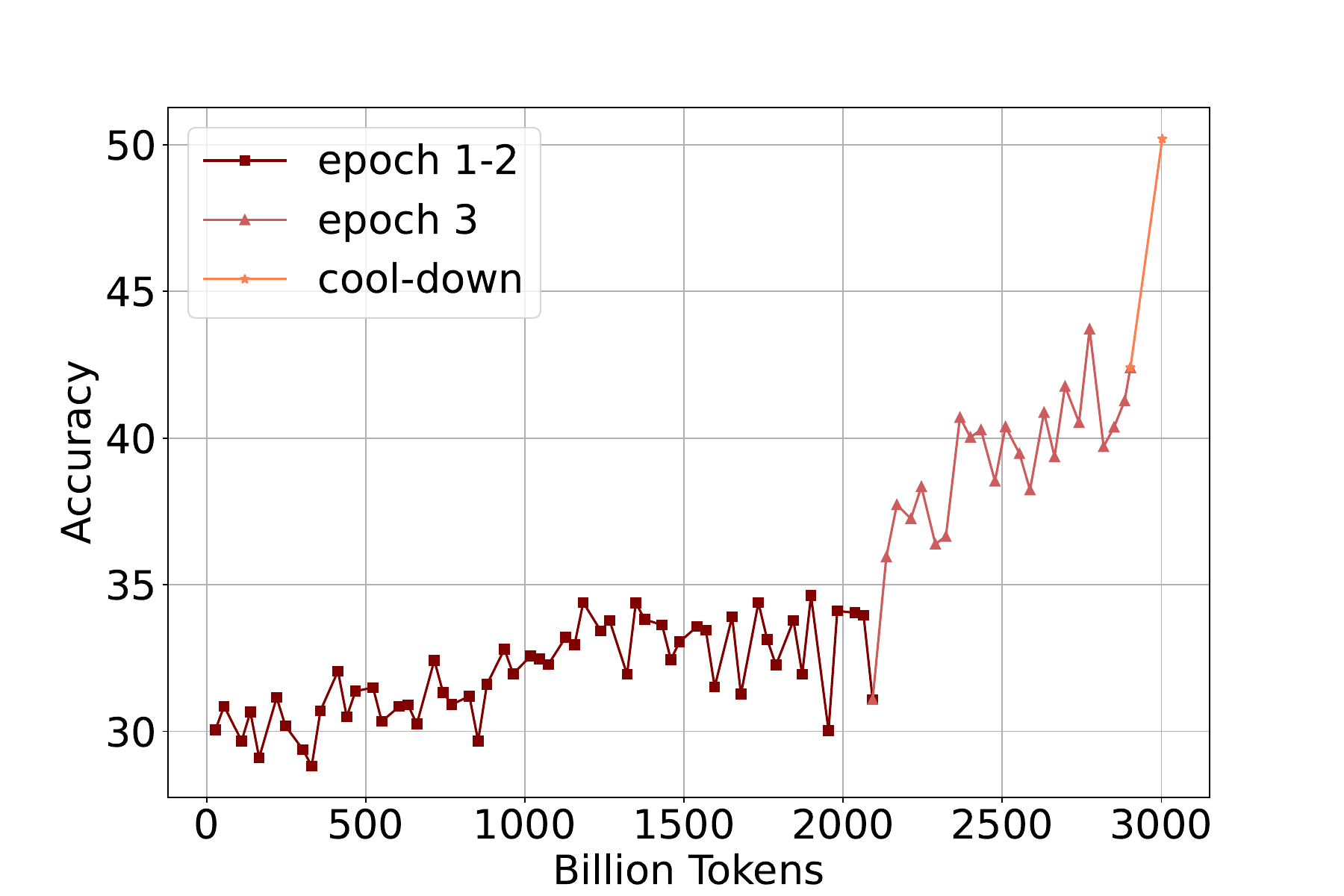} 
   \caption{3-shot accuracy of {\FS} model during pre-training on Arabic MMLU~\citep{koto2024arabicmmlu}. The rate of improvement is noticeable in epoch 3 after applying additional data filtering. Cool-down phase has also given a strong boost in downstream  performance.}
   \label{fig:fs-ar-mmlu}
\end{figure}

Using this recipe, Fanar Star was pre-trained on a total of $\sim$3 trillion tokens. The training loss trajectory of Fanar Star during multi-epoch pre-training stage is presented in Figure~\ref{fig:train-loss}. The loss and perplexity curves decrease consistently which indicates the effective learning throughout the multi-epoch training stage. Furthermore, the model's performance on the Arabic MMLU benchmark, as shown in Figure~\ref{fig:fs-ar-mmlu}, demonstrates progressive improvements across training phases, confirming the efficacy of our pre-training strategy.

\subsubsection{Optimization Configuration}

In both stages, Fanar Star was trained using the standard auto-regressive language modeling objective, with a fixed context length of 4096 tokens. We used AdamW optimizer \citep{loshchilov2018decoupled}. The details of our training configuration is presented in Table~\ref{tab:hyperparam-fanar}. Our pre-training process utilized \texttt{bfloat16} mixed precision to enhance computational efficiency, while critical all-reduce gradient operations were performed in \texttt{fp32} to maintain numerical stability. The global batch size was set to 1344 samples, totaling $\sim$5.5 million tokens per optimization step. The learning rate was managed through a two-phase scheduling approach. Initially, we employed a warm-up stage spanning 2000 steps, during which the learning rate was linearly increased to a maximum value of $3\times10^{-4}$. Following the warm-up phase, a cosine annealing schedule was used to progressively reduce the learning rate to $3\times10^{-5}$ over the course of Stage 1. In Stage 2, the cool-down phase, the learning rate was linearly reduced to zero, effectively training the model on the curated high-quality dataset.

\begin{table}[H]
\centering
\begin{tabular}{@{}l|c c@{}}
\toprule
&  \multicolumn{1}{l}{\textbf{Fanar Star}} & \multicolumn{1}{l}{\textbf{Fanar Prime}}  
\\ \midrule
Warmup Steps  & 2000  & 100  \\
Peak LR       & $3\times10^{-4}$  & $8\times10^{-6}$ \\
Minimum LR    & $3\times10^{-5}$  & $1\times10^{-6}$ \\
Optimizer     & \texttt{AdamW} &  \texttt{AdamW} \\
$\beta_1$     & 0.9   & 0.9   \\
$\beta_2$     & 0.95  & 0.95 \\
$\epsilon$       & $1\times10^{-5}$  & $1\times10^{-8}$ \\
Weight Decay  & 0.1   & 0.01 \\
LR Schedule   & \texttt{cosine}  & \texttt{cosine}  \\
Gradient Clipping & 1.0   & 1.0 \\
Gradient Reduce dtype & \texttt{fp32}   & \texttt{bfloat16}\\
Optimizer State dtype & \texttt{fp32}  & \texttt{bfloat16}    
\\ \bottomrule
\end{tabular}
\caption{Overview of pre-training hyperparameters of Fanar Star and Fanar Prime.}
\label{tab:hyperparam-fanar}
\end{table}

\begin{figure}[H]
    \centering
    \includegraphics[width=0.5\linewidth]{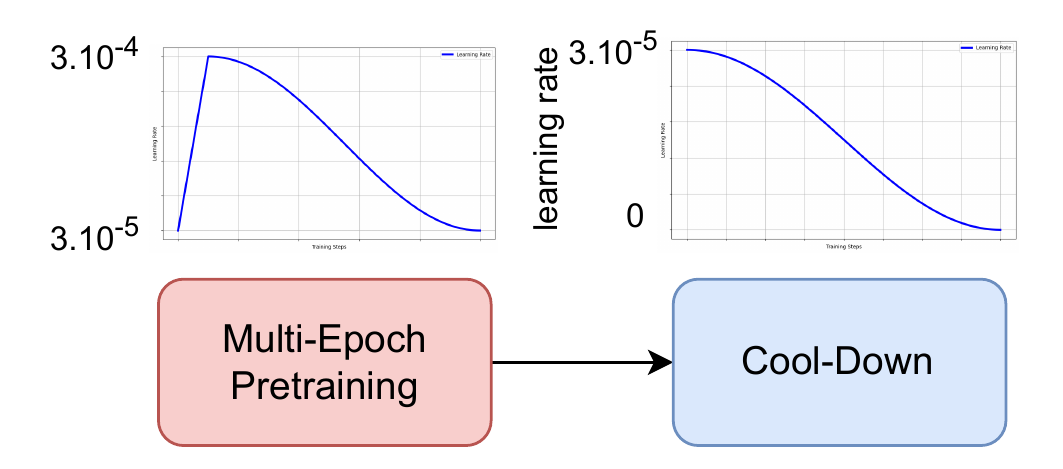}
    \caption{Pre-training phases: 1) multi-epoch training: training for 4 epochs over approximately 4T tokens. 2) cool-down phase: training with high-quality data with a decaying learning rate to zero.}
    \label{fig:pretrain_phases}
\end{figure}

\begin{figure}[H]
    \centering
    \includegraphics[width=0.55\linewidth]{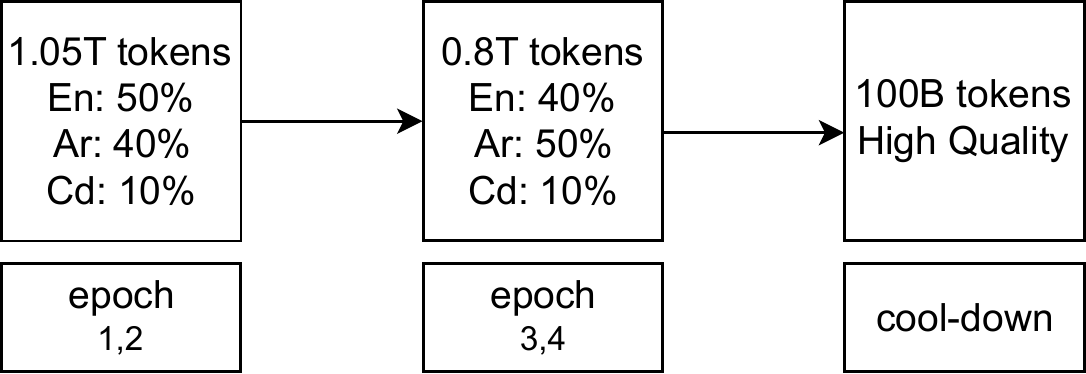}
    \caption{Curriculum learning in pre-training. phases. In the last epoch, additional model-based filtering are applied. The cool-down phase contains high-quality data such as books, encyclopedia, cultural and STEM related materials. Also conversational data and Wikipedia knowledge re-structured in multiple-choice questions to boost benchmarking capability. }
    \label{fig:curriculum_learning}
\end{figure}

\subsection{Continual Pre-training for {\FP}}

\subsubsection{Training Recipe}

\FP\ is built through the continual pre-training of the Gemma-2-9B-Base model \citep{Riviere2024Gemma2I}, a robust multilingual foundation model initially pre-trained on eight trillion tokens through knowledge distillation from a larger, undisclosed model developed by Google. Although specific details regarding the training corpus of Gemma-2-9B are limited, its multilingual capabilities and extensive pre-training make it an ideal candidate for further adaptation.

As part of this adaptation, we performed vocabulary pruning on the original 250,000-token vocabulary of Gemma-2-9B, reducing it to 128,256 tokens for \FP. This reduced model size by decreasing the total parameters from 9.2 billion to 8.78 billion, enhancing computational efficiency without compromising performance.

Similar to the pre-training from scratch in \FS, the continual pre-training process for \FP\ followed a two-stage curriculum strategy, utilizing a similar balanced data composition of 45\% Arabic, 45\% English, and 10\% code. Since we started from a competent model, we chose to slightly alter the data composition as compared to {\FS}. Unlike the training strategy employed for \FS, \FP\ underwent a single-epoch pre-training phase, followed by a cool-down stage designed to enhance generalization. The results presented in this report are for a checkpoint around the ~600B token mark, while the training is continuing to complete the full epoch. The cool-down stage utilized the same high-quality dataset curated for \FS, ensuring consistency and alignment with the overarching training objectives. Through this training recipe, \FP\ was continually pre-trained on a total of 650 billion tokens.

\subsubsection{Optimization Configuration}
We reused the configuration of {\FP} to train {\FS} with a few exceptions. The global batch size was set to 1071 samples, totaling $\sim$4.4 million tokens per optimization step. The learning rate was first linearly increased to $8\times10^{-6}$ for 100 steps, and the cosine annealed down to $1\times10^{-6}$. The hyperparameters for \FP\ are detailed in Table \ref{tab:hyperparam-fanar}.

\subsection{Pre-training Infrastructure and Frameworks}

The pre-training of both \FS\ and \FP\ was conducted on a 168 NVIDIA H100 80GB SXM5 GPUs, distributed across 21 nodes with 8 GPUs per node. These GPUs, powered by the NVIDIA Hopper architecture, are interconnected within each node via NVLink and NVSwitch, providing a bidirectional GPU-to-GPU bandwidth of 900 GB/s (450 GB/s in each direction). Inter-node communication was facilitated through high-speed InfiniBand connections, ensuring efficient data transfer across the cluster.

For the pre-training of Fanar Star, we utilized the OLMo framework\footnote{\url{https://github.com/allenai/OLMo}}, which is renowned for its advanced features and fine-grained control over training processes. This framework offers detailed artifact generation, including visualizations of layer weight dynamics, in-loop evaluation mechanisms, and back-tracing capabilities for training batches. These functionalities were invaluable for debugging, particularly in diagnosing and addressing loss spikes during our pre-training. For instance, the back-tracing feature allowed us to pinpoint problematic training batches and inspect their raw text inputs, helping us identify and resolve dataset issues such as improper language filtering, thereby ensuring cleaner and more balanced training data.

Additionally, we implemented custom in-loop evaluation pipelines tailored to specific downstream tasks. This enabled us to monitor performance trends during training, providing early insights into the model's generalization capabilities. To streamline the data preparation process, we leveraged the Dolma pipeline\footnote{\url{https://github.com/allenai/dolma}} —a flexible and scalable toolbox designed for large-scale language model data curation and preprocessing.

For the continual pre-training of \FP, we adopted the LitGPT framework\footnote{\url{https://github.com/Lightning-AI/litgpt}}. This framework enabled seamless integration with the Gemma-2-9B-Base model, supporting efficient and scalable continual learning processes. Its modular design allowed us to focus on domain-specific optimizations, ensuring that the pre-trained model adapted effectively to new data distributions without compromising its foundational knowledge.

\section{Post-Training}
\label{sec:pt}
\fullwidthbox{
Supervised Fine-Tuning (SFT) and preferential learning to accentuate Arabic 
cultural and safety alignment distinguish the post-training phase.
}

Our post-training objective was to develop a model that could effectively follow general instructions and engage users in meaningful conversations.
Two additional key considerations guided our development process. First, the model needed to demonstrate strong Arabic proficiency to effectively serve our target user base. Second, it was essential to ensure that the model’s responses were not only helpful and harmless but also culturally aligned.
To accomplish this, Fanar’s post-training process combined supervised fine-tuning with preference learning.
The data needed to guide these training steps were acquired either through curating data from public sources and or generation by our post-training team over the course of training using a number of open models and also Fanar. 

Our post-training strategy distinguishes itself from existing techniques through three key steps:
\begin{itemize}
\item	\textbf{Sample-Level Data Quality Validation:} A capability-focused filtering process to curate a high-quality, bilingual supervised fine-tuning dataset from public sources.
\item	\textbf{Multi-Stage Training Workflow:} A progressive training approach that incorporates high-quality, task-representative data samples at multiple stages. This process mirrors the cool-down and annealing stages commonly used in pre-training, to reinforce the model’s ability to handle diverse tasks effectively.
\item	\textbf{Value-Aligned Synthetic Data Generation:} A data generation method tailored to align with Arabic cultural and religious values, addressing the challenge of developing a value-aligned preference/reward model.
\end{itemize}

The two key decisions we needed to address during training were how to generate Arabic instruction and preference data and how to balance the English and Arabic post-training datasets. While several recent models have demonstrated that synthetically generated data can be effective for creating instruction datasets~\citep{adler2024nemotron, dubey2024llama3herdmodels,gunter2024apple,lambert2024t}, the lack of open models with strong Arabic proficiency initially led us to rely on translating English data, particularly for core capability datasets.
To enhance and evaluate Arabic proficiency, we tested several open models and identified instruction-tuned models that demonstrated significant reliability in Arabic.  These included \texttt{Gemma-2-27B-it} from Google \citep{team2024gemma}, \texttt{Qwen2.5-72B-Instruct} from Alibaba \citep{qwen2.5}, \texttt{c4ai-command-r-plus} from Cohere \citep{cohere_for_ai_2024}, and \texttt{Llama-3.1-70B-Instruct} and\\ \texttt{Llama-3.1-405B-Instruct} from Meta \citep{dubey2024llama3herdmodels}. 
We leveraged these models for tasks such as filtering, judging, and data generation, ensuring compliance with the permissiveness of their respective licensing terms. 

Balancing the language composition of post-training datasets presented additional challenges. Although previous studies suggest that language models generate a shared conceptual space across languages, the transferability of task-specific capabilities between languages -- particularly dissimilar ones like English and Arabic -- remains poorly understood \citep{csaki2024sambalingo}. 
To address this, we conducted post-training in both languages by generating data samples tailored to each language, ensuring that each version was contextually appropriate and aligned with cultural and value considerations.

\begin{figure}[H] %
    \centering
    \includegraphics[width=\linewidth]{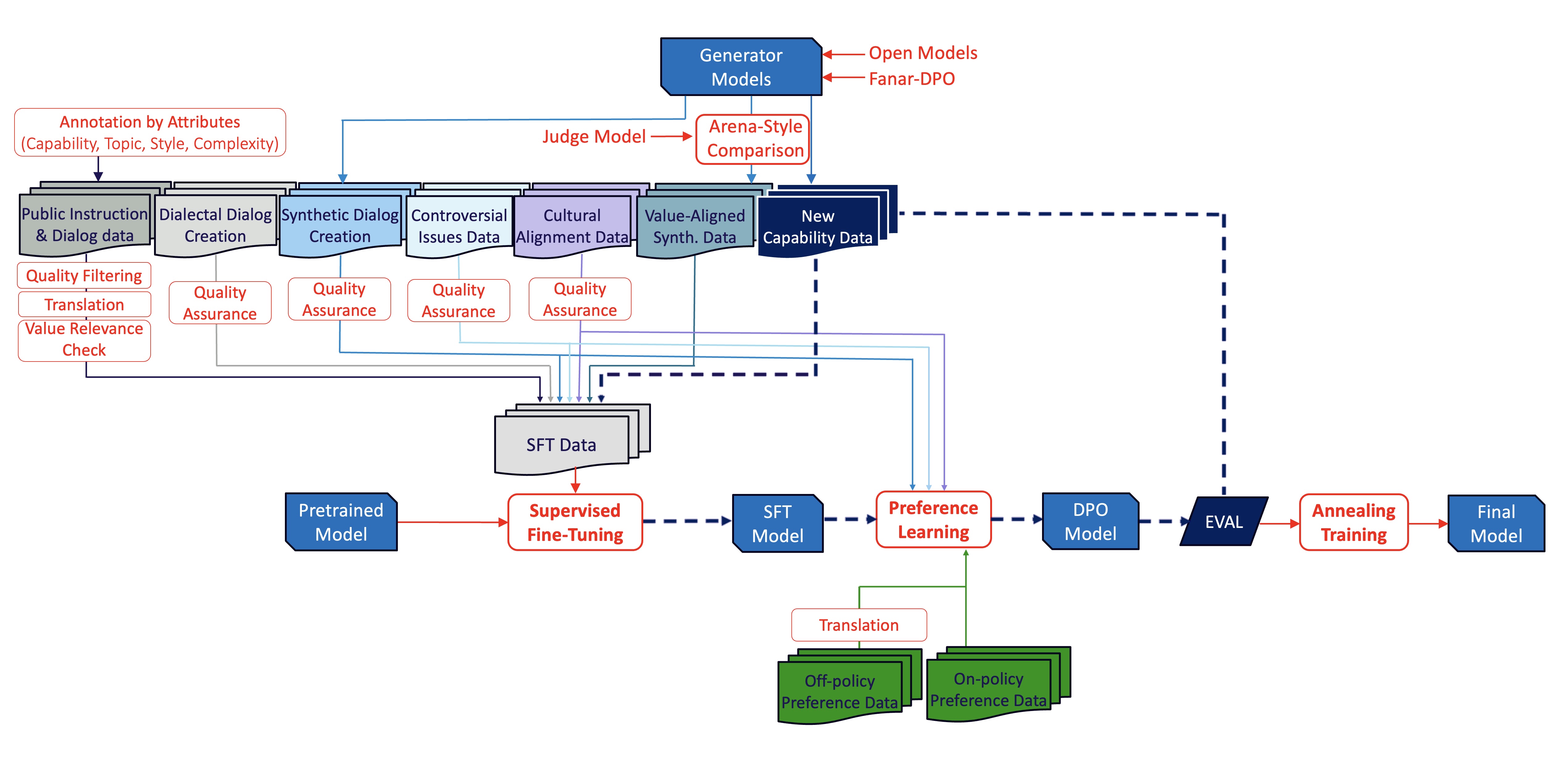} %
    \caption{Post-training workflow for Fanar.}
    \label{fig:workflow}
\end{figure}

Our training process follows a multi-stage training workflow, as illustrated in Fig. \ref{fig:workflow}.
In our earlier experiments, we found that combining general capability data with task-specific data requiring cultural and value awareness diminished the model’s effectiveness in handling the latter.
To address this, we implemented supervised fine-tuning (SFT) in multiple stages, performing additional training on what we identified as higher-quality data using a lower learning rate. Capabilities requiring value alignment were incorporated into the second stage of fine-tuning.
The instruction-tuned model then underwent further training through a preference optimization step, utilizing preference data generated either by our SFT model (on-policy data) or by external models (off-policy data).
Our development process was guided by a combination of benchmarks designed to evaluate the chat capabilities of our models and continuous feedback from external testers using our playground.
Our training concludes with an annealing stage that utilizes a highly curated, small subsample of SFT and preference data.
The same post-training stages were applied to both \FS\ and \FP\ models. 
Both models were trained to support a system prompt as part of their chat template. 

\subsection{Supervised Fine-Tuning}
The SFT data consists of several distinct splits, encapsulating a wide range of capabilities and behaviors, as illustrated in Fig. \ref{fig:workflow}.
This data is primarily obtained through three key processes: extensive filtering and annotation of publicly available SFT data, synthetic data generation, and a small yet critical amount of expert- and vendor-created data.
To optimize performance, we implemented a two-stage SFT training approach rather than a single training round, drawing inspiration from the way pre-training is performed in multiple cooling-off rounds using progressively higher-quality data.
Accordingly, the data relevant to behaviors that require a more nuanced interpretation and more challenging capabilities are introduced or repeated in the second round at a lower learning rate to reinforce learning. 
Our tests revealed that employing this two-stage training approach, rather than utilizing all the data in a single training round, results in a model that achieves higher performance in both automated benchmarks and user evaluations.
To the best of our knowledge, this two-stage SFT process is novel, particularly in its application for creating a more value-aligned model.

\subsubsection{Data Curation from Public Sources}
Generating high-quality samples for behavior mimicking has been a significant area of research, with various criteria proposed to evaluate sample quality \citep{xia2024less, shen2024rethinking, wang2023far, zhao2024long}. Although a vast amount of public instruction and dialogue data is available, much of it is automatically generated en masse using other language models. Consequently, even widely-used and well-regarded datasets often contain a substantial proportion of low-quality samples.
To address this, we began our curation process by annotating each instruction and dialogue sample along four key dimensions: topic, writing style, prompt complexity, and the capability reinforced by each data sample. 
For capability annotation, we expanded upon the capability-based test and evaluation taxonomy introduced in \citep{scale2023evaluation} by incorporating safety to develop a framework of 11 core capabilities.
All classification tasks were conducted using the smallest Llama model, Meta-Llama-3.1-7B-Instruct.

Annotated data samples underwent a three-level filtering process to extract the highest-quality samples from public datasets.
In the first level, each sample’s quality was assessed. For this purpose, we designed specific rubrics for each capability category to quantify how effectively a data sample reinforced that capability. Separate rubrics were applied to single-turn and multi-turn data samples \citep{zhang2024comprehensive}. 
Llama-3.1 models were used to evaluate these rubrics, assigning a quality score to each sample. 
The quality score, along with prompt complexity and response length, was then used to identify accepted samples within each quality category. (Separate criteria were applied to category sample depending on the number of available samples.)
Next, the selected samples were translated into Arabic using in-house translation models as well as Google Translate. The translated samples were further filtered to eliminate incoherent outputs, particularly those involving tasks such as translation, grammar, word riddles, or sorting, which often yielded poor results.
Finally, a value-relevance filter was applied to remove samples misaligned with cultural and religious values, ensuring the curated dataset met ethical and contextual standards. 

\begin{figure}[H] %
    \centering
    \includegraphics[width=1.0\linewidth]{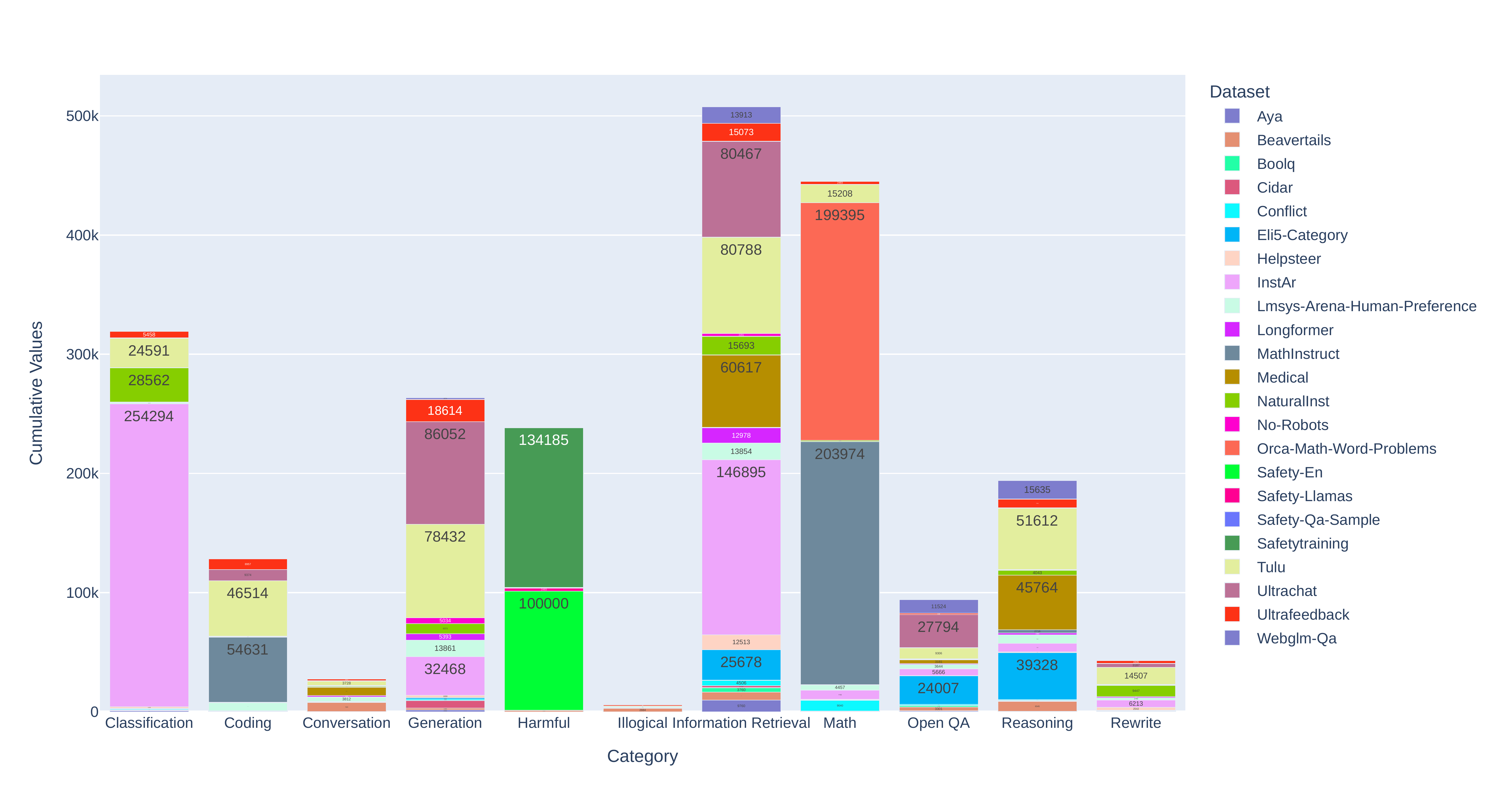} %
    \caption{Composition and provenance characteristics of the core capability dataset curated from public sources.}
        \label{fig:public-SFT-data}
\end{figure}

This process resulted in approximately 2.5 million instructions and dialogues across 11 categories in both languages.
Figure \ref{fig:public-SFT-data} presents the composition of our curated dataset, highlighting provenance characteristics and dataset size, sourced from public data after the three-level filtering process.
Notably, our observations indicate that public datasets often consist of a mix of data samples with varying quality, with few being uniformly high-quality.
This core capability data is utilized during the initial stage of supervised fine-tuning.

\subsubsection{Synthetic Data Generation}
One notable drawback of curating data from public sources is the tendency of the translation process to introduce typical English-language contexts into Arabic. This includes culturally specific elements such as personal names, geographic references, traditions, social norms, and lifestyle practices, which often fail to capture local cultural nuances.
To address this and ensure that our models are culturally and linguistically aligned with the preferences of Arabic-speaking users, we generated a substantial volume of synthetic data, including both single-turn and multi-turn dialogues.

Synthetic data generation, however, posed two key challenges. First, it required careful selection of models for generating completions. Rejection sampling was employed as a critical tool, generating multiple completions and selecting the best among them. However, our models trained on core capability data lacked the cultural awareness necessary to drive this generation effort effectively.
To address this, we leveraged open, large-parameter public models, complemented by well-engineered system prompts specifically designed to instruct these models to create culturally and religiously appropriate samples for Arabic-speaking populations.

The second challenge was validating the quality of the model-generated outputs.
This is typically achieved using a reward or preference model trained on extensive human preference data to assign a goodness score to each completion.
However, due to the lack of sufficient user preference data at the start of our training process, we employed the quantized version of the largest Llama model (Meta-Llama-3.1-405B-Instruct-FP8) as the judge model.
The core steps of our assessment process for a given set of prompts are as follows:
\begin{itemize}
    \item Generate output responses using multiple models for each prompt.
    \item For Arabic outputs, run a Spell Checker\footnote{\url{https://farasa-api.qcri.org/}}~\citep{mubarak2014automatic} to automatically detect spelling errors. Set a threshold to filter out outputs that are deemed as having too many errors. 
    \item Use the judge model to score responses on a scale of 1 to 10, considering scores of 8, 9, or 10 as good responses.
    \item Conduct an arena-style comparison to select the best response for each prompt by performing pairwise comparisons of high-scoring responses. Selection accounts for positional bias and response length.
    \item Create an SFT dataset using the selected prompts and responses.
\end{itemize}
To ensure meaningful cultural alignment, all prompts used at this stage focused on capabilities relating to creative writing, rewriting, in-context retrieval, and conversational queries, introducing a variety of contexts.
These prompts were drawn from two main sources: prompts discarded during the quality filtering stage due to low response scores based on our rubrics, and existing prompts from the curated dataset.

Another issue we identified was the model’s tendency to use English entity names and locations in creative-writing tasks, even when responding to Arabic queries. To address this, we regenerated responses from the curated dataset, specifically targeting queries containing the 100 most common English male and female names. Incorporating 30K contextually appropriate responses effectively mitigated this behavior. Additionally, we generated synthetic data to support a range of other capabilities, including precise instruction following, handling controversial topics, and uncertainty learning.

Our synthetic data generation process extensively utilized carefully crafted prompts to guide larger models with 27B+ parameters, ensuring the responses met our quality standards while aligning with Arabic cultural and religious values.
To achieve this, we developed three types of prompts: system prompts to direct text generation by the models, a system prompt for the judge model to evaluate and score responses from selected generation models, and a system prompt for the judge model to perform arena-style comparisons of pairwise responses.

Overall, through the use of stronger models, we generated close to a million samples in both Arabic and English, mainly focusing on culturally contextualizing and aligning the model’s responses.
The synthetically generated data was utilized in both stages of SFT, with value-alignment-related splits included in their entirety, while other splits were only subsampled.

\subsubsection{New Capability Data}
Our tests and in-loop evaluations during the post-training phase revealed several gaps and weaknesses. To address these, we enhanced the existing dataset with additional targeted datasets, each designed to introduce specific behaviors. These included:
\begin{itemize}
\item A dataset created by rephrasing and expanding a core set of manually generated prompts and responses to cover a closed set of user questions.
\item	Datasets generated by transforming high-quality textual sources focusing on IslamQ\&A, poetry, and humor into instructional formats.
\item	Datasets broadening the model’s capabilities on language tasks, such as  diacritization, question generation, and grammar correction.
\item	A vendor-generated dataset for dialectic dialogues.
\item	Expert-driven instructional and DPO data tailored to address nuanced topics, including controversial Islamic issues.
\end{itemize}

\subsection{Preference Learning}
To further enhance the model’s capabilities, we implemented Direct Preference Optimization (DPO) \citep{rafailov2024direct}, an offline reinforcement learning method that eliminates the need to explicitly build a reward model or sample from the model during training.
Our binary preference data was obtained from two primary sources. 
Off-policy preference data was derived from three key public datasets: UltraFeedback \citep{cui2023ultrafeedback}, HelpSteer \citep{wang2023helpsteer,wang2024helpsteer2}, and Nectar \citep{starling2023}. This data underwent the same quality and value-relevance filtering process applied during the curation of the SFT data to ensure alignment with our standards.
We also generated on-policy preference data by producing both accepted and rejected responses using our models. 
To distinguish between accepted and rejected responses, we employed the Reinforcement Learning with AI Feedback (RLAIF) approach, which primarily relied on the preferences provided by judge models.
However, when generating culturally aligned data to address nuanced cultural issues, we relied on user annotations to determine preferences for model-generated responses.

To establish trust in the use of a large language model (LLM) for preference annotation in Arabic, we conducted a user study.
We selected approximately 700 user-generated prompts where our model’s responses were initially disliked and scored below 7 by the Llama judge model.
Using our generator models, we created responses to these prompts, with the best response selected by the judge through scoring and arena-style comparison, as previously described.
Users were presented with prompts paired with two anonymized responses: one being the rejected response from our model and the other the accepted response generated by other models. To ensure unbiased evaluations, the judge’s selections were concealed, and each pair of responses was independently assessed by 5-6 users.
We aggregated the user annotations by conservatively scoring \textit{likes} and \textit{greats} as +1 and \textit{dislikes} as -1, calculating the total score for both the accepted and rejected responses. The score difference between the accepted and rejected responses was then computed for each prompt, where a negative score indicated disagreement of human annotators with the judge.
Our findings showed that the Llama 405B model achieved approximately 87\% agreement with human annotations, aligning with earlier studies on GPT-4’s agreement with human evaluators \citep{zheng2023judging}.

For generating on-policy data, prompts were selected from our curated dataset and user queries submitted through the playground with disliked responses. Accepted and rejected responses were produced by slightly varying the temperature settings to create pairs of outputs with noticeable deviations.
The responses were evaluated using two judge models, Llama-405B and Gemma-27B. The average score from these models was used to classify the responses. A response was marked as accepted only if its average score was eight or higher, while rejected responses required scores between two and six, ensuring a minimum gap of two points between the paired responses.
This process generated approximately 250K preference data samples, balanced across both languages, with around 20\% comprising on-policy data.

During preference optimization, we explored several DPO variants as alternatives, including IPO~\citep{azar2024general}, KTO~\citep{ethayarajh2024kto}, SimPO~\citep{meng2024simpo}, and ORPO~\citep{hong2024orpo}\footnote{LLaMA-Factory implementations are used.}. However, these approaches led to a decline in performance on our automated benchmarks, so we decided not to adopt them.
Additionally, we observed that applying preference data in batches, rather than processing the entire dataset in a single training run, led to a modest but consistent improvement in our benchmarks (+1-2\%). Updating the reference model every 40K samples yielded the best results, even though it introduced minor off-policy characteristics to the initial accepted and rejected responses.
During testing, we observed that the model occasionally failed to respond in the language of the user query. This issue was traced to an imbalance between the number of Arabic and English DPO samples, with the model defaulting to the majority language. To address this, we balanced the dataset between the two languages and incorporated some mismatched-language responses as rejected responses, which corrected the issue.

\subsection{Annealing Stage}
Our training ends with an annealing phase, where the learning rates for the SFT and preference optimization steps are reduced to near zero while presenting the model with a high-quality subsample of SFT and DPO data.
This final stage serves two main objectives. First, the training process involves a diverse range of tasks with varying levels of complexity, making it challenging to perfectly balance the data to reflect this variability. As a result, task interference becomes inevitable, especially at smaller parameter sizes.
For example, as additional capability data was incorporated into the training, the model began to struggle with certain math problems and nuanced responses that earlier versions handled effectively.
Thus, as the final set of training data presented to the model, this stage reinforces learning by helping the model retain the diverse range of tasks encountered in earlier phases.

In addition, this stage serves as a rapid response mechanism, allowing fast retraining of the model when harmful or culturally misaligned behaviors are identified. Rather than relying solely on such disliked responses to teach a new behavior or reinforce one, these samples are integrated with other capability data to create a balanced dataset, ensuring that previously learned capabilities are preserved.
While this annealing stage led to a slight decrease in automated benchmark performance, it noticeably reduced the user dislike rate. 
In our playground, both models—\FS\ and \FP\ —are deployed. Harder queries, such as those related to math, reasoning, and coding, are routed to \FP\ due to its superior performance in these areas. Since \FS\ handles most user queries, the annealing stage was applied only to it, leaving \FP\ unchanged.

\subsection{Infrastructure \& Hyperparameters}
All post-training activities were carried out on 2–3 nodes, each equipped with 8x H100 GPUs. Table \ref{tab:post-training-params} details the training parameters used during the supervised fine-tuning and preference optimization stages. Notably, the Gemma-based continually pre-trained model required a learning rate an order of magnitude smaller than the model trained from scratch.

\begin{table}[h!]
\centering
\caption{Post-Training Hyperparemeters}
\label{tab:post-training-params}
\begin{tabular}{c|c|c|c|c|c|c}
\toprule
\multicolumn{2}{c|}{Training}                        & Number of & \multicolumn{2}{c|}{Fanar-Star} & \multicolumn{2}{c}{Fanar-Prime}  \\
\multicolumn{2}{c|}{Phase}                           & Sample & Batch & LR & Batch & LR\\
\midrule
\multirow{2}{*}{\rotatebox[origin=c]{270}{SFT}}  & Stage-1 & 3.6M & 256 & 5.0e-06 & 640 & 5.0e-07 \\ %
                                                 & Stage-2 & 834k & 512 & 1.0e-06 & 640 & 1.0e-07\\  %
\midrule                                                 
\multicolumn{2}{c|}{DPO}                                   & 250k & 64 & 1.0e-07 & 128 & 1.0e-07 \\ 
\midrule
\multicolumn{2}{c|}{Annealing-SFT}                         & 5k & 128 & 6.0e-08  & 640 & 6.0e-08 \\ 
\multicolumn{2}{c|}{Annealing-DPO}                         & 4k & 64 & 3.0e-08 & 64 & 3.0e-08\\ 
\bottomrule
\end{tabular}
\end{table}

\subsection{Collection of User Feedback}
In the early stages of our model development, we implemented a playground equipped with evaluation functionality. This allowed users to compare responses from multiple models and provide feedback based on three attributes: \textit{like, dislike}, and \textit{great}. The first two attributes could be assigned to each model output, while \textit{great} was reserved for a single model output that stood out. (Our menus supported both languages depending on the user preference.)
When \textit{dislike} was selected, users were prompted to choose from eight predefined dislike categories, including lack of factuality, length issues, instruction adherence failures, insufficient harmlessness, refusal, cultural misalignment, and grammatical errors. This feedback mechanism enabled continuous monitoring of model improvements and helped identify systematic gaps in capabilities.

Our system engaged a team of 40–250 annotators at different stages, primarily Arabic speakers from various locations, who, at peak capacity, contributed up to 10K prompts and provided feedback on model responses each day.
Over time, our core user base grew to 130 individuals from diverse professional and academic backgrounds. On any given day, 60 to 90 of these users actively participated in model evaluation.
Additionally, students from several universities participated in the model testing efforts for varying durations.
During the early testing phase, users were free to prompt the model with any queries they preferred. However, as development progressed, testing became more guided, focusing more on capability-oriented evaluations.
Over 90\% of the prompts were in Arabic, and when testing was unguided, most user prompts ($\sim$70\%) tended to focus on open-domain question-answering queries.

In addition to highlighting capability gaps and areas of cultural misalignment, user feedback enabled us to identify issues such as grammatical errors and presentation problems, to trace them back to their sources, and to refine our data filtering processes. This feedback also informed adjustments to how we utilized judge and generation models. 
We observed that our model, trained from scratch, had a dislike rate of approximately 13\%, with the vast majority of dislikes related to factual accuracy. This can be safely attributed to the relatively small size of our pre-training corpus. 

\subsection{In-Loop Evaluations}
The training progress was continuously monitored using a combination of benchmarks designed to evaluate the instruction-following and conversational capabilities of our models.
For Arabic, we utilized a combination of five public and internal benchmarks. These included translated versions of MT-Bench \citep{zheng2023judging} and Alpaca-Eval \citep{alpaca_eval}, with modifications to humanities-related questions in MT-Bench to better suit the Arabic context~\citep{boughorbel2023analyzing}. To evaluate Arabic language proficiency, we used the development set of the BALSAM benchmark for tasks relevant to our use cases~\citep{BALSAM}. %
To further enhance evaluation, we developed a general capability benchmark called Almieyar, expanding upon the 10 capability categories used for annotating our SFT data.
User chats were also incorporated as a crucial evaluation source, offering valuable insights into real-world performance. These chats typically consisted of multi-turn dialogues, sometimes extending up to 100 turns, enabling a comprehensive assessment of model capabilities.
To ensure that English proficiency was maintained, we included the English counterparts of these benchmarks alongside IFEval~\citep{zhou2023instruction}. For both languages, evaluations were primarily conducted using the LLM-as-a-Judge framework, with GPT-4o or GPT-4 scoring the model responses.

To validate model improvements, we also incorporated user preferences. Specifically, we identified 555 data samples focusing on various capabilities, primarily inspired by challenging user prompts. Users compared the responses of the baseline model to those of its updated version using these standardized queries. The data samples included both single-turn and multi-turn dialogs; for the latter, users were shown only the response from the final turn.
Models that demonstrated improvements in both automated benchmarks and user comparisons were subsequently deployed in our playground for further testing. The following section presents a performance comparison of our two models against other models of similar parameter sizes across multiple-choice questions, conversational tasks, and instruction-following benchmarks used during post-training.

\section{Evaluation}
\label{sec:benchmark}
\fullwidthbox{
 \FS\ and \FP\ are evaluated against  Arabic-aware peer models on standard and proposed culturally aware benchmarks. 
}

Evaluation of large language models (LLMs), especially in the context of Arabic, remains in its early stages, with no universally accepted framework for comprehensively assessing their capabilities. In this work, we adopt the common practice of benchmarking our models against comparable baselines across a diverse set of tasks and formats. This approach aims to provide a detailed understanding of the models' performance and highlight their unique strengths.

The Fanar model family is trained on diverse datasets encompassing standard Arabic, dialectal Arabic, English, and code. This multilingual and multi-domain training enables the model to excel in various tasks, including reading comprehension, logical reasoning, knowledge extraction, and standard NLP applications. Our evaluation framework leverages widely recognized benchmarks for reasoning, reading comprehension, and question-answering in both Arabic and English. It also includes assessments for conversational abilities and instruction-following tasks. To address gaps in existing benchmarks, we have developed new datasets specifically designed to capture cultural and dialectal nuances. We plan to release these benchmarks for the benefit of the broader research community. The following sections provide detailed descriptions of the selected benchmarks and baseline LLMs, followed by an analysis of the evaluation results.

\subsection{Benchmarks}
We benchmark Fanar and baselines on three types of benchmarks: automatic evaluations, open-ended and conversational assessments, and human evaluations.

\subsubsection{Automatic evaluation with multi-choice questions}
We use LM-Evaluation-Harness\footnote{https://github.com/EleutherAI/lm-evaluation-harness}~\citep{eval-harness} version 0.4.3 as the backbone for most of the automatic evaluation tasks. We use the default task definitions for English tasks and create custom task configurations for Arabic tasks. For the Open Arabic LLM Leaderboard (OALL) benchmark, we use Hugginface's Lighteval\footnote{https://github.com/huggingface/lighteval}~\citep{lighteval} that supports the live leaderboard~\citep{OALL}.

\paragraph{English tasks} we opt for standard English benchmarks that capture key capabilities: 1) \textbf{MMLU} ~\citep{hendrycks2020measuring} tests for general world knowledge through 4-choice questions in 57 subcategories; 2) logical inference is tested through \textbf{PIQA} (Physical Interaction QA)~\citep{bisk2020piqa} comprising 3K binary commonsense questions, \textbf{Hellaswag}~\citep{zellers2019hellaswag} comprising 10K multi-choice continuation questions about commonsense events, \textbf{ARC Challenge}~\citep{clark2018think} comprising 2.5K difficult multiple-choice grade-school level science questions, and \textbf{Winogrande}~\citep{sakaguchi2021winogrande} which presents 1.7K binary fill-in-the-blank type logical questions; 3) mathematical skills are tested with \textbf{GSM8K} which comprises 1K multi-step math word problems.

\paragraph{Standard Arabic tasks} 1) The Arabic section of \textbf{MMMLU}~\citep{MMMLU}: an OpenAI expert translation of the MMLU dataset; 2) \textbf{ArabicMMLU}~\citep{koto2024arabicmmlu} which covers 40 subjects in 14,575 multiple-choice questions in Modern Standard Arabic (MSA); 3) The MSA-translation subset of \textbf{PIQA} provided by the AraDiCE project~\citep{mousi2024aradicebenchmarksdialectalcultural} to test commonsense reasoning; 3) the \textbf{Open Arabic LLM Leaderboard} suite~\citep{OALL} which is the average of 14 tasks.

\paragraph{Cultural benchmarks} 1) \textbf{Arabic Cultural Value Alignment} (ACVA)~\citep{huang2023acegpt} to measure culture understanding through  true/false questions generated by GPT-3.5 Turbo; 2) \textbf{Arab Cultural MCQ} an in-house developed task to capture cultural nuances in Arabic communities, comprising 1K multi-choice questions extracted from relevant cultural guides and sources and curated thoroughly. More details about the benchmark could be found in Appendix~\ref{appx:CulturalMCQ}.  

\paragraph{Dialectal benchmarks} 1) the Arabic part of \textbf{Belebele}~\citep{bandarkar2023belebele}: the average of six subsets of the dialectal reading comprehension task in MSA, Levantine, Gulf, Egyptian, Iraqi and Moroccan dialects; 2) \textbf{Almieyar}: an in-house task that measures competency in different aspects of Arabic language and its dialects. More details about the benchmark could be found in Appendix~\ref{appx:Almieyar}; and 3) human-translated subset of \textbf{PIQA} and \textbf{ArabicMMLU} to Egyptian and Levantine dialects provided by the AraDiCE project~\citep{mousi2024aradicebenchmarksdialectalcultural}.

\subsubsection{Conversation and instruction following evaluations}
To benchmark the models' ability to follow instructions and engage in dialog, we utilize verifiable instructions and capable LLMs as judges in open-ended generation tasks. 
\paragraph{English tasks} we benchmark using 1) MT-Bench~\citep{zheng2023judging}, a set of 80 two-turn open-ended questions that test model's ability in writing, roleplay, reasoning, math, coding, extraction and STEM and humanities knowledge, judged by GPT-4; 2) AlpacaEval~\citep{alpaca_eval} a set of 805 instructions derived from a diverse set of instruction sources, judged by GPT-4; and IFEval~\citep{zhou2023instruction} which comprises 25 verifiable instructions and 541 prompts.

\paragraph{Arabic tasks} we benchmark using 1) the Arabic MT-Bench~\citep{boughorbel2023analyzing} which is a curated and culturally-appropriated translation of the original MT-Bench; 2) a translation of AlpacaEval to Arabic scored by GPT-4; %
and 4) our in-house conversational benchmark, user-chats, which includes 3.5K single- and multi-turn chats  (averaging 2.7 turns per chat, with a maximum of 100 turns) from our testers, spanning a diverse range of topics and capabilities. The user prompts from each chat session are provided to the model, and GPT-4o is used to judge the quality of the answers.

\subsubsection{Human evaluation}
As instruction datasets might not be representative of real usage, we created a conversational test set of 555 single- and multi-turn dialogues, primarily drawn from our user messages. Model responses are generated for each prompt and presented to three human evaluators to express their preferences. To streamline the evaluation, users were shown only the final prompt and its corresponding model response rather than the full sequence of responses.

\subsection{Baselines}

\begin{table}[ht!]
    \centering
    \resizebox{0.6\textwidth}{!}{%
    \begin{tblr}{
    colspec={lccc},
    rowspec={QQ[colour1]Q[colour1]Q[colour1]Q[colour1]Q[colour1]Q[colour1]Q[colour2]Q[colour2]Q[colour2]Q[colour1]Q[colour2]Q[colour1]Q[colour1]Q[colour2]},
    hline{2} = {-}{},
    hline{3} = {-}{},
    vline{2-4} = {1-17}{},
    }    
    & Parameters & \SetCell[c=2]{c}Tokens \\
    &&Pre-training & Continual training\\
    Jais-family-6p7b& 6.7B & 0.48T & -\\
    OLMo-7B-0724-hf &7B &  2.46T & -\\
    \textbf{\FS} &7B & 3T & -\\
    ALLaM-Base (from scratch) &7B & 5.2T & -\\
    Qwen2.5-7B &7B & 18T& - \\
    ALLaM-Base &7B & 2T (Llama 2) & 1.2T \\
    AceGPT-v2-8B  &8B & 15T (Llama 3) & 0.11T\\
    Aya-Expanse-8B & 8B & - & - \\
    Llama 3.1  &8B & 15T & -\\
    \textbf{\FP} &8.7B & 8T (Gemma-2-9b) & 1.0T\\
    Gemma-2-9b   &9B & 8T & -\\
    Jais-13B     &13B & 0.4T & -\\
    Jais-adapted-13b &13B & 2T (Llama 2) & 0.28T\\
    \end{tblr}
    }
    \caption{Fanar and baseline model sizes and training data statistics. The shaded rows indicate the models which have been continually-trained from some base model as compared to training from scratch. A `-' indicates that information is not available or not applicable.}
    \label{tab:baselines}
\end{table}
We choose a number of capable LLMs that have been shown to exhibit excellent bilingual performance in Arabic and English in the same size range as Fanar models. These models are:
\begin{itemize}
    \item Jais~\citep{jaisfamilymodelcard}: a bilingual Arabic-English family of models. We use the 6.7B parameter \texttt{Jais-family-6p7b} model and \texttt{Jais-13B}~\citep{sengupta2023jais}, which are both trained from scratch, (plus their corresponding chat models \texttt{Jais-family-6p7b-chat} and \texttt{Jais-13B-chat}), and  the newer \texttt{Jais-adapted-13B} that has been evolved from Llama 2 (and its corresponding \texttt{Jais-adapted-13B-chat}).
    \item OLMo~\citep{olmofamilymodelcard}: the family of models whose architecture and training framework were the starting point for \texttt{\FS}. We benchmark the \texttt{OLMo-7B-0724-hf} base model  and its corresponding \texttt{OLMo-7B-0724-Instruct-hf} chat variant.
    \item Allam~\citep{bari2024allamlargelanguagemodels}: A family of Arabic-centric models. We report results for the 7B from-scratch \texttt{ALLaM-Base (from scratch)} and the 7B model evolved from Llama 2  \texttt{ALLaM-Base}. We also report results for the corresponding chat models \texttt{ALLaM-Instruct (from scratch)} and \texttt{ALLaM-Instruct}. Because Allam models are only released via endpoints, we reproduce all the results that require access to logits from the Allam report~\citep{bari2024allamlargelanguagemodels}.
    \item Qwen 2.5~\citep{qwenfamilymodelcard}: a series of multilingual models with strong performance in Arabic. We benchmark \texttt{Qwen2.5-7B} and its corresponding fine-tuned sibling \texttt{Qwen2.5-7B-Instruct}.
    \item AceGPT v2~\citep{acefamilymodelcard}: An Arabic-centric model built on top of Llama 2. We benchmark \texttt{AceGPT-v2-8B} and its chat version \texttt{AceGPT-v2-8B-Chat}.
    \item Aya Expanse~\citep{ayafamilymodelcard}: a family of fine-tuned models with highly advanced multilingual capabilities. We select the \texttt{Aya-Expanse-8B} variant.
    \item Llama 3.1~\citep{llamafamilymodelcard}: open models from Meta. We benchmark the similarly-sized \texttt{Llama-3.1-8B} and its chat version \texttt{Llama-3.1-8B-Instruct}.
    \item Gemma 2~\citep{gemmafamilymodelcard}: a family of light-weight models from Google. We benchmark \texttt{Gemma-2-9B}, a student model distilled from a much larger teacher and the starting point of \texttt{\FP}, and its corresponding instruction-tuned version \texttt{Gemma-2-9B-it}.
\end{itemize}

Table~\ref{tab:baselines} presents statistics about the chosen baselines.

\subsection{Evaluation results}
\subsubsection{Base models}
We present in Table~\ref{tab:Ar_base} the benchmarking results of base Fanar and baselines on Arabic benchmarks, and in Table~\ref{tab:En_base} the English benchmarks. All the results reported are in \textit{accuracy} percentage for multi-choice benchmarks with character output (e.g. `A', `B', `C' and `D' as in \texttt{MMLU}), and \textit{normalized accuracy} for multi-choice benchmarks with textual answer (e.g. \texttt{PIQA}). For the generation task \texttt{GSM8K} we report the results for \textit{exact\_match} with flexible extract.

The shaded rows in the tables below indicate the models which have been continually-trained from some base model as compared to training from scratch (refer to Table~\ref{tab:baselines}), and we order the rows by model size. The best score for a benchmark is shown in \textbf{bold}, while the second best is indicated by an \underline{underline}.
\begin{table}[ht!]
    \centering
    \resizebox{\textwidth}{!}{%
    \begin{tblr}{ 
    colspec={lcccccccc},
    rowspec={QQQ[colour1]Q[colour1]Q[colour1]Q[colour1]Q[colour1]Q[colour2]Q[colour2]Q[colour1]Q[colour2]Q[colour1]Q[colour1]Q[colour2]QQ[colour1]Q[colour1]Q[colour1]Q[colour1]Q[colour1]Q[colour2]Q[colour1]Q[colour2]Q[colour1]Q[colour1]Q[colour2]},
    width = \linewidth,
    vline{3} = {3-14}{},
    vline{3} = {17-27}{},
    hline{1-3} = {-}{},
    hline{15} = {2}{-}{2pt},
    hline{16} = {2}{-}{},
    hline{17} = {2}{-}{},
  }
  && MMMLU(Ar) & \SetCell[c=2]{c}ArabicMMLU & & PIQA(Ar) & OALL & ACVA & CulturalMCQ \\
  &&      {\small 0-shot}      & \small 0-shot & \small 3-shot   &   \small 0-shot  &   \small 0-shot   &  \small 5-shot  & \small 3-shot \\
Jais-family-6p7b & 6.7B&  25.34 &  29.12  & 34.81      &\underline{65.18}&    37.55  &  60.13 &34.00  \\
OLMo-7B-0724-hf &7B&      25.37    &  27.22  &  35.35      &49.95&  37.20   &  49.60 & 29.30  \\
\textbf{\FS} &7B&    36.16      & 38.75 &     50.20    &65.13&  42.02    &  68.60 & 62.80 \\
ALLaM-Base (from scratch) &7B&     36.28*     & 44.45*  &  \centering -   &\centering -&   \centering -  & 68.46*  &\centering - \\
Qwen2.5-7B &7B&    51.77        &   \underline{60.15}   & \underline{65.08}     &59.68&    48.66  &  \underline{80.37}  & 65.70 \\
ALLaM-Base &7B&      34.42*    & 41.52*  &  \centering -   &\centering -& \centering -   &  66.18* &\centering - \\
AceGPT-v2-8B  &8B&        41.71    &  45.96 &58.55         &63.17&  43.58   &  78.36  &67.50\\
Llama 3.1  &8B&      43.21      &    46.56   & 55.73    &57.51&  43.01   &  77.72  &60.00 \\
\textbf{\FP} &8.7B&      \textbf{57.30}    & \textbf{61.14} &  \textbf{67.35} &\textbf{65.83}&  \textbf{54.79 }    &   \textbf{81.40 }  &\textbf{71.90}  \\
Gemma-2-9b   &9B&    \underline{54.04}     & 57.80 &   64.32        &63.98&    \underline{50.24}  &  79.66  &\underline{68.60} \\
Jais-13B     &13B&   29.91      &  35.77  & 39.23       &64.96&  38.20   &   62.70   &35.60 \\
Jais-adapted-13b &13B& 34.01 & 42.93 &51.96&65.02&40.79&73.52&60.90& \\

 && Belebele(Ar) & Almieyar(Ar) & PIQA(Egy)& PIQA(Lev) & ArabicMMLU(Egy)& ArabicMMLU(Lev) \\
 &&  \small 3-shot & \small 3-shot &   {\small 0-shot}      & \small 0-shot & \small 0-shot   &   \small 0-shot  \\
Jais-family-6p7b & 6.7B& 34.54 & 32.17 & 60.23 &	58.38   &28.50	&29.46 \\
OLMo-7B-0724-hf &7B&  30.87&41.52 &   51.41 &	50.05 &  26.72&	26.32\\
\textbf{\FS} &7B&  47.76&59.10 &  60.88	&57.78&  29.31	 & 39.75  \\
Qwen2.5-7B &7B&   71.72 & \underline{76.81} &  57.51	&55.44   &   47.33	& 49.26\\
AceGPT-v2-8B  &8B&  60.61 &66.83 &   61.48&	56.75 &      43.40	& 40.96\\
Llama 3.1  &8B& 61.59 &63.84&  55.28 &	53.81    &  41.44	& 38.39\\
\textbf{\FP} &8.7B& \textbf{79.37}  &\textbf{77.68} &61.97	& 57.78     & \textbf{55.68}	& \textbf{55.41}\\
Gemma-2-9b   &9B&  \underline{75.31} &73.82 & 60.17 &	58.05   & \underline{49.61}	& \underline{47.15}\\
Jais-13B     &13B&  35.39 &37.03 & \textbf{62.73} &	\underline{58.65}   &  29.65&	36.38 \\
Jais-adapted-13b &13B& 43.02 &62.34  &\underline{62.19}&	\textbf{59.25}&  38.24 &	37.93
 
    \end{tblr}}
    \caption{Arabic benchmarks for Fanar base models and baselines. All the reported results are in accuracy or normalized accuracy. The top half contains benchmarks of world knowledge/logical inference and cultural benchmarks, and the lower half presents dialectal Arabic benchmarks. \FP achieves the best accuracy in most of the benchmarks.\\
    \\\scriptsize{*~ALLaM results are reproduced from the technical report~\citep{bari2024allamlargelanguagemodels} and not validated with our benchmarking setup}}
    \label{tab:Ar_base}
\end{table}

\texttt{\FP} offers excellent performance in Arabic benchmarks, achieving the best result in most of the benchmarks including the culture ones, with the only exception being dialectal PIQA. For English, it consistently beats its starting point \texttt{Gemma-2-9b-it} in most benchmarks except in PIQA and ARC Challenge, and achieves best or second best score all over.  

\begin{table}[ht!]
    \centering
    \resizebox{.8\textwidth}{!}{%
    \begin{tblr}{ 
    colspec={lcccccccc},
    rowspec={QQQ[colour1]Q[colour1]Q[colour1]Q[colour1]Q[colour2]Q[colour1]Q[colour2]Q[colour1]Q[colour2]Q[colour1]Q[colour1]Q[colour2]},
    width = \linewidth,
    vline{3} = {3-14}{},
    hline{1-3} = {-}{},
  }
  && \SetCell[c=2]{c}MMLU && PIQA & Hellaswag & GSM8K  & ARC Challenge & Winogrande  \\
  &&      \small 0-shot   &  \small 5-shot   & \small 0-shot & \small 0-shot     &   \small 5-shot   &  \small 0-shot  & \small 0-shot   \\
Jais-family-6p7b & 6.7B&  29.03 &32.50&  75.95  & 69.28 &  6.90   &  40.27 & 65.11  \\
OLMo-7B-0724-hf &7B&    49.01     &54.11& 80.25  &  77.83  & 29.42  &  43.69 & 67.56  \\
\textbf{\FS} &7B&   48.27  &48.43& 76.12 & 71.88 &  38.89  & 39.76 & 61.01   \\
Qwen2.5-7B &7B&   \textbf{71.71}    &\textbf{74.18}& 79.71   & 78.95  &  \textbf{83.24}   & 51.37  & 73.01 \\
ALLaM-Base &7B&   40.71*    &-&   79.00* &  76.17*   & 16.98*   &  45.65* & 68.90*   \\
ALLaM-Base (from scratch) &7B&   42.91*    &-& 80.58* &  76.26*  &   16.15*  &  43.52* & 68.43*   \\
AceGPT-v2-8B  &8B&    59.98  &63.55& 80.03  & 76.97 & 30.93   &   49.40 & 73.01   \\
Llama 3.1  &8B&  63.10  &65.10&   81.01  &  78.95  &  51.63  &  53.41  & 73.72 \\
\textbf{\FP} &8.7B&    \underline{69.24}  &\underline{71.14}& \underline{81.34} & \textbf{80.56}  &   \underline{71.42}   &  \underline{60.15}   & \textbf{75.37}    \\
Gemma-2-9b   &9B&     68.50    &70.60& \textbf{82.97} &  \underline{79.82} & 67.10  & \textbf{65.53} & \underline{74.19} \\
Jais-13B     &13B&   33.19     &34.72&  77.91  &  71.77  & 11.14  &  41.89 & 68.43  \\
Jais-adapted-13b &13B& 45.18 &50.42& 78.94 & 78.02&18.95& 48.55 & 71.67 
    \end{tblr}}
    \caption{English benchmarks for Fanar base models and baselines. \FP\ has the highest scores for Hellaswag and Winogrande, is the second best for the rest of the benchmarks after \texttt{Qwen2.5-7B} and  \texttt{Gemma-2-9b-it}, and it also beats its starting point \texttt{Gemma-2-9b-it} in MMLU and GSM8K.\\
    \\\scriptsize{*~ALLaM results are reproduced from the technical report~\citep{bari2024allamlargelanguagemodels} and not validated with our setup}}
    \label{tab:En_base}
\end{table}

\subsubsection{Instruction-tuned models}
We present in Table~\ref{tab:Ar_instruct} the benchmarking results of fine-tuned Fanar and baselines on Arabic benchmarks, and in Table~\ref{tab:En_instruct} the English benchmarks. As before, we report results in \textit{accuracy} for multi-choice benchmarks with character output, \textit{normalized accuracy} for multi-choice benchmarks with textual answer, and  \textit{exact\_match} with flexible extract for \texttt{GSM8K}. For LLM-as-a-judge benchmarks the results are out of 10.

As can be seen from the tables, \texttt{\FP\ Instruct} is a strong competitor in Arabic, producing top scores in many automated and generative tasks.
In nearly all Arabic evaluations, our models ranked among the top two performers, with the Gemma-based model particularly excelling.
This trend was also observed in English benchmarks, where the strengths of the Gemma, Llama, and Qwen models became more pronounced.
\begin{table}[ht!]
    \centering
    \resizebox{\textwidth}{!}{%
    \begin{tblr}{ 
    colspec={lcccccccccc},
    rowspec={QQQ[colour1]Q[colour1]Q[colour1]Q[colour1]Q[colour1]Q[colour2]Q[colour2]Q[colour2]Q[colour1]Q[colour2]Q[colour1]Q[colour1]Q[colour2]
    QQ[colour1]Q[colour1]Q[colour1]Q[colour1]Q[colour1]Q[colour2]Q[colour2]Q[colour1]Q[colour2]Q[colour1]Q[colour1]Q[colour2]
    QQ[colour1]Q[colour1]Q[colour1]Q[colour1]Q[colour1]Q[colour2]Q[colour2]Q[colour1]Q[colour2]Q[colour1]Q[colour1]Q[colour2]Q[colour2]},
    width = \linewidth,
    vline{3} = {3-15}{},
    vline{3} = {18-28}{},
    vline{3} = {30-44}{},
    hline{1-3} = {-}{},
    hline{16} = {2}{-}{2pt},
    hline{17-18} = {2}{-}{},
    hline{29} = {2}{-}{2pt},
    hline{30} = {2}{-}{},
  }
  && MMMLU(Ar) & \SetCell[c=2]{c}ArabicMMLU & & PIQA(Ar) & OALL & ACVA & CulturalMCQ\\
  &&      {\small 0-shot}      & \small 0-shot & \small 3-shot   &   \small 0-shot   &   \small 0-shot   &  \small 5-shot  & \small 3-shot\\
    Jais-family-6p7b-chat & 6.7B& 41.59  &  54.92  &   55.80    &62.51&  48.20    &  72.04 &64.10  \\
    OLMo-7B-0724-instruct-hf &7B&    27.14      &  34.75  &   35.88     &50.22&   38.43  & 50.94  & 34.40  \\
    \textbf{\FS\ Instruct} &7B&    40.67      & 51.32 &   53.24      &\textbf{69.86}&  48.76    &  74.55 & 61.40  \\
    ALLaM-Instruct (from scratch)$^{*}$ &7B&      51.38*     &  \textbf{69.16*}  &  \centering -   &-&   \centering -  & 79.59*  &\centering - \\
    Qwen2.5-7B-Instruct &7B&    {55.63}        &  61.55    &  63.96   &60.55&   54.19   & 78.09   & 68.10\\
    ALLaM-Instruct$^{*}$ &7B&       49.60*    & \underline{66.90*}  &  \centering -   &-& \centering -   &   80.33* &\centering - \\
    AceGPT-v2-8B-Chat  &8B&   51.16         & 61.02  &    62.61     &64.58&  50.16   &  77.66  & \underline{68.90} \\
    Aya-Expanse-8B & 8B & 47.14 & 58.82 & 60.10 &  63.00 & 53.18  & 77.11 & 67.30 \\
    Llama 3.1-Instruct  &8B&    47.58        &   56.50    &  59.05   &58.11&  47.81   &  76.67  & 66.10  \\
    \textbf{\FP\ Instruct} &8.7B& \textbf{58.74} & 64.90
 & \textbf{67.82}  & \underline{67.14}	& \textbf{63.81} 	& \textbf{80.85} & \textbf{70.60}\\
    Gemma-2-9b-it   &9B&   \underline{57.93}   & 63.43 &  \underline{64.16}   &61.26&  \underline{56.11}    &  \underline{80.67}  &68.60 \\
    Jais-13B-chat     &13B&    40.44     &  57.30  &   55.36    & 65.72 &  49.76   & 71.88     &64.90 \\
    Jais-adapted-13b-chat &13B& 44.45 &57.81&58.97& 61.10&46.41&75.24&65.30\\

    && Belebele(Ar) & Almieyar(Ar) & PIQA(Egy)& PIQA(Lev) & ArabicMMLU(Egy)& ArabicMMLU(Lev) \\
    &&  \small 3-shot & \small 3-shot &   {\small 0-shot}      & \small 0-shot & \small 0-shot   &   \small 0-shot  \\
    Jais-family-6p7b-chat & 6.7B& 65.11 &61.72 & 60.12 &	57.24	  &49.11	& 47.49\\
    OLMo-7B-0724-instruct-hf &7B&  34.54 &45.39&   49.67	& 50.27	& 28.18 &	27.88	\\
    \textbf{\FS\ Instruct} &7B&  62.02 &58.98&  {62.30}	&\underline{59.25}&   46.63 &	47.15  \\
    Qwen2.5-7B-Instruct &7B&  74.61 &\underline{75.69}& 58.65	& 56.04  &   48.74	& 53.42 \\
    AceGPT-v2-8B-Chat  &8B& 74.56 &74.31&   61.32 &	56.91	&     54.53 &	53.91\\
    Aya-Expanse-8B & 8B & 70.78 & 70.20 & 59.41	 & 56.53 & 53.52& 53.71 \\
    Llama 3.1-Instruct  &8B& 66.72 & 69.70&  55.39	& 54.24	   & 46.85 &	47.52 \\
    \textbf{\FP\ Instruct} &8.7B& \textbf{82.48} &\textbf{78.30}&\textbf{63.60}&\textbf{59.74}& \textbf{59.22} &  \textbf{60.14}\\
    Gemma-2-9b-it   &9B&  \underline{78.31}  & 73.69& 59.96	& 57.24 & \underline{57.95} &	\underline{59.25}\\
    Jais-13B-chat     &13B&  66.39&62.34& \underline{62.40} &	59.19&	  51.30	& 51.87  &	\\
    Jais-adapted-13b-chat &13B& 67.52&65.46 & 58.05 &	55.77	& 52.87	& 53.59	\\

      && MT-Bench(Ar)& Alpaca LCW (Ar)  & User-Chats \\
      
    Jais-family-6p7b-chat & 6.7B& 6.47&25.60 &8.43 &\\
    OLMo-7B-0724-instruct-hf &7B&2.11& 1.04 &2.60 & \\
    \textbf{\FS\ Instruct} &7B& 6.89& 36.81 &9.46 & \\
    ALLaM-7B-Instruct$^{**}$ &7B&  7.61& 33.16 &9.43 & \\
    Qwen2.5-7B-Instruct &7B& 8.63 & 33.61 &8.65&\\
    AceGPT-v2-8B-Chat  &8B& 7.48  & 25.76 &9.01& \\
    Aya-Expanse-8B & 8B & 8.51  & \textbf{62.95}& \underline{9.52}\\
    Llama 3.1-Instruct  &8B&  6.95  & 14.92 &7.59 & \\
    \textbf{\FP\ Instruct} &8.7B& \textbf{8.93}&  \underline{55.60} & \textbf{9.66}& \\
    Gemma-2-9b-it   &9B& \underline{8.88}  & 49.60 & 9.27&\\
    Jais-13B-chat     &13B& 4.31&6.44 &6.73& \\
    Jais-adapted-13b-chat &13B& 6.61 & 26.62 &8.97&
    \end{tblr}
    }
    \caption{Arabic benchmarks for instruction-tuned Fanar and baselines. We show in the table the same benchmarks that were used for base models, in addition to the LLM-as-a-judge benchmarks and human evaluation in the bottom. \FP\ Instruct shows superior performance in most of the standard benchmarks, all the cultural assessments and dialectal tasks, and offers competitive performance in generative open-ended tasks judged by LLMs and humans.\\
    \\\scriptsize{*~ALLaM results are reproduced from the technical report~\citep{bari2024allamlargelanguagemodels} and not validated with our setup}\\
     \scriptsize{** The ALLaM model was accessed via the Microsoft Azure cloud platform.} 
    }
    \label{tab:Ar_instruct}
\end{table}

\begin{table}[ht!]
    \centering
    \resizebox{.8\textwidth}{!}{%
    \begin{tblr}{ 
    colspec={lccccccccc},
    rowspec={
    QQQ[colour1]Q[colour1]Q[colour1]Q[colour1]Q[colour1]Q[colour2]Q[colour2]Q[colour2]Q[colour1]Q[colour2]Q[colour1]Q[colour1]Q[colour2]
    QQ[colour1]Q[colour1]Q[colour1]Q[colour1]Q[colour1]Q[colour2]Q[colour2]Q[colour1]Q[colour2]Q[colour1]Q[colour1]Q[colour2]Q[colour2]},
    width = \linewidth,
    vline{3} = {3-15}{},
    vline{3} = {17-29}{},
    hline{1-3} = {-}{},
    hline{16} = {2}{-}{2pt},
    hline{17} = {2}{-}{},
  }
  && \SetCell[c=2]{c}{MMLU} && PIQA & Hellaswag & GSM8K & ARC Challenge & Winogrande\\
  &\small &      \small 0-shot &      \small 5-shot      & \small 0-shot & \small 0-shot     &   \small 5-shot   &  \small 0-shot  & \small 0-shot \\
Jais-family-6p7b-chat & 6.7B &50.50 & 49.42 & 74.05 & 72.04 & 55.04    & 44.62  & 62.35  \\
OLMo-7B-0724-instruct-hf &7B&   51.69     &52.29& 79.00 & 79.99  & 18.42  &  47.01 & 65.11  \\
\textbf{\FS\ Instruct} &7B&   47.38 &46.83& 79.11 & 73.34 &  42.84  & 45.90 & 64.48   \\
ALLaM-Instruct (from scratch)$^{*}$ &7B&    42.91*    & - & 80.58* &  76.26*  &   53.60*  &  52.05* & 69.93*  \\
Qwen2.5-7B-Instruct &7B&   \textbf{72.08}    &\textbf{74.21}&  79.92 & {80.44} &  \underline{77.33}   & 55.03  & 71.35  \\
ALLaM-Instruct$^{*}$ &7B&    40.71*    & - &   79.00* &  76.17*   & 49.28*   &  51.45 &70.56* \\
AceGPT-v2-8B-Chat  &8B&  64.86   &66.45& 80.58 & 79.21 &  57.01  &  53.50  & 73.72   \\
Aya-Expanse-8B & 8B & 60.55 & 62.85 & \underline{81.18} & 78.54  & 76.12 & 56.40& 64.80 \\
Llama 3.1-Instruct  &8B&  67.92 &68.04& 80.74   & 79.22  &  74.91  &  55.29  & 73.88  \\

\textbf{\FP\ Instruct} &8.7B& 69.15   &70.97   & \textbf{81.99}
  & \textbf{83.06} & \textbf{83.02} & \textbf{64.93} & \textbf{78.69 }\\

Gemma-2-9b-it   &9B&   \underline{71.34}     & \underline{71.65} & 79.38 & 79.06 & 60.42  & \underline{63.99} & \underline{75.69} \\
Jais-13B-chat     &13B&   49.45    &48.46& 78.24  &  77.59 & 27.98  &  46.84 & 68.59 \\
Jais-adapted-13b-chat &13B& 55.63 &56.64& 80.47& \underline{80.86} &68.23  & 54.27 & 69.77\\

      && MT-Bench& IFEval & Alpaca LCW \\
    Jais-family-6p7b-chat & 6.7B &4.68 & 43.28 &  5.77 \\   
    OLMo-7B-0724-instruct-hf &7B& 5.64& 51.07& 8.53\\
    \textbf{\FS\ Instruct} &7B &5.16& 67.98  & 10.40\\
    ALLaM-7B-Instruct$^{**}$ &7B    & - &47.36 & - \\
    Qwen2.5-7B-Instruct &7B    &\underline{8.30} &66.67 & 32.78\\
    AceGPT-v2-8B-Chat  &8B   &7.14& 59.83 & 11.96 \\
    Aya-Expanse-8B & 8B & 7.17 &  66.90 & \underline{44.68} \\
    Llama 3.1-Instruct  &8B &7.62  &\textbf{82.25}&  27.31\\
    \textbf{\FP\ Instruct} &8.7B& 7.48& 74.82 & 27.71 \\
    Gemma-2-9b-it   &9B  &\textbf{8.38}&\underline{78.65} & \textbf{49.65} \\
    Jais-13B-chat     &13B &5.59&59.71 & 1.87 \\
    Jais-adapted-13b-chat &13B &4.18& 34.77& 8.33
    \end{tblr}
    }
    \caption{English benchmarks for instruction-tuned Fanar and baselines. We show the same benchmarks as the base model in the top, in addition to LLM-as-a-judge and verifiable generative tasks in the bottom. \FP\ Instruct offers competitive performance in English even though it was developed as an Arabic-centric model.\\
    \\\scriptsize{*~ALLaM results are reproduced from the technical report~\citep{bari2024allamlargelanguagemodels} and not validated with our setup}\\
    \scriptsize{** The ALLaM model was accessed via the Microsoft Azure cloud platform. MT-Bench and Alpaca were not evaluated during the time of access.}  
    }
    \label{tab:En_instruct}
\end{table}

\section{Integrating Multimodal Support}
\label{sec:multimodal}
\fullwidthbox{
 The training process for the speech and image modalities is explained, including efforts to ensure inclusive support for multi-dialectal Arabic speech and culturally aligned image generation. 
}

Fanar provides support for both speech and image modalities. Users of Fanar can interact with the models using speech as input and can receive the model-generated responses in audio format, enhancing accessibility and user experience. 
We leverage in-house state-of-the-art bilingual speech technology that supports both English and Arabic, including a wide range of Arabic dialects. 
For image modality our objective was to generate imagery which is culturally aligned with and appropriate for the Arabic-speaking region.

\subsection{Speech Modality} %

\begin{figure}[H]
    \centering
    \includegraphics[width=0.8\linewidth]{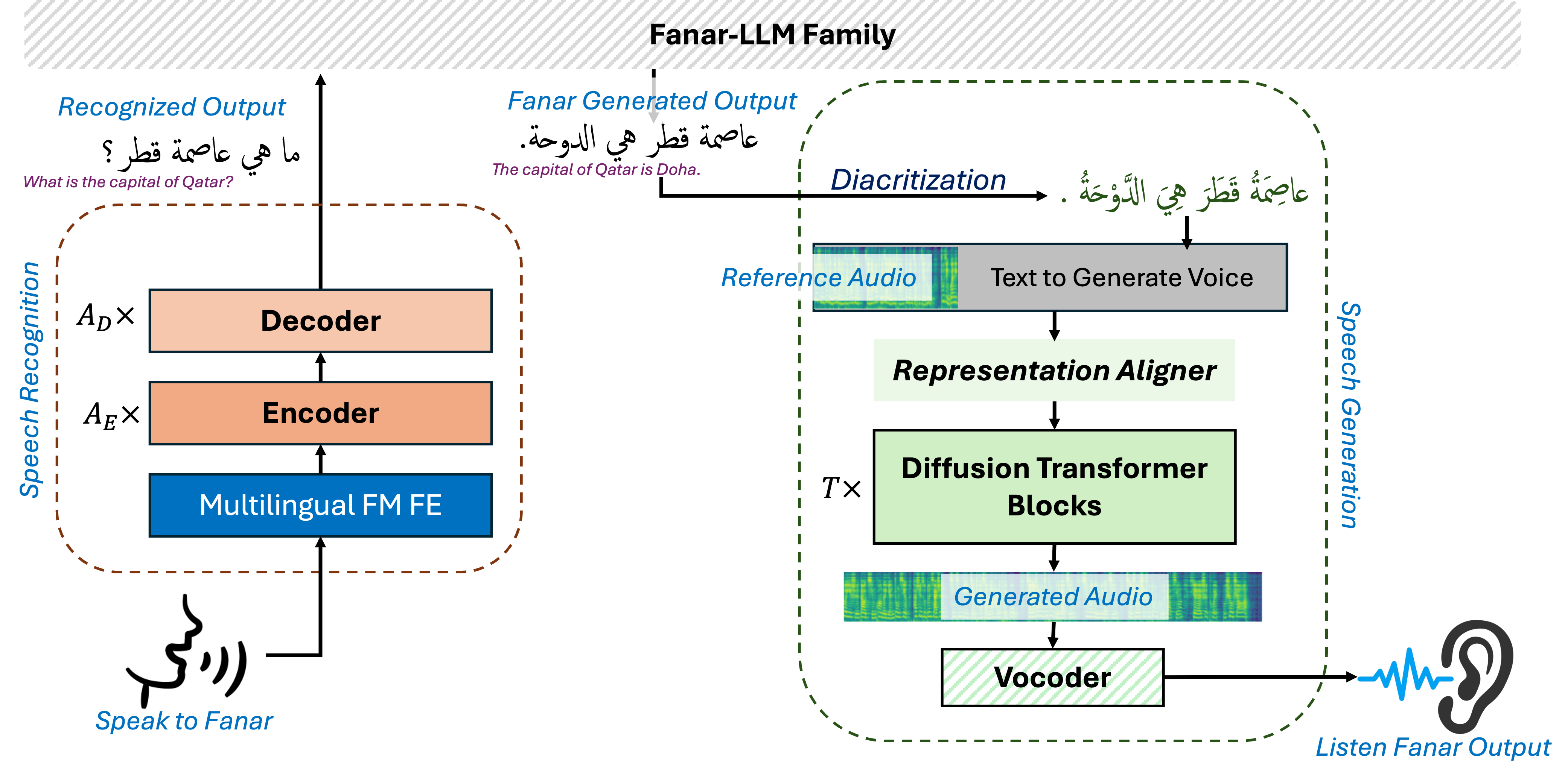}
    \caption{Overview of Fanar spoken interaction Capabilities enabled by automatic speech recognition (ASR) and text-to-speech generation (TTS) modules. FM: Foundation model, FE: feature extractor. $A_E$ and $A_D$ represent the numbers of encoder and decoder blocks respectively in the ASR, and $T$ represents number diffusion blocks in the models. }
    \label{fig:speech_mod}
\end{figure}

Speech is one of the most crucial and natural forms of interaction. This multifaceted modality does not just convey language but encapsulates human emotion, the intention behind their action, and personal identification while reflecting the speakers' surrounding condition. 
Understanding such rich information is paramount to achieving advanced AI capabilities. 
Therefore, with Fanar, we added support for processing spoken instruction and generating output with natural-sounding speech. An overview of Fanar's spoken interaction support is presented in Figure \ref{fig:speech_mod}.

\paragraph{Inclusive Arabic Speech Recognition System} Fanar integrates a state-of-the-art in-house Arabic-English automatic speech recognition (ASR) system that supports multiple Arabic dialects regardless of the speaker's nativeness, accentedness, or environmental context. 
Additionally, Fanar accommodates accented English and seamlessly handles code-switching between languages (Ar $\leftrightarrow$ En) and within dialects (e.g., MSA $\leftrightarrow$ EGY). 

\noindent\textbf{Model Design:} To design such an inclusive model, we leveraged a multilingual self-supervised foundation model (FM) to extract rich contexual representation from the raw speech input. 
We then passed the extracted  representation through the layers of conformer-based encoders, followed by transformer decoders. 
We then train the model utilizing multitask learning objectives -- combining the decoder's cross-entropy loss and CTC loss with a weighting factor, $\alpha$ \citep{chowdhury_towards_2021}. Finally, we average the weight of the best 10 model checkpoints based on the validation set. 

\noindent
\textbf{Training Strategy and Data:} We trained the speech model with a total of $\approx 15,000$ hours of data in English and Arabic. 
For Arabic, the dataset comprised both MSA and dialectal Arabic from diverse domains, such as broadcasts, podcasts, meetings, and academic conferences. Incorporating dialectal support was pivotal to Fanar for expanding accessibility - reaching people of varying education levels and age groups while upholding cultural inclusion \citep{arabic_2021}.
However, the scarcity of annotated dialectal speech data posed significant challenges. To address this, we employed data augmentation techniques and a customized curriculum-based batching strategy. The customized batching approach introduced dialects incrementally based on their phonetic and linguistic proximity to MSA, ensuring effective model generalization \citep{ElKheir2024Beyond}.
For English, the training data is selected from widely used publicly available speech datasets, capturing a wide range of speaking styles, domains, and accents. 
The model is highly Arabic-centric, with 75\% of the vocabulary tokens (out of total $10,000$ bpe tokens) representing the Arabic language, and only 25\% tokens are reserved for English. 

\begin{figure}[H]
    \centering
    \includegraphics[width=0.8\linewidth]{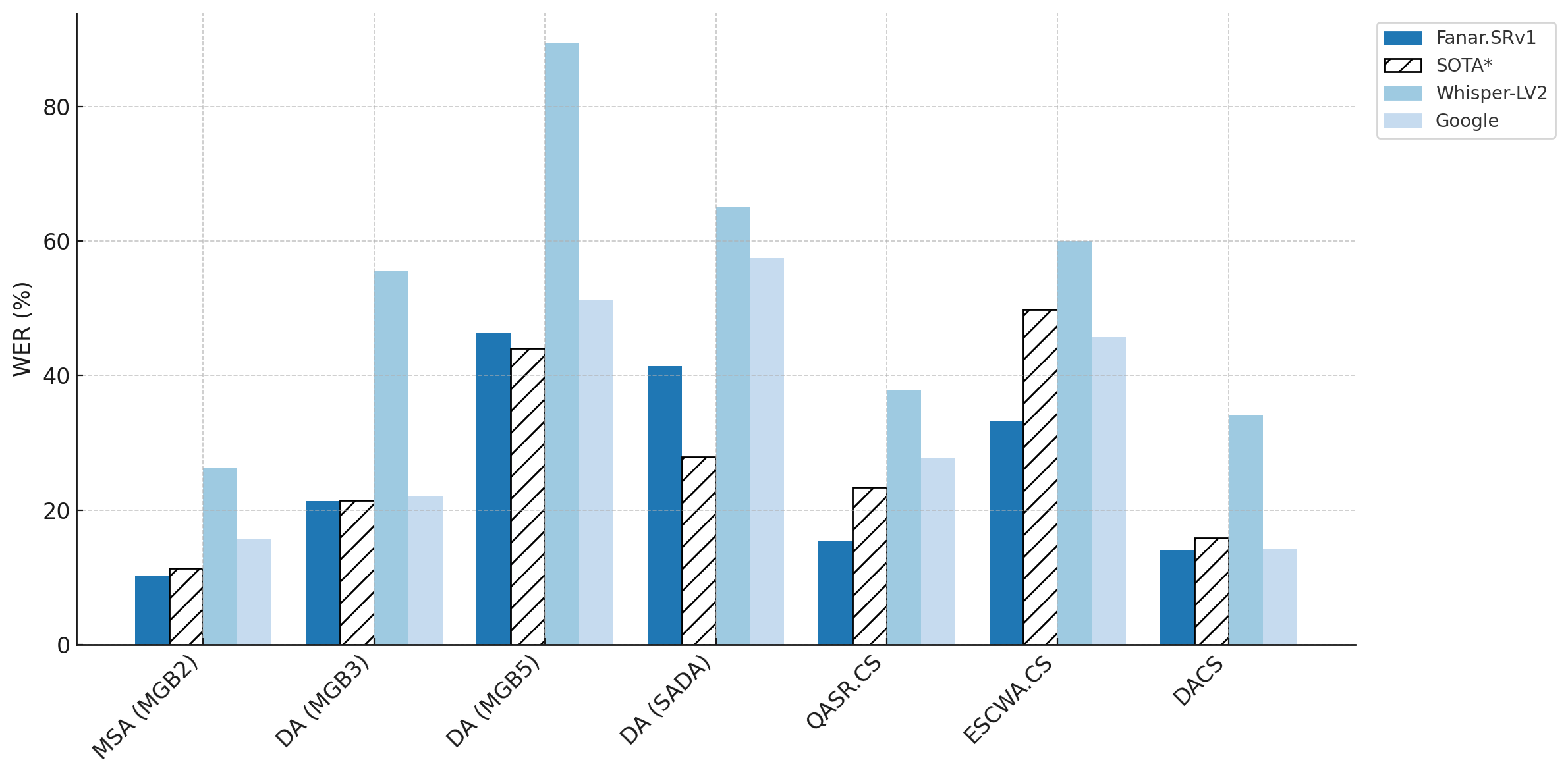}
    \caption{Reported word error rate (WER) on widely used Arabic testsets, presented in LAraBench \citep{abdelali-2024-larabench} showcasing Fanar's ASR performance compared to Google and OpenAI's whisper models. All SOTA and third-party ASR results are from LAraBench, except the SADA results which are from the SADA-dataset paper \citep{alharbi2024sada}. The SOTA results represent performance when the model is trained specifically on the training set portion of the test data. Note that SADA, ESCWA.CS test sets are completely unseen to the Fanar ASR model. }
    \label{fig:speech_mod_results}
\end{figure}

\noindent\textbf{Model Performance:} We evaluate the performance of Fanar SR using widely recognized test sets and compare the result with comprehensive benchmarks reported in LAraBench \citep{abdelali-2024-larabench}. 
Following LAraBench, for Arabic, we reported QASR-test \citep{mubarak_qasr_2021} for MSA and MGB3 \citep{ali2017speech},  MGB5 \citep{ali2019mgb} for dialectal performance. In addition, we also included performance for the latest SADA \citep{alharbi2024sada} dataset for evaluating dialectal varieties, and showcasing the model's inclusivity across regional dialects. For English, we benchmarked with the popular LibriSpeech \citep{panayotov2015librispeech} dataset.
For assessing the model's performance on code-switching tasks, we used the QASR.CS \citep{chowdhury_towards_2021} and ESCWA.CS \citep{chowdhury_towards_2021} for Arabic-English intra-utterance code-switching and DACS.CS \citep{chowdhury2020effects} for dialectal code-switching scenarios.

Our results, in Figure \ref{fig:speech_mod_results}, show the efficacy of the Fanar ASR. The results indicate that the in-house model consistently outperforms the available large-scale speech models such as Google ASR and Whisper-Largev2 for the LAraBench \citep{abdelali-2024-larabench} and SADA  \citep{alharbi2024sada} datasets. Notably, Fanar ASR also achieve significantly better or comparable results, in most of these datasets, with respect to the current in-domain systems (SOTA) reported in LAraBench. 
Fanar ASR is primarily an Arabic-centric speech recognition model. However, it demonstrates a satisfactory performance on English benchmarks, achieving a WER of 7.5\% on Libri-Clean, and 12\% on Libri-Other.
Remarkably, this performance is achieved while utilizing only 25\% of the model's vocabulary tokens, highlighting the model's capability of transcribing English in this constrained setting.

\paragraph{Text to Speech Generation}
To improve accessibility, Fanar integrates the state-of-the-art in-house Arabic text-to-speech (TTS) model following recent flow-based diffusion model architechture supporting in-context learning. 
The model outputs synthesized speech that is natural, culturally adaptive, and intelligible to a diverse audience. 

\noindent\textbf{Model Design:} To design the TTS, we opt for a fully non-autoregressive architecture that offers human-level naturalness and state-of-the-art speaker similarity and intelligibility with only small reference speaker audio. We adopted a flow matching strategy for generating mel spectrogram using diffusion transformer model (DiT) \citep{chen-etal-2024-f5tts,eskimez2024e2}. 

To generate the spoken response from Fanar, we first diacritize the input prompt using Farasa diacretizer \citep{abdelali-etal-2016-farasa}. Next, the diacritized text is converted into a sequence of characters and padded with filler tokens to match the length of the input speech.
Given the reference mel spectrogram of the speaker (audio prompt) and the text representation, the model exploit ConvnextV2 \citep{woo2023convnext} to address text-speech alignment within the in-context learning framework. 
The aligned representation is then passed through DiT blocks to generate the text and reference audio conditioned mel spectrogram. We then converted the mel spectrogram into the final raw speech output using the vocoder \citep{siuzdak2023vocos}.
Unlike traditional approaches, our adapted architecture eliminated the need for duration models, text encoders, or phoneme alignment, making it well suited for phonetically rich and diverse languages like Arabic.

\subsection{Image Modality}

We developed the image generation model of Fanar by fine-tuning Stable Cascade~\citep{pernias2023wuerstchen}. Compared to other open-source projects, Stable Cascade has satisfactory characteristics, including generation quality, generation speed, configurability, documentation and license.

Stable Cascade operates at three steps. The first step is stage C where given a textual prompt, a small (typically $12 \times 12$) image is created with tens of channels. This image is given to stage B which upscales the image to $256 \times 256$ with $4$ channels. Stage B is not conditioned on the prompt. Finally, the output of stage B is given to stage A which upscales the image to $1024\times 1024$ with three channels. The models for the three stages C, B and A have respectively 3.6B, 700M and 20M parameters. Stages B and A primarily increasing image resolution. We focus on fine tuning stage C to achieve our desired results.

Stable Cascade works with English Prompts. We support Arabic and English. If we are given an Arabic prompt, we query the model with the English translation of the prompt. If the prompt is in English, we use the same prompt. We augment the text prompt with some textual descriptions to help increase the quality and the details of the result.

\textbf{Problems with Stable Cascade}: We observe two problems with publicly available open source models. We refer to these problems as the problems of knowledge and preferences. The problem of knowledge is similar to variance problem which refers to lack of knowledge about certain topics of interest. The problem of preference refers to the bias of the model and the training data.

\textbf{Problem of Knowledge}: The base Stable Cascade model is trained using LAION-5B~\citep{schuhmann2022laion5bopenlargescaledataset} dataset. In this dataset, many topics familiar in the Western culture are represented. However, some concepts that are familiar in Qatar and the broader Middle East are under represented in this dataset. For example, the model generates a more elaborate image for ``American children playing basketball" than for ``Qatari children playing basketball". In practice, the base Stable Cascade model generates higher quality and more accurate images in the American context, while they do poorly in the Qatari and Middle Eastern context. Since the knowledge about Arab culture is represented in only a very small fraction of training data, even if the size of training data is increased, the model cannot desirably learn the local topics. This is because the limited capacity of the model is allocated proportionally to the training data. Therefore, if we increase the size of data, still certain concepts will be inevitably under-represented in the training data.

\textbf{Problem of Preference}: Knowledge and preference are often correlated, however, this is not always the case. In human psychology, familiarity heuristic~\citep{tversky1974judgment} or ``information bias" is a useful heuristic where concepts that are more familiar, are also generally preferred more. Even though this heuristic is helpful, it can also lead to some adverse effects, including discrimination, archetypes, misalignment, and safety issues. For example, a person may be familiar with violence, but does not prefer it. We observe that no matter how we train our model, this information bias will persist unless we explicitly enforce preferences.

\textbf{Solution to the Knowledge problem}: No matter how we increase data size, certain important concepts will be under-represented in the training data. To make sure we exhaustively cover important cultural concepts we took a principled approach. We prepared a taxonomy of visual concepts specific to the Arab world and the Middle East. This taxonomy tree covers familiar visual concepts including clothing, architecture, natural scenes, familiar people, and everyday life from Arab and Middle Eastern countries. To make sure important concepts are not missed in the training data, our taxonomy has four layers. For example, we have: Food and Drink $\rightarrow$ Beverages $\rightarrow$ Pakistan $\rightarrow$ Falooda. For each concept within this taxonomy tree, we collected several related images. Using this process, we had more fine grained control over the topics that we wanted our model to be familiar with.

To create the taxonomy tree, we used Wikipedia articles from multiple languages and performed some manual adjustment. Our taxonomy tree includes more than 5000 visual concepts in four layers of abstraction. We used the elements of this taxonomy tree to obtain images which we use to create a dataset of about 200,000 images with captions. We then applied quality filters (e.g., ignore low resolution or grayscale images) and reduced the data to about 100,000 high-quality images. We fine-tuned several image generation models using several different sources of data. In addition, we collected images from cultural Wikipedia pages and obtained other images from Qatari sources. For each source of data we created a separate dataset. We then fine-tuned a separate model given each dataset.

\begin{figure}[ht!]
    \centering
        \includegraphics[width=\linewidth]{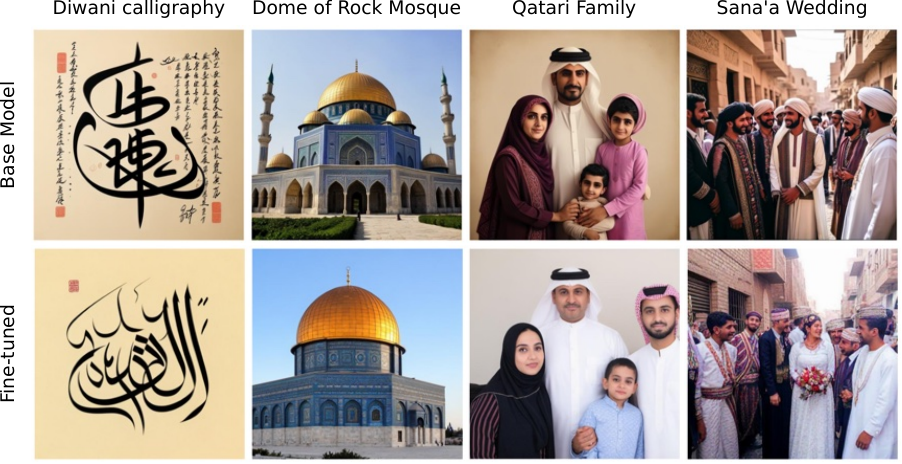}
    \caption{Comparison between images generated by the base Stable Cascade model versus our fined-tuned version. Column names indicate input prompt. The full prompt for the right-most column is: ``A picture of a wedding party in the streets of old Sana'a and a group of people are happy and wearing traditional dress''. Notice that the images from the fine-tuned model are more culturally nuanced compared to the original counterparts. The corresponding images are generated using identical initializations.}
    \label{fig:image_modality}
\end{figure}

\textbf{Model Averaging}: Knowing the ideal proportion of different datasets in the training data is difficult before training; therefore, we used a training process to have more principled control over the mix of data. We partitioned the training data into a few categories. Then, for each category we fine tuned a separate model. Finally, we averaged the models after they were trained. Then, a group of users reviewed hundreds of generated images and assigned a quality score to each fine-tuned model. We then picked weights to each model accordingly and averaged the best performing models. The final average performs better than each individual model. Figure~\ref{fig:image_modality} compares the outputs of our final averaged model vs the base model.

\textbf{Solution to the Preference problem}: After we adjusted the distribution of our training dataset and used it to train models, there are still unwanted biases in our model. To fix these biases, we first input a list of preferences in textual format in the form of ``A should B" or ``A should not B". (For example ``Muslims should not drink alcohol"). Then we create three textual prompts for image generation: neutral prompt, negative prompt, and positive prompt. Neutral prompt contains A and its variations. Negative prompt contains A and not B. Positive prompt contains A and B. Then we generate three images using the three prompts using the same Stable Cascade model. Finally, we use DPO to train the model to prefer ``A and B" over ``A and not B" given prompt A.

\section{Retrieval Augmented Generation}
\label{sec:rags}
\fullwidthbox{
 The various RAG systems used in Fanar are explained. These systems handle prompts related to specific domains in order to improve the accuracy of model-generated content.
}

Fanar extensively uses Retrieval Augmented Generation (RAG) in various forms. RAGs are used for generating Islamic content, biographical
information, accessing recent information and for attributing fact-related prompts.

\subsection{Islamic RAG}
Islamic RAG consists of two main components. The first is a data pipeline designed to store curated texts from Islamic literature within a vector database (Milvus). This process involves text chunking, where each document is divided into segments of up to 2048 tokens, with a 50-token overlap between consecutive segments. Chunking is performed while ensuring that splits occur at semantic boundaries (e.g., paragraphs or lines) to preserve coherence. The hyperparameters of the RAG system, including the embeddings model, chunk size, re-ranking strategy, and others, were fine-tuned using carefully selected domain-specific benchmark datasets.
The second component is the prompt engine. When a prompt is submitted, a similarity search is conducted within the vector database to retrieve the top twenty documents that match the prompt and exceed a predefined similarity threshold. This threshold was calibrated using the same benchmark datasets. The retrieved documents are subsequently processed by a re-ranking model, which selects the top four documents based on relevance. These documents are further filtered to fit within the context window of the language model. Finally, a prompt is constructed and sent to Fanar, instructing it to generate a response based on the provided documents. The prompt design was optimized to maximize context utilization and minimize out-of-context references.

\subsection{Recency and Biography RAG}
An orchestrator uses a classifier to determine whether a prompt is related to recent information, e.g., ``what is the weather in Doha", or
biographical information, e.g., ``who is Mohamed Salah,''. The prompt is routed to the Recency or the Biography RAG where it
rewritten as search query by \FP\ and relevant information is accessed through a search engine. The results of the search are provided
to \FP\ which summarizes the information and creates a response with the included source.

\subsection{Attribution RAG}
\label{sec:attribution_rag}

Attribution refers to the capacity of an LLM model to generate and provide evidence, often in the form of references or citations, that substantiates the claims or statements it produces~\citep{bohnet2022attributed}. This evidence is derived from identifiable sources, such as webpages with unique URLs, ensuring that the claims can be logically inferred, comprehended, and verified by LLM users (e.g., chatbot users). LLM attribution is used to enable users to validate the claims made by the model, promote the generation of responses that align with the cited sources, and to establish a framework for evaluating the completeness and relevance of the supporting evidence in relation to the presented claims~\citep{li2023survey}.

There are three main types of LLM attribution~\citep{li2023survey}: (1) direct model-driven attribution~\citep{sun2022recitation}, where the model itself provides the attribution for its generated responses, (2) post-retrieval answering~\citep{xu2023search}, where information is explicitly retrieved from specific references and then the model is conditioned on this information to generate responses, and (3) post-generation attribution~\citep{gao2022rarr}, where the generated response is used along with the user's query to search for supporting references from a large corpus, after which the response is modified and attributed using the references.

We decided to avoid making direct modifications to Fanar models, as that may affect the way they generate responses and other down-stream tasks (e.g., model evaluation, guardrails). As such, attribution in Fanar is an opt-in feature and uses post-generation attribution, making it a stand-alone, independent service. In particular, we view attribution as a post-generation RAG task that performs fact-checking and revision of Fanar responses. It takes in a user query and Fanar's response to the query, and then uses an LLM model to generate a revised response with references based on top-$k$ most relevant documents from a  corpus of facts-related documents (e.g., Wikipedia articles, Data Commons factoids). In concept, attribution in Fanar implements a research-then-revise methodology~\citep{gao2022rarr} using prompt engineering, where relevant evidence is retrieved from the corpus for attribution and the response is minimally modified to make it consistent with the evidence while preserving its properties, such as style and structure.

\section{Discussion and Future Plans}
\label{sec:future}
In this work we have presented our first release of Fanar, an Arabic-centric multimodal generative AI platform. The core capabilities of Fanar are encapsulated in \FS\ and \FP,\, two LLMs  that operate concurrently, each  responsible for answering different
types of prompts. \FS\ is a 7B parameter model trained from scratch on nearly 1 trillion tokens including 300 billion tokens in original Arabic and nearly  50 billion tokens in translated MSA and dialectal Arabic to enhance coverage. The \FP\ model is continually trained on the base Gemma-2 9B model with the same 1 trillion set.  In addition, the  Fanar platform provides multimodal capability in speech and image generation and four RAG systems
for Islamic content, recent information, factual content attribution and select biographies. An overarching orchestrator coordinates
all the Fanar services and enables safety filters to verify content at the prompt and response level. During the course of the project we have learned several lessons and identified several directions for future research and
development.

\begin{enumerate}
\item Lack of Arabic data (both MSA and dialectal) that has both broad coverage and clean is the biggest bottleneck towards building
a very large scale Arabic-centric LLM. Coupling \FS\ with \FP\ is our pragmatic approach to take advantage of the coverage of English data while
maintaining Arabic characteristics. On Arabic-centric benchmarks,  results indeed show that our solution is
robust. With human testers the Fanar ``dislike'' rate for prompts is around 15\% and most of it due to 
factuality related errors which is an indicator that we need to  scale to bigger models. We have further attempted to
reduce errors by extensively using RAGs and supporting attribution. However new approaches maybe required to increase 
the confidence of using LLMs for critical tasks.
\item 
While the current version of Fanar platform has an orchestrator module to coordinate the prompts,
it is not agentic. Fanar is reactive rather than pro-active whereby it can reason and plan with minimal
guidance and pursue goals autonomously. The next generation of Fanar  will include forms of reasoning
and verification capabilities either endogenously or through tool calling. An emerging area is to explore the 
use of enhanced test time computation for hard tasks (e.g.,  math reasoning) where multiple responses are generated
and an external reward models selects the best answer~\citep{snell2024scalingllmtesttimecompute}. 
\item The Fanar platform has multimodal capabilities but they are not intrinsic to 
the autoregressive model with image and speech tokens that can be combined to perform the appropriate
multimodality task. For the next version, we plan to make speech, image and text generation
as part of a unified generative model. We will also provide support for video generation that
is both culturally appropriate and efficient.
\item 
The first release of the Fanar platform is accompanied by a small class of applications in education,
news summarization and a chatbot to access government information. Our goal is to integrate Fanar
at the enterprise level including in the government and the private sector. For Fanar to evolve and thrive,
real use cases with visible productivity benefits  will have to drive its evolution.
\item Finally, we return to the data issue. In the current landscape of GenAI, web publishers are increasingly grappling with how to maintain free and open access to their content while ensuring fair compensation. With the rise of LLM-driven tools such as ChatGPT, publishers often see diminished traffic to their websites, leading to reduced advertisement revenue. Moreover, legal questions are emerging regarding whether LLM providers should pay for access to copyrighted content that are used to train their models. As a result, over 5\% of all Web data and 25\% of high-quality Web sources in AI training are now restricted, mainly due to publishers putting their content behind paywalls and login screens~\citep{longpre2024consent}. QCRI has launched a parallel project, \textbf{TokenX}, to address the data issue in a principled, market-driven manner. In particular, TokenX provides a new service that leverages blockchain technology and LLM attribution (see \S\ref{sec:attribution_rag}) to incentivize publishers to keep their content openly-accessible by paying them whenever the content is used for attribution by subscribed LLM providers. We aim to use TokenX to encourage Arabic speakers to generate content both in MSA and dialects across a wide range of disciplines and possibly create an Arabic language renaissance.
\end{enumerate}

\newpage
\appendix
\section{Contributions}
\label{sec:contribution}
The Fanar project was divided into work packages and sub-projects and the specific contributions of
the authors are listed below. The names within each sub-group are ordered alphabetically. 

\noindent
\textit{Data Collection and Quality:} Hamdy Mubarak (lead), Mohammad Shahmeer Ahmad, Sabri Boughorbel, Abubakr Mohamed, Tasnim Mohiuddin, Ahmad Musleh, Zan Naeem, Omar Sinan, Yifan Zhang  \\
\textit{Pre-Training ({Fanar Star}):} Sabri Boughorbel (lead), Mohammad Shahmeer Ahmad, Fahim Dalvi, Tasnim Mohiuddin, Zan Naeem,  Amin Sadeghi \\
\textit{Pre-Training ({Fanar Prime}):} Fahim Dalvi (lead), Sabri Boughorbel, Tasnim Mohiuddin, Amin Sadeghi \\
\textit{Post-Training:} Husrev Taha Sencar (lead), Enes Altinisik, Masoomali Fatehkia \\
\textit{Benchmarking and Evaluation:}  Majd Hawasly and Ehsannedin Asgari (leads),  Sabri Boughorbel, Fahim Dalvi, Mus'ab Husaini, Hamdy Mubarak \\
\textit{Fanar Morphology-based Tokenizer:} Ehsannedin Asgari (lead) \\
\textit{Machine Translation:} Nadir Durrani (lead), Fahim Dalvi, Kareem Darwish, Basel Mousi \\
\textit{Image Generation:} Amin Sadeghi (lead), Ehsannedin Asgari, Kareem Darwish, Abubakr Mohamed, Hamdy Mubarak, Zan Naeem \\
\textit{Speech:} Shammur Chowdhury (lead), Mohammad Shahmeer Ahmad, Kareem Darwish, Abubakr Mohamed, Hamdy Mubarak \\
\textit{Applications:} Kareem Darwish (lead), Mohammad Shahmeer Ahmad, Yifan Zhang \\
\textit{Engineering and Deployment:} Mohamed Elfeky (lead), Ummar Abbas, Fahim Dalvi, Mus'ab Husaini, Soon-Gyo Jung, Ji Kim Lucas,  Yifan Zhang\\
\textit{Islamic RAG:} Ummar Abbas (lead),  Kareem Darwish, Walid Magdy, Hamdy Mubarak,   Mourad Ouzzani, Omar Sinan, Mohammed Shinoy \\
\textit{Attribution:} Yazan	Boshmaf (lead), Soon-Gyo Jung, Ji Kim Lucas, Yifan Zhang \\
\textit{Infrastructure:} Anastasios Fragkopoulos (lead) \\
\textit{External User Study and System testing:} Hamdy Mubarak and Majd Hawasly (leads) \\
\textit{Scientific and Management Leadership:} Mohamed	Eltabakh (lead), Sanjay	Chawla, Ahmed 	Elmagarmid, Mourad	Ouzzani

\subsection{Acknowledgments}
A project of this scope would not have been possible without contributions from a diverse array of individuals and partner organizations. We would like to express our heartfelt gratitude to all who have helped support Fanar's development.

We would like to thank Qatar's  Ministry of Communications and Information Technology (MCIT) for sponsorship of the project; our partners, LiverX for project management and Google Cloud for infrastructure support.  We would further like to thank our partners at the Arab Center for Research and Policy Studies for evaluation and benchmarking Fanar using custom in-house benchmarks, as well as our data provider partners, Al Jazeera, Qatar University, Qatar National Library (QNL) and the Ministry of Endowments and Islamic Affairs.

Special thanks goes to Omid Ghahroodi, Nizi Nazar, Hind AL-Merekhi, Fatema Ahmad, Marzia Nouri, Pardissadat Zahraei, Aisha Hamad M A Al-Naimi, Reema Khater, Mohammad Ali Sadraei, Ali Nazari, Mohammad Ali Seif Kashani, Bryson Jandwa and Michael Aupetit.
  Finally, we would like to express our gratitude to numerous volunteer testers at Qatar University and across different Arab countries whose valuable feedback has enabled us to improve Fanar. A complete list of volunteers will be available at \url{https://fanar.qa/volunteers}.

\newpage
\section{Arabic-Speaking Map and Language Statistics}
\begin{figure}[ht!]
    \centering
    \begin{subfigure}[t]{0.9\textwidth} %
        \centering
    \includegraphics[width=\textwidth, trim={2cm 1.5cm 2cm 1.5cm}, clip]{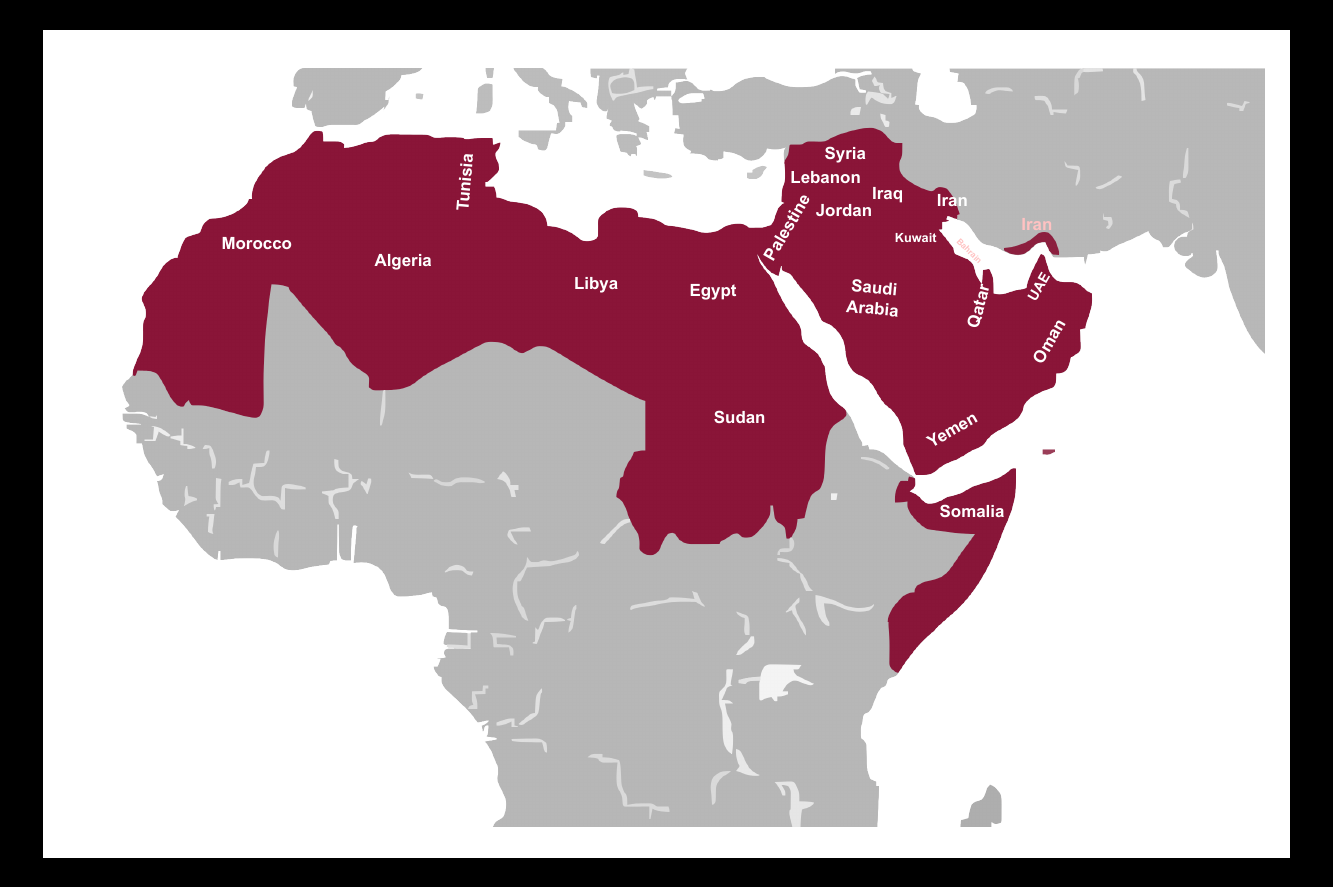} %
        \caption{Overview of Arabic-speaking Countries and Regions in the Middle East, North Africa, and the Arabian Peninsula (MENA Region).}
        \label{fig:top}
    \end{subfigure}
    
    \vspace{0.5cm} %

    \begin{subfigure}[t]{1\textwidth} %
        \centering
    \includegraphics[width=\textwidth]{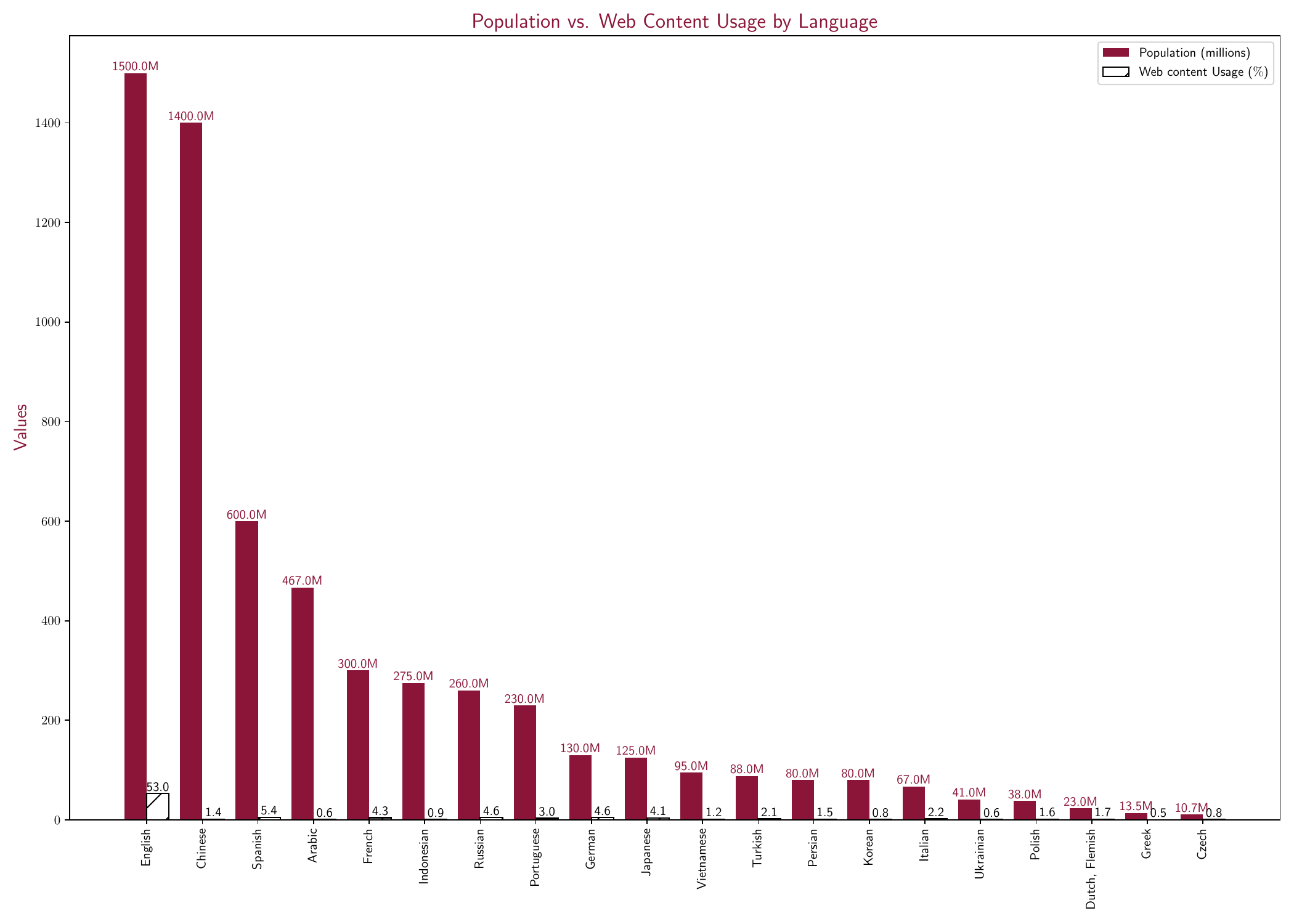} %
        \caption{Contrasting the language population and web-data availability of different languages.}
        \label{fig:bottom-left}
    \end{subfigure}

    \caption{(a) Overview of Arabic-speaking countries; (b) Population vs. web content statistics for the top 10 most popular languages in the world.}
    \label{fig:panel}
\end{figure}

\section{In-house Benchmarks}

\subsection{Almieyar: capability-based benchmarking}
\label{appx:Almieyar}

The evaluation of LLMs has emerged as a critical research focus, frequently discussed at recent ACL and ML conferences, particularly in 2023 and 2024. Answering benchmark questions in many existing evaluation datasets, such as MMLU \citep{MMMLU}, requires a mix of skills. For example, solving a logic puzzle often requires a combination of linguistic proficiency, logical reasoning, and domain expertise. However, this entanglement makes it challenging to provide clear feedback on the specific strengths and weaknesses of the models. In our view, a diverse set of essential capabilities should be benchmarked for LLMs. To this end, we introduce Al-Mieyar, a family of capability-focused benchmarks that target Arabic as primary languages/culture, while also providing a recipe that can be adapted to any language/culture of interest. Al-Mieyar targets a set of LLM-related capabilities, which will be introduced in a separate report. Here, we focus specifically on its language capabilities segment, Almieyar-Language.

\noindent\textbf{Almieyar-Recipe:} The creation of the Almieyar dataset involved three steps: (i) template-based question generation using an LLM guided by a refined taxonomy, (ii) human review by native Arabic speakers for linguistic and cultural accuracy, and (iii) linguistic validation by professionals to ensure quality.

\begin{figure}[ht!]
    \centering
    \includegraphics[width=0.8\textwidth]{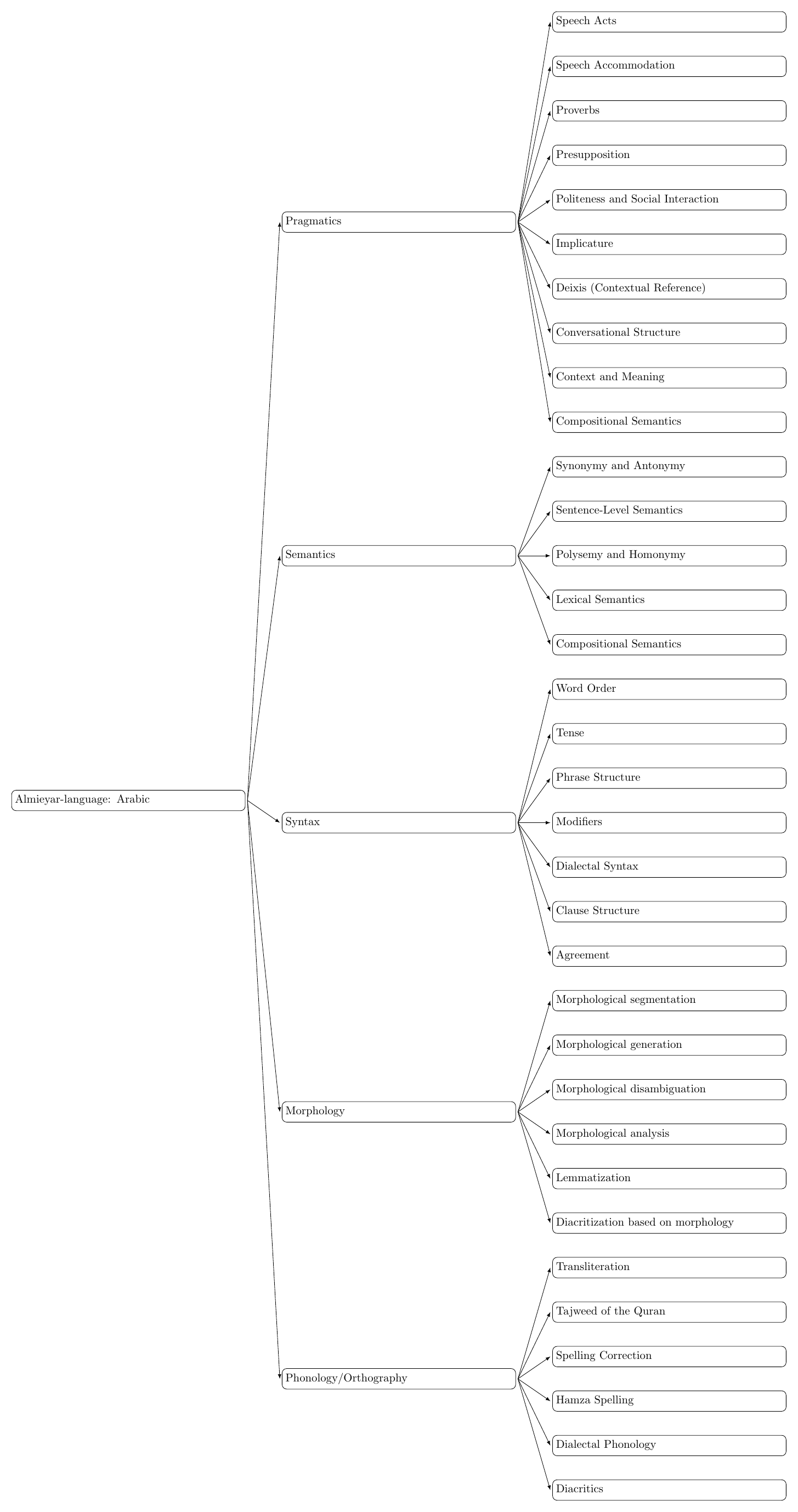} 
    \caption{Hierarchical representation of Almieyar-language categories and sub-categories in Arabic: Phonology/Orthography, Morphology, Syntax, Semantics, and Pragmatics. Each category is further detailed with its sub-categories.}
    \label{fig:almieyarlanguagetree}
\end{figure}

\noindent\textbf{Almieyar-Language:} despite advances in benchmarks targeting common sense and logical reasoning, relatively few focus exclusively on linguistic understanding. To the best of our knowledge, \textit{Holmes} is one of the few datasets in this area \citep{waldis2024holmes}, but it focuses solely on English and lacks detailed taxonomies for linguistic evaluation. Furthermore, it excludes important linguistic categories, such as phonology, and provides no resources for non-English languages. To address these limitations, we introduce \textbf{Almieyar-language}, a benchmark designed to evaluate linguistic understanding in Arabic, along with a recipe for extending it to other languages. \textit{Almieyar-language} is grounded in linguistic capabilities and offers a structured approach to dataset generation and evaluation.

Linguistic theory classifies language knowledge into five core layers: phonology, morphology, syntax, semantics, and pragmatics. We developed a comprehensive taxonomy that encompasses these layers from a universal perspective, with a particular focus on the unique characteristics of Arabic. This focus is informed by the specific features of Arabic outlined in \citep{habash2010introduction} for each of the five layers. The taxonomy of linguistic capabilities in Arabic is summarized in Figure \ref{fig:almieyarlanguagetree}.

 By addressing Arabic’s rich dialectal diversity, with contributions from speakers of 16 dialects, Almieyar-language emphasizes lexical semantics and pragmatics, including dialect-specific elements. The evaluation of the Fanar model on the ``Almieyar-language'' dataset, in comparison to other prominent models, provides detailed insights into its performance across the five core linguistic layers, assessed over a high-quality set of approximately one thousand questions, drawn from diverse linguistic categories.

 \begin{figure}[ht!]
    \centering
    \includegraphics[width=0.8\textwidth]{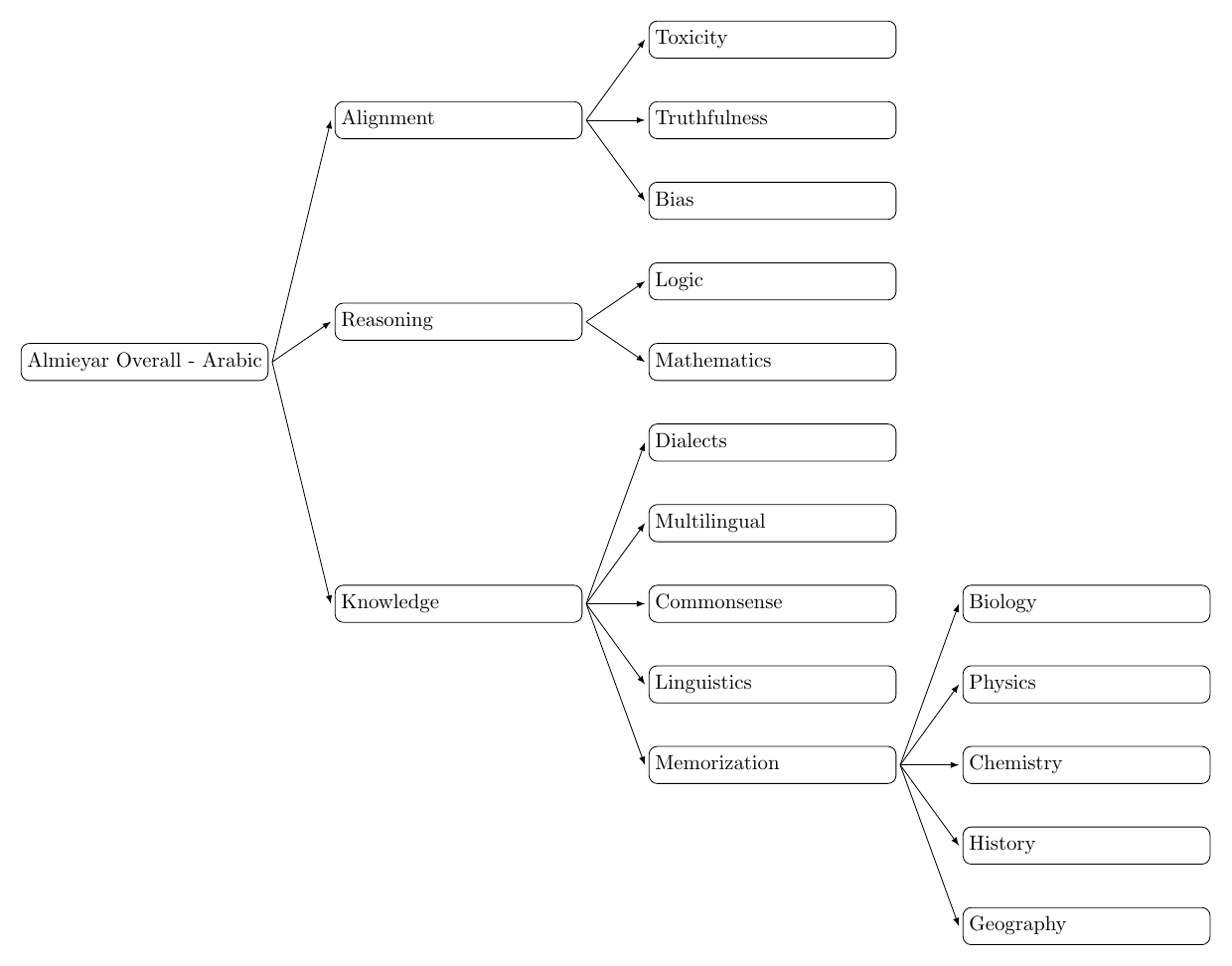} 
    \caption{Hierarchical structure of the ``Almieyar-Overall'' segment, encompassing high-quality benchmarking questions on knowledge, reasoning, alignment, developed to provide quick feedback following SFT training.}
    \label{fig:almieyaroveralltree}
\end{figure}

\noindent\textbf{Almieyar-Overall:} This dataset contains high-quality and manually reviewed benchmarking questions on knowledge, reasoning, alignment, and robustness, in Arabic is another segment of Almieyar developed in-house to provide quick feedback on a diverse range of topics following SFT training. Figure \ref{fig:almieyaroveralltree} illustrates the hierarchical structure of the ``\textbf{Almieyar-Overall}'' segment.
 ``\textbf{Almieyar-Culture}'' and ``\textbf{Almieyar-Multimodality}'' are also among the segments that will be detailed in a separate report.

\subsection{Arab Cultural MCQ}
\label{appx:CulturalMCQ}

To help bridge the gap in high-quality benchmarks for Arabic culture, we created a new multi-choice question dataset that addresses cultural issues, values and nuances in Arab countries. The dataset comprises 1K questions that were curated carefully from a corpus of 7.5K generated automatically by GPT-4o. The LLM was presented with relevant web pages that contain high-quality content about Arabic culture\footnote{E.g., \url{https://culturecrossing.net/}, \url{https://www.expatica.com/}, Wikipedia culture pages.}, and instructed to create a number of multi-choice questions that revolve around the input text.

Then, the questions went through three stages of thorough validation and quality control:
\begin{enumerate}
\item An expert reviewed the language and the factuality of each question from the 7.5K corpus. In certain cases, the question body or some of the choices were changed by the expert.
\item Another round of review selected questions that are relevant to the Arabic culture, not particularly easy or obvious, and that have meaningful answer choices.
\item Lastly, semantic clustering was performed on the reviewed questions. This used \texttt{bge-multilingual\\-gemma2}\footnote{https://huggingface.co/BAAI/bge-multilingual-gemma2}, an LLM-based multilingual embedding model with good performance in Arabic~\citep{bge-m3}. The questions with the highest similarity were reviewed manually again to remove the redundant questions, resulting in 1K questions. 
\end{enumerate}

Below are a few examples from the culture questions development set. %

\begin{enumerate}
    \item \noindent
\begin{RLtext}
هذا سؤال عن الثقافة والعادات في السودان. ما اسم الشراب الذي يتم تحضيره في رمضان في السودان ويعتبر مميزًا؟
\end{RLtext}
    \begin{RLtext}- العصير الطبيعي \\ \end{RLtext}
    \begin{RLtext}- الكركديه \\ \end{RLtext}
    \begin{RLtext}- التبلدي \\ \end{RLtext}    {\color{green}\begin{RLtext}- الحلو مر	\end{RLtext}} 
    
\item 
\noindent
\begin{RLtext}
هذا سؤال عن الثقافة والعادات في قطر. في أي مناسبات ينشد النهام أغانيه؟
\end{RLtext}
 {\color{green}\begin{RLtext}- أثناء الرحلات البحرية\end{RLtext}}
\begin{RLtext}- في الحفلات الموسيقية\end{RLtext}
 \begin{RLtext}- في الاجتماعات العائلية\end{RLtext}
 \begin{RLtext}- خلال فعاليات اجتماعية عامة\end{RLtext} 

\item 
\noindent\begin{RLtext}
هذا سؤال عن الثقافة والعادات في اليمن. ما هي طريقة الصباغة الأكثر شهرة في اليمن؟
\end{RLtext}
     \begin{RLtext}- التلوين الحر\end{RLtext}
     \begin{RLtext}- الرش\end{RLtext}
     {\color{green}\begin{RLtext}- الوصايل\end{RLtext}}
     \begin{RLtext}- الغمر\end{RLtext}

\end{enumerate}

\section{LLM Security and Safety}
LLMs, like any other software system, are vulnerable to attacks from malicious adversaries. At QCRI we have a parallel project that evaluates the safety of LLMs~\citep{ae}.
\textbf{aiXamine}
includes over 30 tests that cover a range of security threats and safety vulnerabilities. The tests evaluate both data and model safety. For data vulnerabilities, these tests can detect issues such as poisoned samples that inject backdoors, biases or noise that compromise performance, and samples that may violate copyright laws. For models, tests are designed to evaluate robustness against adversarial and privacy attacks, detect backdoors and unfair or biased behavior, check the model’s tendency to produce hallucinations and refuse innocuous requests, evaluate performance on out-of-distribution examples, verify watermarks, and assess content safety to prevent harmful or inappropriate outputs in the form of text or code. Table~\ref{tab:image-datasets} below
shows the results of \textbf{aiXamine} on select open source models and how they compare against 
\FS.

\begin{table*}[h]
\resizebox{\textwidth}{!}{%
  \begin{tabular}{lcccccc}
    \toprule
    \textbf{Model} & \multicolumn{1}{c}{\textbf{Adversarial}} & \multicolumn{1}{c}{\textbf{Code}} & \multicolumn{1}{c}{\textbf{Model \& Data}} & \multicolumn{1}{c}{\textbf{OOD}} & \multicolumn{1}{c}{\textbf{Safety \&}} & \textbf{Average} \\
    \textbf{Name} & \multicolumn{1}{c}{\textbf{Robustness}} & \multicolumn{1}{c}{\textbf{Security}} & \multicolumn{1}{c}{\textbf{Privacy}} & \multicolumn{1}{c}{\textbf{Robustness}} & \multicolumn{1}{c}{\textbf{Alignment}} & \textbf{Score}\\
    \midrule
    Fanar-7B & 58.24 & \textbf{85.8} & 81.14 & 89.6 & 97.68 & \textbf{82.49}\\
    LLaMA-3.2-3B-Instruct & 43.38  & 49.85 & \textbf{99.82} & 86.07 & 95.14 & 74.85\\
    LLaMA-3.1-8B-Instruct & 57.48 & 49.1 & 77.05 & 87.63 & 97.07 & 73.67\\
    Qwen-2-7B-Instruct & \textbf{61.78} & 46.18 & - & 90.65 & 95.32 & 73.48\\
    Llama-2-7b-chat-hf & 51.24 & 48.87 & 79.07 & 85.93 & \textbf{99.35} & 72.89 \\
    Llama-3-8B-Instruct & 49.97 & 48.37 & 74.53 & \textbf{91.40} & 97.52 & 72.36\\
    Mistral-7B-Instruct-v0.1 & 45.00 & 46.00 & - & 88.47 & 71.28 & 62.69 \\
    Mistral-7B-Instruct-v0.2 & 46.02 & 45.59 & - & 70.56 & 84.44 & 61.65\\
    gemma-2-9b-it & 16.62 & 50.53 & 96.00 & 49.62 & 92.57 & 61.07\\
    \bottomrule
  \end{tabular}
  }
  \caption{Comparison of scores achieved by different models across the different aiXamine services.}
  \label{tab:image-datasets}
\end{table*}

\bibliographystyle{apalikeModified}
\bibliography{fanar_ref,anthology}
\end{document}